\let\mypdfximage\pdfximage
\def\pdfximage{\immediate\mypdfximage}
\documentclass[times,twocolumn,final,authoryear]{elsarticle}
\usepackage{ycviu}

\usepackage[dvipsnames,svgnames,x11names]{xcolor}
\usepackage{tikz}
\usetikzlibrary{arrows.meta,shapes,calc,matrix,fit,positioning,backgrounds,decorations.markings,fadings}
\usepackage{pgfplots}
\usepackage{pgfplotstable}
\pgfplotsset{compat=1.9}
\usepackage{xstring}

\usepgfplotslibrary{external}
\newcommand{\extfig}[2]{\tikzsetnextfilename{#1}{#2}}

\IfBeginWith*{\jobname}{fig/extern/}{\finalcopy}{}

\tikzstyle{every picture}+=[
	remember picture,
	every text node part/.style={align=center},
	every matrix/.append style={ampersand replacement=\&},
]
\tikzstyle{tight} = [inner sep=0pt,outer sep=0pt]
\tikzstyle{node}  = [draw,circle,tight,minimum size=12pt,anchor=center]
\tikzstyle{op}    = [draw,circle,tight]
\tikzstyle{dot}   = [fill,draw,circle,inner sep=1pt,outer sep=0]
\tikzstyle{pt}    = [fill,draw,circle,inner sep=1.5pt,outer sep=.2pt]
\tikzstyle{box}   = [draw,rectangle,inner sep=3pt]
\tikzstyle{high}  = [black!60]
\tikzstyle{group} = [high,box,opacity=.5]
\tikzstyle{dim1}  = [fill opacity=.3,text opacity=1]
\tikzstyle{dim2}  = [fill opacity=.5,text opacity=1]
\tikzstyle{dim3}  = [fill opacity=.7,text opacity=1]
\tikzstyle{rectc} = [tight,transform shape]
\tikzstyle{rect}  = [rectc,anchor=south west]

\newcommand{\leg}[1]{\addlegendentry{#1}}

\tikzset{every mark/.append style={solid}}
\pgfplotsset{
	grid=both, width=\columnwidth, try min ticks=5,
	every axis/.append style={font=\small},
	every axis plot/.append style={thick,mark=none,mark size=1.8,tension=0.18},
	legend cell align=left, legend style={fill opacity=0.8},
	xticklabel={\pgfmathprintnumber[assume math mode=true]{\tick}},
	yticklabel={\pgfmathprintnumber[assume math mode=true]{\tick}},
	nodes near coords math/.style={
		nodes near coords={\pgfmathprintnumber[assume math mode=true]{\pgfplotspointmeta}},
	},
}

\pgfplotsset{
	dash/.style={mark=o,dashed,opacity=0.6},
	dott/.style={mark=o,dotted,opacity=0.6},
	nolim/.style={enlargelimits=false},
	plain/.style={every axis plot/.append style={},nolim,grid=none},
}

\pgfdeclarelayer{bg4}
\pgfdeclarelayer{bg3}
\pgfdeclarelayer{bg2}
\pgfdeclarelayer{bg1}
\pgfdeclarelayer{fg1}
\pgfdeclarelayer{fg2}
\pgfdeclarelayer{fg3}
\pgfdeclarelayer{fg4}
\pgfsetlayers{bg4,bg3,bg2,bg1,main,fg1,fg2,fg3,fg4}

\pgfkeys{/tikz/.cd, aspect/.store in=\aspect, aspect=1}
\pgfkeys{/tikz/.cd, depth/.store in=\depth, depth=.5}
\pgfkeys{/tikz/.cd, stepx/.store in=\step, stepx=1}

\tikzstyle{geom} = [line join=bevel,aspect=1,depth=.5,z={(\depth*\aspect,\depth)}]
\tikzstyle{wire} = [geom,draw,thick]

\def\cx[#1,#2,#3]{#1}
\def\cy[#1,#2,#3]{#2}
\def\cz[#1,#2,#3]{#3}
\def\ex[#1,#2,#3]{#1,0,0}
\def\ey[#1,#2,#3]{0,#2,0}
\def\ez[#1,#2,#3]{0,0,#3}

\makeatletter
\renewcommand\paragraph{\@startsection{paragraph}{4}{\z@}{1ex}{-1em}{\normalfont\normalsize\bfseries}}
\makeatother

\usepackage{times}
\usepackage{epsfig}
\usepackage{graphicx}
\usepackage{amsmath}
\usepackage{amssymb}
\usepackage{xspace}

\usepackage{natbib}
\usepackage{enumitem}
\usepackage{multirow}
\usepackage{array,booktabs}
\usepackage{bbm}
\usepackage{colortbl}

\usepackage[pagebackref=true,breaklinks=true,colorlinks,bookmarks=false]{hyperref}

\usepackage[capitalize]{cleveref}
\crefname{section}{Sec.}{Secs.}
\Crefname{section}{Section}{Sections}
\Crefname{table}{Table}{Tables}
\crefname{table}{Tab.}{Tabs.}

\journal{Computer Vision and Image Understanding}
\setcitestyle{numbers}

\begin{document}

\thispagestyle{empty}

\newcommand{\head}[1]{{\smallskip\noindent\textbf{#1}}}
\newcommand{\alert}[1]{{\color{red}{#1}}}
\newcommand{\sm}{\scriptsize}
\newcommand{\eq}[1]{(\ref{eq:#1})}

\newcommand{\Th}[1]{\textsc{#1}}
\newcommand{\mr}[2]{\multirow{#1}{*}{#2}}
\newcommand{\mc}[2]{\multicolumn{#1}{c}{#2}}
\newcommand{\tb}[1]{\textbf{#1}}
\newcommand{\ch}{\checkmark}

\newcommand{\red}[1]{{\textcolor{red}{#1}}}
\newcommand{\blue}[1]{{\textcolor{blue}{#1}}}
\newcommand{\green}[1]{{\textcolor{green}{#1}}}
\newcommand{\gray}[1]{{\textcolor{gray}{#1}}}

\newcommand{\citeme}[1]{\red{[XX]}}
\newcommand{\refme}[1]{\red{(XX)}}

\newcommand{\fig}[2][1]{\includegraphics[width=#1\linewidth]{fig/#2}}
\newcommand{\figh}[2][1]{\includegraphics[height=#1\linewidth]{fig/#2}}
\newcommand{\figa}[2][1]{\includegraphics[width=#1]{fig/#2}}
\newcommand{\figah}[2][1]{\includegraphics[height=#1]{fig/#2}}

\newcommand{\tran}{^\top}
\newcommand{\mtran}{^{-\top}}
\newcommand{\zcol}{\mathbf{0}}
\newcommand{\zrow}{\zcol\tran}

\newcommand{\ind}{\mathbbm{1}}
\newcommand{\expect}{\mathbb{E}}
\newcommand{\nat}{\mathbb{N}}
\newcommand{\zahl}{\mathbb{Z}}
\newcommand{\real}{\mathbb{R}}
\newcommand{\proj}{\mathbb{P}}
\newcommand{\prob}{\operatorname{P}}
\newcommand{\normal}{\mathcal{N}}

\newcommand{\mif}{\textrm{if}\ }
\newcommand{\other}{\textrm{otherwise}}
\newcommand{\minimize}{\textrm{minimize}\ }
\newcommand{\maximize}{\textrm{maximize}\ }
\newcommand{\st}{\textrm{subject\ to}\ }

\newcommand{\id}{\operatorname{id}}
\newcommand{\const}{\operatorname{const}}
\newcommand{\sgn}{\operatorname{sgn}}
\newcommand{\var}{\operatorname{Var}}
\newcommand{\mean}{\operatorname{mean}}
\newcommand{\trace}{\operatorname{tr}}
\newcommand{\diag}{\operatorname{diag}}
\newcommand{\vect}{\operatorname{vec}}
\newcommand{\cov}{\operatorname{cov}}
\newcommand{\sign}{\operatorname{sign}}
\newcommand{\prj}{\operatorname{proj}}

\newcommand{\softmax}{\operatorname{softmax}}
\newcommand{\clip}{\operatorname{clip}}

\newcommand{\defn}{\mathrel{:=}}
\newcommand{\peq}{\mathrel{+\!=}}
\newcommand{\meq}{\mathrel{-\!=}}

\newcommand{\paren}[1]{\left({#1}\right)}
\newcommand{\mat}[1]{\left[{#1}\right]}
\newcommand{\set}[1]{\left\{{#1}\right\}}
\newcommand{\floor}[1]{\left\lfloor{#1}\right\rfloor}
\newcommand{\ceil}[1]{\left\lceil{#1}\right\rceil}
\newcommand{\inner}[1]{\left\langle{#1}\right\rangle}
\newcommand{\norm}[1]{\left\|{#1}\right\|}
\newcommand{\abs}[1]{\left|{#1}\right|}
\newcommand{\frob}[1]{\norm{#1}_F}
\newcommand{\card}[1]{\left|{#1}\right|\xspace}

\newcommand{\diff}{\mathrm{d}}
\newcommand{\der}[3][]{\frac{\diff^{#1}#2}{\diff#3^{#1}}}
\newcommand{\ider}[3][]{\diff^{#1}#2/\diff#3^{#1}}
\newcommand{\pder}[3][]{\frac{\partial^{#1}{#2}}{\partial{{#3}^{#1}}}}
\newcommand{\ipder}[3][]{\partial^{#1}{#2}/\partial{#3^{#1}}}
\newcommand{\dder}[3]{\frac{\partial^2{#1}}{\partial{#2}\partial{#3}}}

\newcommand{\wb}[1]{\overline{#1}}
\newcommand{\wt}[1]{\widetilde{#1}}

\def\xssp{\hspace{1pt}}
\def\ssp{\hspace{3pt}}
\def\msp{\hspace{5pt}}
\def\lsp{\hspace{12pt}}

\newcommand{\cA}{\mathcal{A}}
\newcommand{\cB}{\mathcal{B}}
\newcommand{\cC}{\mathcal{C}}
\newcommand{\cD}{\mathcal{D}}
\newcommand{\cE}{\mathcal{E}}
\newcommand{\cF}{\mathcal{F}}
\newcommand{\cG}{\mathcal{G}}
\newcommand{\cH}{\mathcal{H}}
\newcommand{\cI}{\mathcal{I}}
\newcommand{\cJ}{\mathcal{J}}
\newcommand{\cK}{\mathcal{K}}
\newcommand{\cL}{\mathcal{L}}
\newcommand{\cM}{\mathcal{M}}
\newcommand{\cN}{\mathcal{N}}
\newcommand{\cO}{\mathcal{O}}
\newcommand{\cP}{\mathcal{P}}
\newcommand{\cQ}{\mathcal{Q}}
\newcommand{\cR}{\mathcal{R}}
\newcommand{\cS}{\mathcal{S}}
\newcommand{\cT}{\mathcal{T}}
\newcommand{\cU}{\mathcal{U}}
\newcommand{\cV}{\mathcal{V}}
\newcommand{\cW}{\mathcal{W}}
\newcommand{\cX}{\mathcal{X}}
\newcommand{\cY}{\mathcal{Y}}
\newcommand{\cZ}{\mathcal{Z}}

\newcommand{\vA}{\mathbf{A}}
\newcommand{\vB}{\mathbf{B}}
\newcommand{\vC}{\mathbf{C}}
\newcommand{\vD}{\mathbf{D}}
\newcommand{\vE}{\mathbf{E}}
\newcommand{\vF}{\mathbf{F}}
\newcommand{\vG}{\mathbf{G}}
\newcommand{\vH}{\mathbf{H}}
\newcommand{\vI}{\mathbf{I}}
\newcommand{\vJ}{\mathbf{J}}
\newcommand{\vK}{\mathbf{K}}
\newcommand{\vL}{\mathbf{L}}
\newcommand{\vM}{\mathbf{M}}
\newcommand{\vN}{\mathbf{N}}
\newcommand{\vO}{\mathbf{O}}
\newcommand{\vP}{\mathbf{P}}
\newcommand{\vQ}{\mathbf{Q}}
\newcommand{\vR}{\mathbf{R}}
\newcommand{\vS}{\mathbf{S}}
\newcommand{\vT}{\mathbf{T}}
\newcommand{\vU}{\mathbf{U}}
\newcommand{\vV}{\mathbf{V}}
\newcommand{\vW}{\mathbf{W}}
\newcommand{\vX}{\mathbf{X}}
\newcommand{\vY}{\mathbf{Y}}
\newcommand{\vZ}{\mathbf{Z}}

\newcommand{\va}{\mathbf{a}}
\newcommand{\vb}{\mathbf{b}}
\newcommand{\vc}{\mathbf{c}}
\newcommand{\vd}{\mathbf{d}}
\newcommand{\ve}{\mathbf{e}}
\newcommand{\vf}{\mathbf{f}}
\newcommand{\vg}{\mathbf{g}}
\newcommand{\vh}{\mathbf{h}}
\newcommand{\vi}{\mathbf{i}}
\newcommand{\vj}{\mathbf{j}}
\newcommand{\vk}{\mathbf{k}}
\newcommand{\vl}{\mathbf{l}}
\newcommand{\vm}{\mathbf{m}}
\newcommand{\vn}{\mathbf{n}}
\newcommand{\vo}{\mathbf{o}}
\newcommand{\vp}{\mathbf{p}}
\newcommand{\vq}{\mathbf{q}}
\newcommand{\vr}{\mathbf{r}}
\newcommand{\Vs}{\mathbf{s}}
\newcommand{\vt}{\mathbf{t}}
\newcommand{\vu}{\mathbf{u}}
\newcommand{\vv}{\mathbf{v}}
\newcommand{\vw}{\mathbf{w}}
\newcommand{\vx}{\mathbf{x}}
\newcommand{\vy}{\mathbf{y}}
\newcommand{\vz}{\mathbf{z}}

\newcommand{\vone}{\mathbf{1}}
\newcommand{\vzero}{\mathbf{0}}

\newcommand{\valpha}{{\boldsymbol{\alpha}}}
\newcommand{\vbeta}{{\boldsymbol{\beta}}}
\newcommand{\vgamma}{{\boldsymbol{\gamma}}}
\newcommand{\vdelta}{{\boldsymbol{\delta}}}
\newcommand{\vepsilon}{{\boldsymbol{\epsilon}}}
\newcommand{\vzeta}{{\boldsymbol{\zeta}}}
\newcommand{\veta}{{\boldsymbol{\eta}}}
\newcommand{\vtheta}{{\boldsymbol{\theta}}}
\newcommand{\viota}{{\boldsymbol{\iota}}}
\newcommand{\vkappa}{{\boldsymbol{\kappa}}}
\newcommand{\vlambda}{{\boldsymbol{\lambda}}}
\newcommand{\vmu}{{\boldsymbol{\mu}}}
\newcommand{\vnu}{{\boldsymbol{\nu}}}
\newcommand{\vxi}{{\boldsymbol{\xi}}}
\newcommand{\vomikron}{{\boldsymbol{\omikron}}}
\newcommand{\vpi}{{\boldsymbol{\pi}}}
\newcommand{\vrho}{{\boldsymbol{\rho}}}
\newcommand{\vsigma}{{\boldsymbol{\sigma}}}
\newcommand{\vtau}{{\boldsymbol{\tau}}}
\newcommand{\vupsilon}{{\boldsymbol{\upsilon}}}
\newcommand{\vphi}{{\boldsymbol{\phi}}}
\newcommand{\vchi}{{\boldsymbol{\chi}}}
\newcommand{\vpsi}{{\boldsymbol{\psi}}}
\newcommand{\vomega}{{\boldsymbol{\omega}}}

\newcommand{\rLambda}{\mathrm{\Lambda}}
\newcommand{\rSigma}{\mathrm{\Sigma}}

\newcommand{\vLambda}{\bm{\rLambda}}
\newcommand{\vSigma}{\bm{\rSigma}}

\makeatletter
\newcommand*\bdot{\mathpalette\bdot@{.7}}
\newcommand*\bdot@[2]{\mathbin{\vcenter{\hbox{\scalebox{#2}{$\m@th#1\bullet$}}}}}
\makeatother

\makeatletter
\DeclareRobustCommand\onedot{\futurelet\@let@token\@onedot}
\def\@onedot{\ifx\@let@token.\else.\null\fi\xspace}

\def\eg{\emph{e.g}\onedot} \def\Eg{\emph{E.g}\onedot}
\def\ie{\emph{i.e}\onedot} \def\Ie{\emph{I.e}\onedot}
\def\cf{\emph{cf}\onedot} \def\Cf{\emph{Cf}\onedot}
\def\etc{\emph{etc}\onedot} \def\vs{\emph{vs}\onedot}
\def\wrt{w.r.t\onedot} \def\dof{d.o.f\onedot} \def\aka{a.k.a\onedot}
\def\etal{\emph{et al}\onedot}
\makeatother

\newcommand{\relu}{\operatorname{relu}}
\newcommand{\gap}{\operatorname{GAP}}
\newcommand{\up}{\operatorname{up}}

\newcommand{\cam}{\textrm{CAM}}
\newcommand{\gcam}{\textrm{Grad-CAM}}
\newcommand{\scam}{\textrm{Score-CAM}}

\newcommand{\Fdef}{Mask\xspace}
\newcommand{\Fref}{Diff\xspace}
\newcommand{\MIOFref}{IODiff\xspace}
\newcommand{\MIODref}{IOMask\xspace}

\newcommand{\AG}{\operatorname{AG}}
\newcommand{\AGf}{Average Gain\xspace}
\newcommand{\Agf}{Average gain\xspace}
\newcommand{\agf}{average gain\xspace}

\newcommand{\AC}{\operatorname{AC}}
\newcommand{\ACf}{Average Contract\xspace}

\newcommand{\AD}{\operatorname{AD}}
\newcommand{\I}{\operatorname{I}}
\newcommand{\D}{\operatorname{D}}
\newcommand{\AI}{\operatorname{AI}}
\newcommand{\OM}{\operatorname{OM}}
\newcommand{\LE}{\operatorname{LE}}
\newcommand{\Fo}{\operatorname{F1}}
\newcommand{\prc}{\operatorname{precision}}
\newcommand{\rec}{\operatorname{recall}}
\newcommand{\BA}{\operatorname{BoxAcc}}
\newcommand{\spg}{\operatorname{SP}}
\newcommand{\epg}{\operatorname{EP}}
\newcommand{\SM}{\operatorname{SM}}
\newcommand{\iou}{\operatorname{IoU}}

\newcommand{\ronan}[1]{#1}
\newcommand{\iavr}[1]{#1}
\newcommand{\hw}[1]{#1}
\newcommand{\stephane}[1]{#1}
\newcommand{\redred}[1]{#1}

\begin{frontmatter}

\title{Opti-CAM: Optimizing saliency maps for interpretability}

\author[1]{Hanwei \snm{Zhang}\corref{cor1}} 
\cortext[cor1]{Corresponding author: 
  Tel.: +3306313058;  }
\ead{zhanghanwei0912@gmail.com}
\author[1]{Felipe \snm{Torres}}
\author[1]{Ronan \snm{Sicre}}
\author[2]{Yannis \snm{Avrithis}}
\author[1]{Stephane \snm{Ayache}}

\address[1]{Centrale Marseille, Aix Marseille Univ, CNRS, LIS, Marseille, France}
\address[2]{Institute of Advanced Research on Artificial Intelligence (IARAI)}

\received{1 May 2013}
\finalform{10 May 2013}
\accepted{13 May 2013}
\availableonline{15 May 2013}
\communicated{H. Zhang}


\begin{abstract}
Methods based on \emph{class activation maps} (CAM) provide a simple mechanism to interpret predictions of convolutional neural networks by using linear combinations of feature maps as saliency maps. By contrast, masking-based methods optimize a saliency map directly in the image space or learn it by training another network on additional data.

In this work we introduce Opti-CAM, combining ideas from CAM-based and masking-based approaches. Our saliency map is a linear combination of feature maps, where weights are optimized per image such that the logit of the masked image for a given class is maximized. We also fix a fundamental flaw in two of the most common evaluation metrics of attribution methods. On several datasets, Opti-CAM largely outperforms other CAM-based approaches according to the most relevant classification metrics. We provide empirical evidence supporting that localization and classifier interpretability are not necessarily aligned.
\end{abstract}

\begin{keyword}
Interpretability; Explainable AI; Saliency map; Class activation maps; Computer vision; 
\MSC 41A05\sep 41A10\sep 65D05\sep 65D17
\KWD Keyword1\sep Keyword2\sep Keyword3

\end{keyword}

\end{frontmatter}

\section{Introduction}
\label{sec:intro}

The success of \emph{deep neural networks} (DNN) and their increasing penetration into most sectors of human activity has led to growing interest in understanding how these models make their predictions. Unlike shallow methods, DNN have a high complexity and it is not possible to directly explain their inference process in a human understandable manner. This challenge has opened up an entire research field~\citep{guidotti2018survey, montavon2018methods, samek2021explaining, bodria2021benchmarking, li2021interpretable}.

In this work, we are interested in the interpretability of deep neural networks through the generation of \emph{saliency maps}, highlighting regions of an image that are responsible for the prediction. This originates in \emph{gradient-based} methods \citep{simonyan2013deep, yosinski2015understanding}, including variants of backpropagation \citep{zeiler2014visualizing, springenberg2014striving, bach2015pixel}. CAM~\citep{zhou2016learning} introduced class-specific linear combinations of feature maps, and led to several alternative weighting schemes \citep{ramaswamy2020ablation, wang2020score, muhammad2020eigen}, including the use of gradients \citep{selvaraju2017grad, chattopadhay2018grad}. On the other hand, \emph{occlusion-} or \emph{masking-based} methods \citep{dabkowski2017real, fong2017interpretable, fong2019understanding, schulz2020restricting} remove regions in the image space while improving classification performance.

Score-CAM~\cite{wang2020score} uses each feature map as a mask and defines a corresponding weight based on the resulting increase of class score; hence, it is both CAM-based and masking-based but does not use gradients. It resembles the numerical gradient approximation, in that it needs \emph{one forward pass per weight}. Instead, the analytical approach would be to use a linear combination of feature maps as a mask, express the class score as a function of the weights and measure the gradient analytically, in a \emph{single backward pass}. Then, \emph{why not use gradient descent to maximize the class score?} The optimal mask should highlight regions for which the network is most confident.

\emph{Masking-based} methods, such as extremal perturbations~\citep{fong2019understanding} or IBA~\citep{schulz2020restricting}, do use gradient descent to maximize the class score. The mask is now a variable in the input or feature space and the class score is expressed as a function of the mask directly. Because the variable being optimized is a high-dimensional image or tensor, additional constraints or regularizers are needed to control \eg the smoothness and the salient area. This translates to more hyperparameters or more expensive optimization.

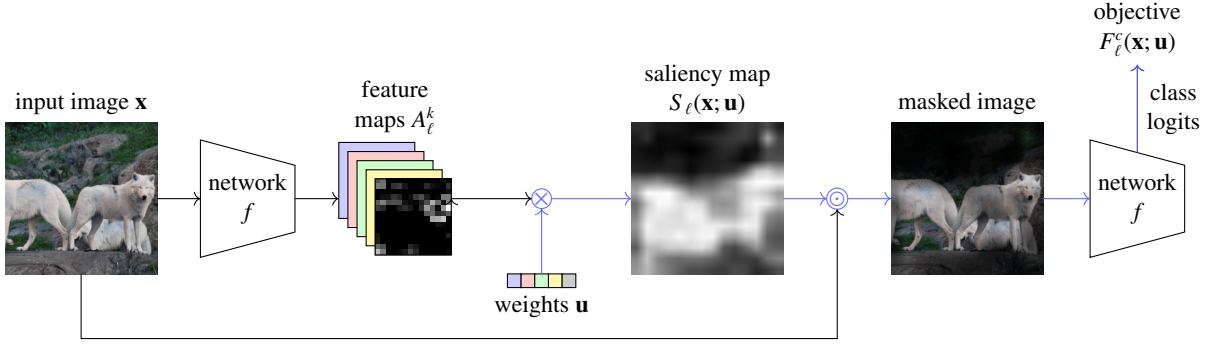
\begin{figure*}[t]
\centering
\begin{tikzpicture}[
	scale=.3,
	font={\small},
	node distance=.5,
	label distance=3pt,
	net/.style={draw,trapezium,trapezium angle=75,inner sep=3pt},
	enc/.style={net,shape border rotate=270},
	txt/.style={inner sep=3pt},
	frame/.style={draw,minimum size=1cm},
	feat/.style={frame},
	sq/.style={minimum size=.18cm},
	elem/.style={draw,sq},
	vec/.style={draw,minimum width=.8cm,minimum height=.15cm},
	var/.style={blue!60},
	B/.style={fill=blue!20},
	R/.style={fill=red!20},
	G/.style={fill=green!20},
	Y/.style={fill=yellow!40},
	P/.style={fill=black!20},
]
\matrix[
	tight,row sep=0,column sep=16,
	cells={scale=.3,},
] {
	\&\&\&\&\&\&\&
	\node[txt] (loss) {objective \\ $F^c_\ell(\vx; \vu)$}; \\
	\node[label=90:{input image $\vx$}] (in) {\figah[2cm]{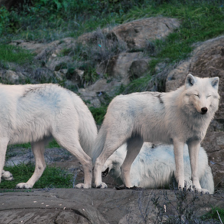}}; \&
	\node[enc] (net) {network \\ $f$}; \&
	\foreach \s/\c in {-2/B,-1/R,0/G,1/Y,2/P}
		{\node[feat,\c] (feat\s) at ($.4*(\s,-\s)$) {};}
	\node        at (feat2) {\figah[1cm]{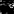}};
	\node[frame] at (feat2) {};
	\coordinate[label=90:{feature \\ maps $A^k_\ell$}]
	            (feat-north) at (feat-2.north -| feat0.north);
	\coordinate (feat-west)  at (feat-2.west  |- feat0.west);
	\coordinate (feat-east)  at (feat2.east   |- feat0.east);
	\&
	\node[var,op] (cam) {$\times$};
	\foreach \s/\c in {-2/B,-1/R,0/G,1/Y,2/P}
		{\node[elem,\c] (elem\s) at ($.6*(\s,-6)$) {};}
	\node[sq,label=-90:{weights $\vu$}] (weight) at (elem0) {};
	\&
	\node[var,label=90:{saliency map \\ $S_\ell(\vx; \vu)$}] (sal) {\figah[2cm]{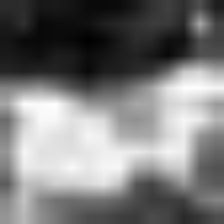}}; \&
	\node[var,op] (mask) {$\odot$}; \&
	\node[label=90:{masked image}] (masked) {\figah[2cm]{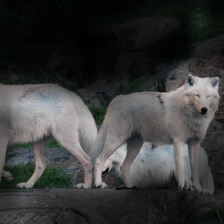}}; \&
	\node[enc] (net2) {network \\ $f$}; \\[8]
	\&\&\&
	\coordinate (mid); \\
};

\draw[->]
	(in) edge (net)
	(net) edge (feat-west)
	(feat-east) edge (cam)
	(net2) edge node[pos=.5,right] {class \\ logits} (loss)
	;

\draw[var,->]
	(weight) edge (cam)
	(cam) edge (sal)
	(sal) edge (mask)
	(mask) edge (masked)
	(masked) edge (net2)
    (net2) edge (loss)
	;

\draw[->]
	(in) |- (mid)
	(mid) -| (mask)
	;

\end{tikzpicture}

\caption{Overview of Opti-CAM. We are given an input image $\vx$, a fixed network $f$, a target layer $\ell$ and a class of interest $c$. We extract the feature maps from layer $\ell$ and obtain a saliency map $S_\ell(\vx; \vu)$ by forming a convex combination of the feature maps ($\times$) with weights determined by a variable vector $\vu$~\eq{v-sal}. After upsampling and normalizing, we element-wise multiply ($\odot$) the saliency map with the input image to form a ``masked'' version of the input, which is fed to $f$. The objective function $F^c_\ell(\vx; \vu)$ measures the logit of class $c$ for the masked image~\eq{obj}. We find the value of $\vu^*$ that maximizes this logit by optimizing along the path highlighted in blue~\eq{opt}, as well as the corresponding optimal saliency map $S_\ell(\vx; \vu^*)$~\eq{o-sal}.}
\label{fig:idea}
\vspace{-0.4cm}
\end{figure*}

Motivated by the above, we introduce Opti-CAM, illustrated in \autoref{fig:idea}. We form a linear combination of feature maps, where the weights are a variable. Treating it as a saliency map, we form a masked version of the input image that is fed again to the network. Then, the logit of a given class for the masked version of the input is maximized to obtain the optimal weights. Thus, Opti-CAM can be seen as an analytical counterpart of Score-CAM that is optimized iteratively, or as a masking-based method where the mask to be optimized lies in the linear span of the feature maps, like CAM-based methods.

The evaluation metrics most relevant to using a saliency map as a mask are \emph{average drop} ($\AD$) and \emph{average increase} ($\AI$)~\cite{chattopadhay2018grad}. The problem is that the two metrics are not defined in a symmetric way. As a result, there exists a trivial attribution method called Fake-CAM~\cite{poppi2021revisiting} that outperforms the state of the art in both metrics. To address this, we introduce the symmetric counterpart of $\AD$, which we call \emph{\agf} ($\AG$), to be paired with $\AD$ as a replacement of $\AI$. As expected, Fake-CAM fails $\AG$.

In summary, we make the following contributions:
\begin{enumerate}[itemsep=2pt, parsep=0pt, topsep=3pt]
	\item We introduce Opti-CAM, a simple model for saliency map generation that combines ideas from CAM-based and masking-based approaches. \redred{Opti-CAM does not need any extra data, network or training.}
	\item Compared with gradient-free methods~\citep{wang2020score,petsiuk2018rise,ramaswamy2020ablation}, it finds the optimal feature map weights and is on par or faster, assuming that the number of iterations is less than the number of channels.
	\item We introduce a new evaluation metric, \emph{\agf} ($\AG$), to be paired with \emph{average drop} ($\AD$) as a replacement of \emph{average increase} ($\AI$)~\cite{chattopadhay2018grad}.
	\item On several datasets,	we improve the state of the art by a large margin, \redred{reaching near-perfect performance} according to the most relevant classification metrics.
	\item We shed more light into how a classifier may exploit background context.
\end{enumerate}

\section{Related Work}

A large number of works study \emph{explainability}, \emph{interpretability} or \emph{attribution} of machine learning models, especially DNN~\citep{guidotti2018survey, montavon2018methods, samek2021explaining, bodria2021benchmarking, li2021interpretable}. These works can be categorized into \emph{transparency} and \emph{post-hoc interpretability}~\citep{lipton2018mythos, guidotti2018survey}. The former addresses how to design an internally understandable model. Here we are interested in the latter, which treats the studied network as a black box and interprets its inner processing~\citep{ribeiro2016should, lundberg2017unified, fong2017interpretable, elliott2021explaining, selvaraju2017grad, petsiuk2018rise}. Among post-hoc methods, LIME~\citep{ribeiro2016should} and SHAP~\citep{lundberg2017unified} are well-known model-agnostic methods that rate feature importance. More specifically, we are interested in the generation of \emph{saliency maps}. These methods are mostly based on gradients, CAM~\citep{zhou2016learning}, occlusion, or a combination.


\paragraph{Gradient-based methods}

Gradient-based methods~\citep{adebayo2018local,springenberg2014striving,baehrens2010explain} use the gradient of a target class score with respect to the input to measure the effect of different image regions on the prediction. In~\citep{simonyan2013deep}, the gradient is directly treated as a saliency map. Inspired by DeconvNet~\citep{zeiler2014visualizing}, \emph{guided backpropagation}~\citep{springenberg2014striving} improves the explanation by setting negative gradients to zero using ReLU units. Other methods~\citep{shrikumar2017learning, zhang2018top, bastings2020elephant} are inspired by Layer-wise Relevance Propagation (LRP)~\citep{bach2015pixel}. SmoothGrad~\citep{smilkov2017smoothgrad} and \emph{integrated gradients}~\cite{sundararajan2017axiomatic} accumulate gradients into saliency maps, while NormGrad~\citep{rebuffi2020there} attempts to unify gradient-based methods. A different approach is to use adversarial attacks~\citep{elliott2021explaining, jalwana2020attack}. Several of these methods do not satisfy the fundamental property of implementation invariance~\cite{sundararajan2017axiomatic}.


\paragraph{CAM-based methods}

\emph{Class activation maps} (CAM)~\citep{zhou2016learning} is a visualization method that highlights the image regions most relevant to a target class by a linear combination of feature maps. A number of variants use different definitions of weights. Many rely on gradients, including GradCAM~\citep{selvaraju2017grad}, GradCAM++~\citep{chattopadhay2018grad}, XGradCAM~\citep{fu2020axiom} and LayerCAM~\citep{jiang2021layercam}. Gradient-free methods, including Ablation-CAM~\citep{ramaswamy2020ablation}, Score-CAM~\cite{wang2020score} and SS-CAM~\citep{wang2020ss}, rather measure the effect on the target class score of each feature map acting as a mask on the input. We inherit the idea of masking but for linear combinations of feature maps and we iteratively optimize the coefficients by analytical gradient computation. Our method is thus faster when the number of iterations is less than the number of channels.


\paragraph{Occlusion (masking)-based methods}

These methods use a number of candidate masks, measure their effect on the prediction, then combine them in a single saliency map. RISE~\citep{petsiuk2018rise} randomly masks input images and uses the class score as a weight to define a linear combination. \emph{Meaningful perturbations} \citep{fong2017interpretable} and \emph{extremal perturbations}~\citep{fong2019understanding} directly optimize the mask in the image space by using gradients. They require a large number of parameters as well as regularizers, \eg for smoothness. \emph{Information bottleneck attribution} (IBA)~\citep{schulz2020restricting} optimizes the mask in the feature space as a tensor instead. Score-CAM~\cite{wang2020score} is also an occlusion-based method, using individual feature maps as candidate masks. The same holds for our Opti-CAM, but for candidate masks constrained in the linear span of the feature maps. Compared with~\citep{fong2019understanding,schulz2020restricting}, we have fewer parameters and do not require a regularizer.


\paragraph{Learning-based methods}

While occlusion-based methods compute or optimize a mask for a particular image at inference, learning-based methods use an additional network or branch and they train it on extra data and image-level labels to predict a saliency map given an input image. This includes for example generators \citep{chang2018explaining} or auto-encoders \citep{dabkowski2017real, phang2020investigating, zolna2020classifier}. This approach may be compared with weakly-supervised object detection~\citep{bilen2016weakly}, segmentation~\citep{KoLa16} or instance segmentation~\citep{AhCK19}. IBA~\citep{schulz2020restricting} includes a learning-based approach in the feature space. Apart from requiring extra data, it is not satisfying in the sense that the learned decoder would need to be explained too. Our method does not need any extra data, network, or training.


\paragraph{Evaluation of attribution methods}

Evaluating saliency maps is challenging because no ground truth attributions exist. \emph{Average drop} ($\AD$) and \emph{average increase} ($\AI$), also known as increase in confidence~\cite{chattopadhay2018grad} are well-established metrics. They consider the effect on the predicted class probabilities by masking the input image with the saliency map. There is a fundamental flaw in using $\AD$, $\AI$ as a pair of metrics, which we fix by replacing $\AI$ by a new metric, \emph{average gain} ($\AG$).

\emph{Insertion} (I) and \emph{deletion} (D) sequentially insert or delete pixels by decreasing order of saliency and observe the effect on the prediction. The resulting images are out-of-distribution (OOD)~\cite{gomez2022metrics} and the metrics favor small and compact regions. Localization metrics measure how the saliency maps are aligned with object bounding boxes, which ignores the importance of background context~\cite{shetty2019not, rao2022towards}. We demonstrate that localization and attribution are not well-aligned as tasks.

\section{Opti-CAM}
\label{sec:opticam}

\subsection{Preliminaries}
\label{sec:prelim}

\paragraph{Notation}
\label{sec:notation}

Consider a classifier network $f: \cX \to \real^C$ that maps an input image $\vx \in \cX$ to a logit vector $\vy = f(\vx) \in \real^C$, where $\cX$ is the image space and $C$ is the number of classes. We denote by $y_c = f(\vx)_c$ the predicted logit and by $p_c = \softmax(\vy)_c \defn e^{y_c} / \sum_j e^{y_j}$ the predicted probability for class $c$. For layer $\ell$ with $K_\ell$ channels, we denote by $A^k_\ell = f^k_\ell(\vx) \in \real^{h_\ell \times w_\ell}$ the feature map for channel $k \in \{1,\dots,K_\ell\}$, with spatial resolution $h_\ell \times w_\ell$. Because of $\relu$ non-linearities, we assume that feature maps are non-negative. Similarly, we denote by $S_\ell \in \real^{h_\ell \times w_\ell}$ a 2D saliency map.


\paragraph{Background: CAM-based saliency maps}
\label{sec:back}

Given a layer $\ell$ and a class of interest $c$, we consider saliency maps given by the general formula
\begin{equation}
	S^c_\ell(\vx) \defn h \left( \sum_k w^c_k A^k_\ell \right),
\label{eq:sal}
\end{equation}
where $w^c_k$ are weights defining a linear combination over channels and $h$ is an activation function. CAM~\citep{zhou2016learning} is defined for the last layer $L$ only with $h$ being the identity mapping and $w^c_k$ being the classifier weight connecting the $k$-th channel with class $c$. Grad-CAM~\citep{selvaraju2017grad} is defined for any layer $\ell$ with $h = \relu$ and weights
\begin{equation}
	w^c_k \defn \gap \left( \pder{y_c}{A^k_\ell} \right),
\label{eq:gcam}
\end{equation}
where $\gap$ is global average pooling.
The motivation for $\relu$ is that we are only interested in features that have a positive effect on the class of interest, \ie pixels whose intensity should be increased in order to increase $y_c$.

Score-CAM~\cite{wang2020score} is also defined for any layer $\ell$ with $h = \relu$ and weights $w^c_k \defn \softmax(\vu^c)_k$.  Softmax normalization considers positive channel contributions only and attends to few feature maps.
Here, vector $\vu^c \in \real^{K_\ell}$ measures the increase in confidence for class $c$ that compares a known baseline image $\vx_b$ with the input image $\vx$ masked according to feature map $A^k_\ell$, for all channels $k$:
\begin{equation}
	u^c_k \defn f(\vx \odot n(\up(A^k_\ell)))_c - f(\vx_b)_c,
\label{eq:s-cam}
\end{equation}
where $\odot$ is the Hadamard product. For this to work, the feature map $A^k_\ell$ is adapted to $\vx$ first: $\up$ denotes upsampling to the spatial resolution of $\vx$ and
\begin{equation}
	n(A) \defn \frac{A - \min A}{\max A - \min A}
\label{eq:norm}
\end{equation}
\redred{is a normalization of matrix $A$ into $[0,1]$.} While Score-CAM does not need gradients, it requires as many forward passes through the network as the number of channels in the chosen layer, which is computationally expensive.


\paragraph{Motivation}
\label{sec:motiv}

\iavr{Score-CAM considers each feature map as a mask in isolation. How about linear combinations?} Given a vector $\vw \in \real^{K_\ell}$ with $w_k$ its $k$-th element, let
\begin{equation}
	F(\vw) \defn f \left( \vx \odot n \left( \up \left(
		\displaystyle\sum_k w_k A^k_\ell
	\right) \right) \right)_c.
\label{eq:s-obj}
\end{equation}
\ronan{If we assume that $\vx_b = \vzero$ in~\eq{s-cam} and define $n(\vzero) \defn \vzero$ in~\eq{norm}, then we can rewrite the right-hand side of~\eq{s-cam} as
\begin{equation}
	\frac{F(\vw_0 + \delta \ve_k) - F(\vw_0)}{\delta},
\label{eq:s-cam2}
\end{equation}
where $\vw_0 = \vzero$, $\delta = 1$ and $\ve_k$ is the $k$-th standard basis vector of $\real^{K_\ell}$. This resembles the numerical approximation of the derivative $\pder{F}{w_k}(\vw_0)$, except that $\delta$ is not small as usual. One could compute derivatives efficiently by standard backpropagation instead. It is then possible to iteratively optimize $F$ with respect to $\vw$, starting at any $\vw_0$.}

\iavr{As an alternative, consider masking-based methods relying on optimization in the input space, like \emph{meaningful perturbations} (MP)~\cite{fong2017interpretable} or \emph{extremal perturbations}~\citep{fong2019understanding}. In general, optimization takes the form
\begin{equation}
	S^c(\vx) \defn \arg\max_{\vm \in \cM} f(\vx \odot n(\up(\vm)))_c + \lambda R(\vm).
\label{eq:mask}
\end{equation}
Here, a mask $\vm$ is directly optimized and does not rely on feature maps, hence the saliency map $S^x(\vx)$ is not connected to any layer $\ell$. The mask is at the same or lower resolution than the input image. In the latter case, upsampling is still necessary.

In this approach, one indeed computes derivatives by backpropagation and indeed iteratively optimizes $\vm$. However, because $\vm$ is high-dimensional, there are constraints expressed by $\vm \in \cM$, \eg $\vm$ has a certain norm, and regularizers like $R(\vm)$, \eg $\vm$ is smooth in a certain way. This makes optimization harder or more expensive and introduces more hyperparameters like $\lambda$. One could simply constrain $\vm$ to lie in the linear span of $\{A_\ell^k\}_{k=1}^{K_\ell}$ instead, like all CAM-based methods.}


\subsection{Method}
\label{sec:method}

\paragraph{Saliency maps}

As motivated by \autoref{sec:motiv}, we obtain a saliency map as a convex combination of feature maps by optimizing a given objective function with respect to the weights.
In particular, following~\citep{wang2020score}, we use channel weights $w_k \defn \softmax(\vu)_k$, where $\vu \in \real^{K_\ell}$ is a variable.
We then consider saliency map $S_\ell$ in layer $\ell$ as a function of both the input image $\vx$ and variable $\vu$:
\begin{equation}
    S_\ell(\vx; \vu) \defn \sum_k \softmax(\vu)_k A^k_\ell.
\label{eq:v-sal}
\end{equation}
Comparing with~\eq{sal}, $h$ is the identity mapping, because feature maps are non-negative and weights are positive.


\paragraph{Optimization}

Now, given a layer $\ell$ and a class of interest $c$, we find the vector $\vu^*$ that maximizes the classifier confidence for class $c$, when the input image $\vx$ is masked according to saliency map $S_\ell(\vx; \vu^*)$:
\begin{equation}
	\vu^* \defn \arg\max_{\vu} F^c_\ell(\vx; \vu),
\label{eq:opt}
\end{equation}
where we define the objective function
\begin{equation}
	F^c_\ell(\vx; \vu) \defn g_c(f(\vx \odot n(\up(S_\ell(\vx; \vu))))).
\label{eq:obj}
\end{equation}
Here, the saliency map $S_\ell(\vx; \vu)$ is adapted to $\vx$ exactly as in~\eq{s-cam} in terms of resolution and normalization. For \emph{normalization function} $n$, the default is~\eq{norm}. The \emph{selector function} $g_c$ operates on the logit vector $\vy$; the default is to select the logit of class $c$, \ie $g_c(\vy) \defn y_c$. Other choices, including the definition of $F^c_\ell$ itself, are investigated in \autoref{sec:ablation} \redred{and in the supplementary material.}


\paragraph{Opti-CAM}

Putting everything together, we define
\begin{equation}
	S^c_\ell(\vx) \defn S_\ell(\vx; \vu^*) = S_\ell(\vx; \arg\max_{\vu} F^c_\ell(\vx; \vu)),
\label{eq:o-sal}
\end{equation}
where $S_\ell$ and $F^c_\ell$ are defined by~\eq{v-sal} and~\eq{obj} respectively. The objective function $F^c_\ell$~\eq{obj} depends on variable $\vu$ through $S_\ell$~\eq{v-sal}, where the feature maps $A^k_\ell = f^k_\ell(\vx)$ are fixed. Then,~\eq{obj} involves masking and a forward pass through the network $f$, which is also fixed.

\autoref{fig:idea} is an abstract illustration of our method, \iavr{called Opti-CAM}, without details like upsampling and normalization~\eq{obj}. Optimization takes place along the highlighted path from variable $\vu$ to objective function $F^c_\ell$. The saliency map is real-valued and the entire objective function is differentiable in $\vu$. We use Adam optimizer~\citep{kingma2014adam} to solve the optimization problem~\eq{opt}.


\paragraph{Discussion}

By maximizing~\eq{obj}, the saliency map focuses on the regions contributing to class $c$, while masked regions contribute less. This way, the influence of background in the average pooling process is reduced.

The saliency map is expressed as a linear combination of feature maps~\eq{v-sal}, with normalized weights. Hence, the saliency map is discouraged from taking up the entire image, both by the $\softmax$ competition~\eq{v-sal} and by the fact that feature maps only respond to particular locations.

\iavr{In case $g_c(\vy) \defn y_c$,~\eq{o-sal} takes the form of direct masking~\eq{mask} with $R(\vm) = \vzero$ and
\begin{equation}
	\cM \defn \{ S_\ell(\vx; \vu) : \vu \in \real^{K_\ell} \}.
\label{eq:mask-m}
\end{equation}
This constraint makes ours a CAM-based method. It dispenses the need for regularizers, because we only optimize one vector over the feature dimensions. In addition, it does not complicate the optimization process in any way. It is only a different parametrization.}


\iavr{
\section{\AGf ($\AG$)}

\redred{Average drop ($\AD$) and average increase ($\AI$)~\cite{chattopadhay2018grad} are well-established classification metrics. They measure the effect on the predicted class probabilities by masking the input image with the saliency map.} Let $p^c_i$ and $o^c_i$ be the predicted probability for class $c$ given as input the $i$-th test image $\vx_i$ and its masked version respectively. Masking refers to element-wise multiplication with the saliency map, which is at the same resolution as the original image with values in $[0,1]$. Let $N$ be the number of test images. Class $c$ is taken as the ground truth.

\emph{Average drop} ($\AD$) quantifies how much predictive power, measured as class probability, is lost when we only mask the image; lower is better:
\begin{equation}
	\AD(\%) \defn \frac{1}{N} \sum_{i=1}^N \frac{[p^c_i - o^c_i]_+}{p^c_i} \cdot 100.
\label{eq:ad}
\end{equation}

\emph{Average increase} ($\AI$), also known as \emph{increase in confidence}, measures the percentage of images where the masked image yields a higher class probability than the original; higher is better:
\begin{equation}
	\AI(\%) \defn \frac{1}{N} \sum_i^N \ind_{p^c_i < o^c_i} \cdot 100.
\label{eq:ai}
\end{equation}

$\AD$ and $\AI$ are not defined in a symmetric way. $\AD$ measures changes in class probability whereas $\AI$ measures a percentage of images. It is possible that the percentage is high while the actual increase is small. Hence, it is possible that an attribution method improves both. Indeed, \citep{poppi2021revisiting} observes that a trivial method called Fake-CAM outperforms state-of-the-art methods, including Score-CAM, by a large margin. Fake-CAM simply defines a saliency map where the top-left pixel is set to zero and is uniform elsewhere. This questions the purpose of $\AD$ and $\AI$.

Although the authors of~\citep{poppi2021revisiting} make this impressive observation, they use it to motivate the definition of a number of metrics that are orthogonal to the task at hand, \ie measuring the effect of masking to the classifier. By contrast, we address the problem by introducing a new metric to be paired with $\AD$ as a replacement of $\AI$. We define the new metric as follows.

\emph{\Agf} ($\AG$) quantifies how much predictive power, measured as class probability, is gained when we mask the image; higher is better:
\begin{equation}
	\AG(\%) \defn \frac{1}{N} \sum_{i=1}^N \frac{[o^c_i - p^c_i]_+}{1-p^c_i} \cdot 100.
\label{eq:ag}
\end{equation}
This definition is symmetric to the definition of average drop, in the sense that \redred{in absolute value, the numerator in the sum of $\AD, \AG$ is the positive and negative part of $p^c_i - o^c_i$ respectively and the denominator is the maximum value that the numerator can get as a function of $o^c_i$, given that $0 < o^c_i < p^c_i$ and $p^c_i < o^c_i < 1$ respectively.} The two metrics thus compete each other, in the sense that changing $o^c_i$ to improve one leaves the other unchanged or harms it. As we shall see, an extreme example is Fake-CAM, which yields near-perfect $\AD$ but fails completely on $\AG$.
}

\section{Experiments}
\label{sec:exp}

We evaluate Opti-CAM and compare it quantitatively and qualitatively against other state-of-the-art methods on a number of datasets and networks. \iavr{We report classification metrics with execution times and we provide visualizations, an ablation study and a study on the suitability of localization ground truth. A sanity check, additional classification results, localization metrics, more ablations, more visualizations \redred{and code} are given in supplementary material}.

\subsection{Datasets}
\label{sec:data}

\paragraph{ImageNet}

We use the validation set of ImageNet ILSVRC 2012~\citep{krizhevsky2012imagenet,ILSVRC15}, which contains $50,000$ images evenly distributed over the $1,000$ categories. For the ablation study and for timing, we sample $1,000$ images from this set. Concerning the localization experiments, bounding boxes from the localization task of ILSVRC\footnote{\url{https://www.image-net.org/challenges/LSVRC/2012/index.php}} are used on the same validation set.

\paragraph{Medical data}

\iavr{We use two medical image datasets, namely \emph{Chest X-ray} \citep{kermany2018labeled} and \emph{Kvasir} \citep{pogorelov2017kvasir}. Complete qualitative and quantitative results are given in the supplementary. Here we only provide visualizations.}


\paragraph{Networks}
\label{sec:setup}

For all datasets, we use the pretrained ResNet50~\citep{he2016deep} and VGG16~\cite{simonyan2014very} networks with batch normalization~\citep{ioffe2015batch} from the Pytorch model zoo\footnote{\url{https://pytorch.org/vision/0.8/models.html}}. \iavr{For ImageNet, we further use the pretrained ViT-B (16-224)~\citep{dosovitskiy2020image} and DeiT-B (16-224)~\citep{pmlr-v139-touvron21a} from Pytorch image models (timm)\footnote{\url{https://github.com/rwightman/pytorch-image-models}}.
Regarding medical datasets, we fine-tune the networks as discussed in the supplementary material, where we also provide the setting details.
}


\subsection{Evaluation}
\label{sec:eval}

\paragraph{Metrics}

\ronan{We use \emph{average drop} ($\AD$) and \emph{average increase} ($\AI$)~\cite{chattopadhay2018grad} metrics, as well as the proposed \emph{\agf} ($\AG$), to measure the effect on classification performance of masking the input image by a saliency map. In the supplementary, we also report \emph{insertion} (I) and \emph{deletion} (D)~\citep{petsiuk2018rise} and highlight their limitations. Using classification metrics, we show the limitations of using the localization ground truth for the evaluation of attribution methods. In the supplementary, we provide a number of localization metrics from the \emph{weakly-supervised object localization} (WSOL) task of ILSVRC2014\footnote{\url{https://www.image-net.org/challenges/LSVRC/2014/index\#}}.}

\paragraph{Methods}

\iavr{We compare against the following state-of-the-art methods: Grad-CAM~\citep{selvaraju2017grad}, Grad-CAM++~\cite{chattopadhay2018grad}, Score-CAM~\citep{wang2020score}, Ablation-CAM~\citep{ramaswamy2020ablation}, XGrad-CAM~\citep{fu2020axiom}, Layer-CAM~\citep{jiang2021layercam} and ExtremalPerturbation~\citep{fong2019understanding}. Implementations are obtained from the PyTorch CAM library\footnote{\url{https://github.com/jacobgil/pytorch-grad-cam}} or TorchRay\footnote{\url{https://github.com/facebookresearch/TorchRay}}. For transformer models, we also compare against raw attention~\citep{dosovitskiy2020image}, rollout~\citep{abnar2020quantifying} and TIBAV~\cite{chefer2021transformer}\footnote{\url{https://github.com/hila-chefer/Transformer-Explainability}}.}

\paragraph{Image normalization}

It is standard that images are normalized before feeding them to a network. By doing so however, we cannot reproduce the results published for the baseline methods; rather, all results are improved dramatically. We can obtain results similar to published ones by \emph{not} normalizing. We believe normalization is important and we include it in all our experiments. In the supplementary, we provide more details and results without normalization\redred{, as well as code that allows for reproduction and verification of our results}.

\subsection{Image classification}

Opti-CAM is evaluated quantitatively using classification metrics and qualitatively by visualizing saliency maps.

\begin{table}
\centering
\footnotesize
\setlength{\tabcolsep}{1.4pt}
\begin{tabular}{lrrrr|rrrr} \toprule
\mr{2}{\Th{Method}}                                & \mc{4}{\Th{ResNet50}} & \mc{4}{\Th{VGG16}} \\ \cmidrule{2-9}
                                                   & {$\AD\!\downarrow$} & {$\AG\!\uparrow$} & {$\AI\!\uparrow$} & \mc{1}{T} & {$\AD\!\downarrow$} & {$\AG\!\uparrow$} & {$\AI\!\uparrow$} & \mc{1}{T} \\ \midrule
Fake-CAM~\citep{poppi2021revisiting}               &  0.8 &  1.6 & 46.0 &  0.00 &  0.5 &  0.6 & 42.6 &  0.00 \\ \midrule
Grad-CAM~\citep{selvaraju2017grad}                 & 12.2 & 17.6 & 44.4 &  0.03 & 14.2 & 14.7 & 40.6 &  0.02 \\
Grad-CAM++~\cite{chattopadhay2018grad}             & 12.9 & 16.0 & 42.1 &  0.03 & 17.1 & 10.2 & 33.4 &  0.02 \\
Score-CAM~\citep{wang2020score}                    &  8.6 & 26.6 & 56.7 & 15.22 & 13.5 & 15.6 & 41.7 &  3.11 \\
Ablation-CAM~\citep{ramaswamy2020ablation}         & 12.5 & 16.4 & 42.8 & 18.26 & 15.5 & 12.6 & 36.9 &  2.98 \\
XGrad-CAM~\citep{fu2020axiom}                      & 12.2 & 17.6 & 44.4 &  0.03 & 13.8 & 14.8 & 41.2 &  0.02 \\
Layer-CAM~\citep{jiang2021layercam}                & 15.6 & 15.0 & 38.8 &  0.08 & 48.9 &  3.1 & 13.5 &  0.07 \\
ExPerturbation~\citep{fong2019understanding}       & 38.1 &  9.5 & 22.5 & 152.96 & 43.0 &  7.1 & 20.5 & 83.20 \\
\rowcolor{cyan!10}
Opti-CAM (ours)                                    & \tb{ 1.5} & \tb{68.8} & \tb{92.8} &  4.15 &  \tb{1.3} & \tb{71.2} & \tb{92.7} & 3.94 \\
\bottomrule
\end{tabular}
\caption{\emph{Classification metrics} on ImageNet validation set, using CNNs. $\AD$/$\AI$: average drop/increase~\citep{chattopadhay2018grad}; $\AG$: average gain (ours); $\downarrow$ / $\uparrow$: lower / higher is better; T: \iavr{Average time (sec) per batch of 8 images. Bold: best, excluding Fake-CAM.}}
\label{tab:imagenet-cnn}
\end{table}

\paragraph{CNN}

\autoref{tab:imagenet-cnn} shows ImageNet classification metrics using \Th{VGG16} and \Th{ResNet50}. Our Opti-CAM brings impressive performance in terms of average drop ($\AD$) and Average Increase ($\AI$) metrics. That is, not only impressive improvement over baselines, but near-perfect: near-zero $\AD$ and above 90\% $\AI$. \redred{Our new metric $\AG$ is lower, around 70\% for Opti-CAM, but this is still several times higher than for all the other methods.}

\iavr{
Interestingly, Fake-CAM~\citep{poppi2021revisiting} is the winner in terms of $\AD$ and second or third best in $\AI$ after Opti-CAM and Score-CAM, but fails completely $\AG$. This is expected and makes Fake-CAM uninteresting as it should be: By only masking one pixel, the classification score can hardly drop (0.8\% on ResNet50) and while it increases very often (on 46\% of images), the gain is as little as the drop (0.7\%). This makes the pair ($\AD$, $\AG$) sufficient as primary metrics and $\AI$ can be thought of as secondary, if important at all.

In the supplementary material we report \emph{insertion} (I) and \emph{deletion} (D) metrics along with failure cases of Opti-CAM. The latter indicate that our saliency maps are not incorrect as a whole, but capturing more parts of the object, more instances or more background context results in larger or several disconnected salient regions. This does not let the classifier focus on a single discriminative region when pixels are processed sequentially by increasing saliency. Rather, I/D favor smaller and more compact saliency maps.
}

\autoref{tab:imagenet-cnn} also includes average execution time per image over the 1000-image ImageNet subset for all methods. Opti-CAM is slower than gradient-based methods that require only one pass through the network, \iavr{but on par or faster than gradient-free methods. Indeed, we use a maximum of 100 iterations with one forward/backward pass per iteration, while Score-CAM and Ablation-CAM perform as many forward passes as channels. Hence they are much slower on ResNet50 than VGG16. ExtremalPerturbation does not depend on the number of channels but is very slow by performing a complex optimization in the image space.}

\begin{table}
\centering
\footnotesize
\setlength{\tabcolsep}{1.4pt}
\begin{tabular}{lrrrr|rrrr} \toprule
\mr{2}{\Th{Method}}                     & \mc{4}{\Th{ViT-B}} & \mc{4}{\Th{DeiT-B}} \\ \cmidrule{2-9}
                                        & {$\AD\!\downarrow$} & {$\AG\!\uparrow$} & {$\AI\!\uparrow$} & \mc{1}{T} & {$\AD\!\downarrow$} & {$\AG\!\uparrow$} & {$\AI\!\uparrow$} & \mc{1}{T} \\ \midrule
Fake-CAM~\citep{poppi2021revisiting}    &  0.3 &  0.4 & 48.3 &  0.00 &  0.6 &  0.3 & 44.6 &  0.00 \\ \midrule
Grad-CAM~\citep{selvaraju2017grad}      & 69.4 &  2.5 & 12.4 &  0.14 & 33.5 &  1.7 & 12.5 &  0.11 \\
Grad-CAM++~\cite{chattopadhay2018grad}  & 86.3 &  1.5 &  1.0 &  0.15 & 50.7 &  0.9 &  7.2 &  0.13 \\
Score-CAM~\citep{wang2020score}         & 32.0 &  6.2 & 33.0 & 23.69 & 53.6 &  2.2 & 12.2 & 22.47 \\
XGrad-CAM~\citep{fu2020axiom}           & 88.1 &  0.4 &  4.3 &  0.13 & 80.5 &  0.3 &  4.1 &  0.12 \\
Layer-CAM~\citep{jiang2021layercam}     & 82.0 &  0.2 &  2.9 &  0.24 & 88.9 &  0.4 &  2.6 & 0.24\\
ExPerturbation~\citep{fong2019understanding}&28.8&6.2&24.4&133.52&60.9&2.0&8.5&129.12\\
RawAtt~\citep{dosovitskiy2020image}     & 92.6 &  0.2 &  2.8 &  0.02 & 95.3 &  0.0 &  1.8 &  0.02 \\
Rollout~\citep{abnar2020quantifying}    & 42.1 &  5.6 & 20.9 &  0.02 & 55.2 &  0.8 &  7.9 &  0.02 \\
TIBAV~\cite{chefer2021transformer}      & 81.7 &  0.8 &  5.8 &  0.16 & 62.3 &  0.7 &  7.1 &  0.16 \\
\rowcolor{cyan!10}
Opti-CAM (ours)                         & \tb{ 0.6} &   \tb{18.0} & \tb{90.1} &    16.05 & \tb{ 0.9} & \tb{26.0} & \tb{83.5} &    15.17 \\ \bottomrule
\end{tabular}
\caption{\emph{Classification metrics} on ImageNet validation set, using transformers. $\AD$/$\AI$: average drop/increase~\citep{chattopadhay2018grad}; $\AG$: average gain (ours); $\downarrow$ / $\uparrow$: lower / higher is better. \iavr{T: Average time (sec) per batch of 8 images. Bold: best, excluding Fake-CAM.}}
\label{tab:imagenet-trans}
\end{table}

\paragraph{Transformers}

\autoref{tab:imagenet-trans} shows ImageNet classification metrics using ViT \iavr{and DeiT}. Unlike CAM-based methods that rely on a class-specific linear combination of feature maps, raw attention~\citep{dosovitskiy2020image} and rollout~\citep{abnar2020quantifying} use the attention map of the [CLS] token from the last attention block and from all blocks respectively. \iavr{This attention map depends only on the particular image and not on the target class, hence it is not really comparable. TIBAV~\cite{chefer2021transformer} uses both instance-specific and class-specific information.

Opti-CAM outperforms all other methods dramatically, reaching near-zero $\AD$ and $\AI$ above 80 or 90\%. \redred{According to our new $\AG$ metric, Opti-CAM still works while all other methods fail, but $\AG$ is much more conservative than $\AI$. On ViT-B for example, the classification score increases for 90.1\% of the images by masking with Opti-CAM, but the gain is only 18.0\% on average.}}

\begin{figure*}[t]
\newcommand{\sizeS}{.13}
\newcommand{\sizeP}{.13}
\newcommand{\hh}{.175\textwidth}
\newcommand{\ww}{.200\textwidth}
\setlength{\tabcolsep}{2pt}
\centering
\footnotesize
\begin{tabular}{cccccccc}
	& Input image &  Grad-CAM  & Grad-CAM++ & Score-CAM & Ablation-CAM & XGrad-CAM & Opti-CAM 
 \\

	\rotatebox{90}{~Grass Snake} &
	\includegraphics[trim={36mm 10mm 32mm 10mm},clip, width=\sizeP\textwidth]{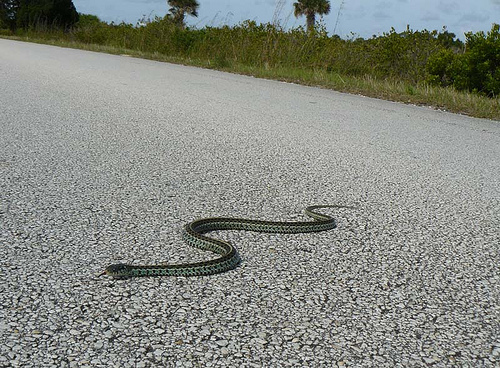}&
	\fig[\sizeS]{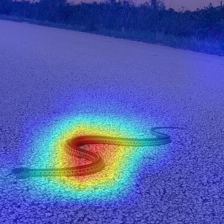} &
	\fig[\sizeS]{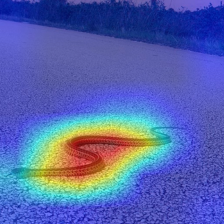} &
	\fig[\sizeS]{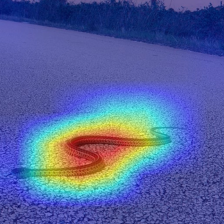} &
	\fig[\sizeS]{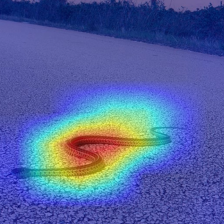} &
	\fig[\sizeS]{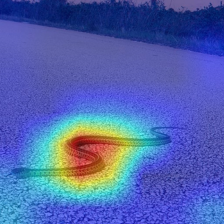} &
	\fig[\sizeS]{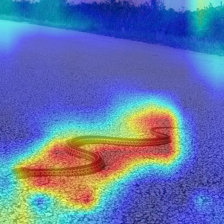}  \\

	\rotatebox{90}{~Tricycle} &
	\includegraphics[trim={28mm 10mm 39mm 10mm},clip, width=\sizeP\textwidth]{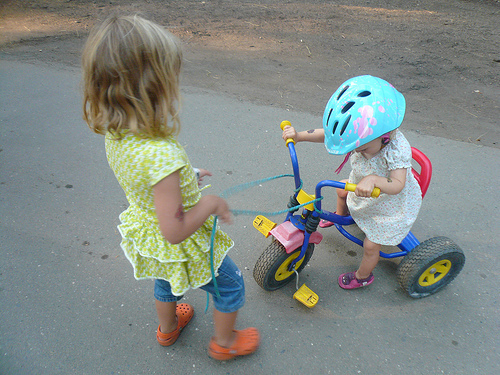}&
	\fig[\sizeS]{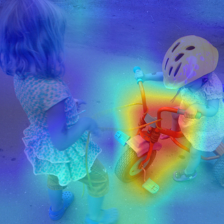} &
	\fig[\sizeS]{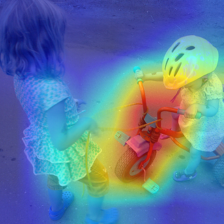} &
	\fig[\sizeS]{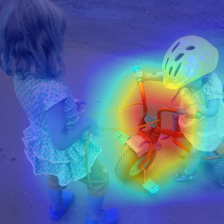} &
	\fig[\sizeS]{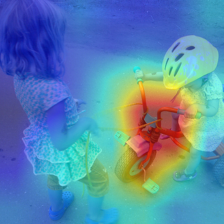} &
	\fig[\sizeS]{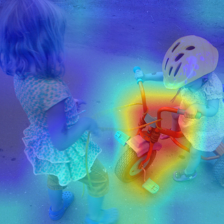} &
	\fig[\sizeS]{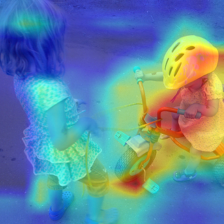}  \\

	\rotatebox{90}{~Pneumonia} &
	\fig[\sizeS]{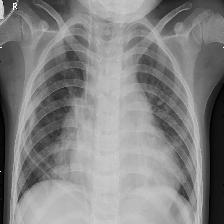} &
	\fig[\sizeS]{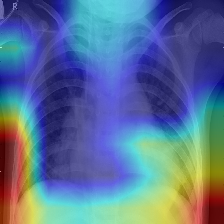} &
	\fig[\sizeS]{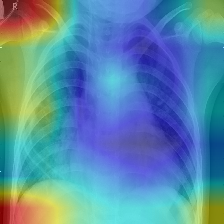} &
	\fig[\sizeS]{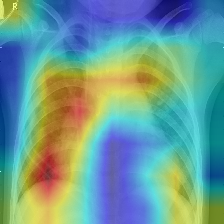} &
	\fig[\sizeS]{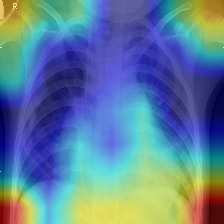} &
	\fig[\sizeS]{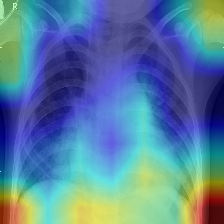} &
	\fig[\sizeS]{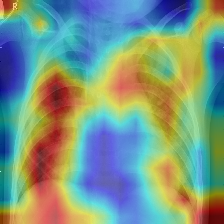}  \\

	\rotatebox{90}{~Pylorus} &
	\fig[\sizeS]{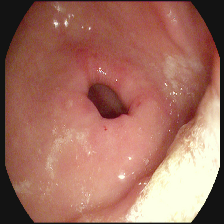} &
	\fig[\sizeS]{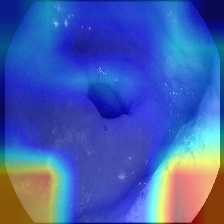} &
	\fig[\sizeS]{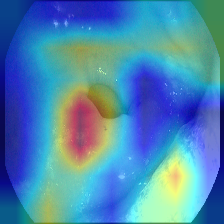} &
	\fig[\sizeS]{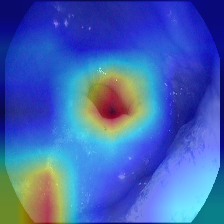} &
	\fig[\sizeS]{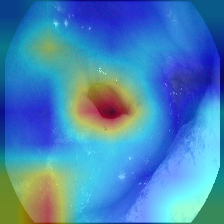} &
	\fig[\sizeS]{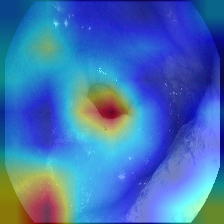} &
	\fig[\sizeS]{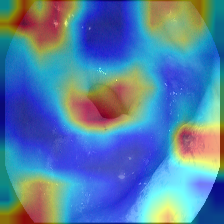}  \\

\end{tabular}
\caption{Saliency maps obtained by different methods for ImageNet (top two rows), Chest X-ray (row 3) and Kvasir (row 4) with VGG. \iavr{Ground truth class shown on the left of the input image.}}
\label{fig:vis-in-chest-n-kvasir-resnet}
\end{figure*}

\paragraph{Visualization}

\autoref{fig:vis-in-chest-n-kvasir-resnet} illustrates saliency map examples from ImageNet, Chest X-ray and Kvasir datasets. Opti-CAM saliency map is in general more spread out. This better highlights full objects, multiple instances or \iavr{background context, which may be taken into account by the model. On Chest X-ray, Opti-CAM and Score-CAM are the only methods that capture the chest, while all others focus on image corners.} More examples on datasets and networks \redred{as well as quantitative evaluation on medical data} are given in the supplementary material.

\subsection{Object localization}

\iavr{Localization metrics are used to measure the precision of saliency maps relative to ground truth bounding boxes of the foreground object of interest. These metrics originate from weakly supervised localization (WSOL). However, the objectives of WSOL and explaining the decision of a DNN are not necessary aligned, since context may play an important role in the decision~\cite{shetty2019not, rao2022towards}.

To investigate the relative importance of the object and its context, we measure classification metrics} when using the bounding box $B$ itself as saliency map as well as its complement $I \setminus B$, where $I$ is the image. We also evaluate the intersection $B \cup S$ of the saliency map $S$ with the bounding box and with its complement ($S \setminus B$).

As shown in \autoref{tab:localization}, the ground truth region of the object is not the only one responsible for the network decision. For example, the bounding box fails both when used as a saliency map itself and when combined with any saliency map, by harming all classification metrics. \iavr{Even the complement is more effective than the bounding box itself, either alone or when combined.} These findings support the hypothesis that localization metrics based on the ground truth bounding box are not necessarily appropriate for evaluating explanations of network decisions. Classification metrics are clearly more appropriate in this sense.

\iavr{Nevertheless, we report localization metrics in the supplementary material. \redred{In summary, although its saliency maps are more spread out, Opti-CAM outperforms other methods on a number of metrics.}}

\begin{table}[t]
\footnotesize
\centering
\setlength{\tabcolsep}{1.0pt}
\begin{tabular}{lccc|ccc|ccc} \toprule
\mr{2}{\Th{Method}}                            & \mc{3}{\Th{$\AD\!\downarrow$}} & \mc{3}{\Th{$\AG\!\uparrow$}}& \mc{3}{\Th{$\AI\!\uparrow$}} \\ \cmidrule{2-10}
                                               & {$S$} & {$B \!\cap\! S$} & {$S \!\setminus\! B$} & {$S$} & {$B \!\cap\! S$} & {$S \!\setminus\! B$}& {$S$} & {$B \!\cap\! S$} & {$S \!\setminus\! B$} \\ \midrule
$S \defn B$                                    & 67.2 &   -- &   -- &  2.3 &   -- &   -- &  9.2 &   -- &   -- \\
$S \defn I \setminus B$                        & 44.0 &   -- &   -- &  2.8 &   -- &   -- & 16.3 &   -- &   -- \\ \midrule
Fake-CAM~\citep{poppi2021revisiting}           &  0.5 & 67.2 & 44.1 &  0.7 &  2.3 &  2.8 & 42.0 &  9.2 & 18.9 \\ \midrule
Grad-CAM~\citep{selvaraju2017grad}             & 15.0 & 72.6 & 52.1 & 15.3 &  1.8 &  6.0 & 40.4 &  8.4 & 19.4 \\
Grad-CAM++~\cite{chattopadhay2018grad}         & 16.5 & 72.9 & 53.1 & 10.6 &  1.6 &  4.1 & 35.2 &  7.3 & 17.1 \\
Score-CAM~\citep{wang2020score}                & 12.5 & 71.5 & 50.5 & 16.1 &  2.2 &  6.3 & 42.5 &  8.6 & 20.8 \\
Ablation-CAM~\citep{ramaswamy2020ablation}     & 15.1 & 72.8 & 52.1 & 13.5 &  1.7 &  5.6 & 39.9 &  7.8 & 19.0 \\
XGrad-CAM~\citep{fu2020axiom}                  & 14.3 & 72.6 & 51.4 & 15.1 &  1.8 &  6.0 & 42.1 &  8.0 & 20.1 \\
Layer-CAM~\citep{jiang2021layercam}            & 49.2 & 84.2 & 74.4 &  2.7 &  0.4 &  1.2 & 12.7 &  4.4 &  7.3 \\
ExPerturbation~\citep{fong2019understanding}   & 43.8 & 81.6 & 71.0 &  7.1 &  1.4 &  3.2 & 18.9 &  5.6 & 11.1 \\
\rowcolor{cyan!10}
Opti-CAM (ours)                                & \tb{1.4} & \tb{62.5} & \tb{34.8} & \tb{66.3} & \tb{8.7} & \tb{25.8} & \tb{92.5} & \tb{18.6} & \tb{47.1} \\ \bottomrule
\end{tabular}
\caption{\emph{Bounding box} study. Classification metrics on ImageNet validation set using VGG16. $B$: ground-truth box used by localization metrics; $I$: entire image; $S$: saliency map. $\AD$/$\AI$: average drop/increase~\citep{chattopadhay2018grad}; $\AG$: average gain (ours); $\downarrow$ / $\uparrow$: lower / higher is better; bold: best, excluding Fake-CAM.}
\label{tab:localization}
\end{table}

\subsection{Ablation study}
\label{sec:ablation}

We perform an ablation study of different choices of the objective function~\eq{obj} and normalization~\eq{norm} of the saliency map. \redred{More choices of~\eq{obj}, layer $\ell$, number of iterations and learning rates, selector function $g_c$
and initialization of $\vw$ are studied in the supplementary material.}


\paragraph{Normalization function}

For normalization function $n$~\eq{obj}, we investigate three choices:
\begin{align}
	\textrm{range}   : \quad & n(A) \defn \textstyle \frac{A - \min A}{\max A - \min A}  \label{eq:n-rng}  \\
	\textrm{maximum} : \quad & n(A) \defn \textstyle \frac{A}{\max A}                    \label{eq:n-max}
	 \\
 	\textrm{sigmoid} : \quad & n(a_{ij}) \defn \frac{1}{1+e^{-a_{ij}}}             \label{eq:n-sig},
\end{align}
where $a_{ij}$ is element $(i,j)$ of matrix $A$. The default is~\eq{n-rng}, normalizing by the range of values in the saliency map, as in Score-CAM~\eq{norm}; while~\eq{n-max} normalizes by the maximum value and~\eq{n-sig} by the sigmoid function element-wise.


\paragraph{Objective function}

We refer to the default definition of $F^c_\ell$~\eq{obj} as \Fdef because it maximizes the logit for the masked image.
We also consider an alternative definition of objective function $F^c_\ell$, which encourages the masked version to preserve the prediction of original image:
\begin{equation}
	F^c_\ell(\vx; \vu) \defn -\abs{g_c(f(\vx)) - g_c(f(\vx \odot n(\up(S_\ell(\vx; \vu)))))}.
\label{eq:ref}
\end{equation}
This function is named \Fref as it minimizes the difference of logits between the masked and the original image.


\paragraph{Results}

\autoref{tab:ablate} shows classification metrics for the different choices of Opti-CAM, as well as comparison to other methods for reference, for the small subset of ImageNet validation set.

We observe that the choice of normalization function has little effect overall and Sigmoid offers lower performance. Note that the minimum value of saliency maps is often zero or close to zero: Saliency maps are non-negative as convex combinations of non-negative feature maps~\eq{v-sal}. By contrast, the choice of loss function has more impact on performance and we observe that \Fdef~\eq{obj} is superior on all cases.

\newcommand{\ob}[1]{\textcolor{brown}{\tb{#1}}}
\newcommand{\ab}[1]{\textcolor{blue}{\tb{#1}}}
\begin{table}
\centering
\footnotesize
\setlength{\tabcolsep}{3pt}
\begin{tabular}{lccrrrrr} \toprule
{\Th{Method}} & {$F^c_\ell$} & {$n$} & {$\AD\!\downarrow$}& {$\AG\!\uparrow$} & {$\AI\!\uparrow$} \\ \midrule
Fake-CAM~\citep{poppi2021revisiting}              & &          &     0.5  &      0.7  &     42.1  \\
\midrule
Grad-CAM~\citep{selvaraju2017grad}                & &          &    15.0  &     15.3  &     40.4  \\
Grad-CAM++~\cite{chattopadhay2018grad}               & &          &    16.5  &     10.6  &     35.2  \\
Score-CAM~\citep{wang2020score}                   & &          &    12.5  &     16.1  &     42.6  \\
Ablation-CAM~\citep{ramaswamy2020ablation}             & &          &    15.1  &     13.5  &     39.9  \\
XGrad-CAM~\citep{fu2020axiom}                        & &          &    14.3  &     15.1  &     42.1  \\
Layer-CAM~\citep{jiang2021layercam}               & &          &    49.2  &      2.7  &     12.7  \\
ExPerturbation~\citep{fong2019understanding}      & &          &    43.8  &      7.1  &     18.9  \\
\midrule
\mr{2}{Opti-CAM (ours)} & \Fdef~\eq{obj}  & Range~\eq{n-rng}   & \tb{1.4} & \tb{66.3} & \tb{92.5} \\
                        & \Fref~\eq{ref}  & Range~\eq{n-rng}   &     7.1  &     18.5  &     54.9  \\ \midrule
\mr{2}{Opti-CAM (ours)} & \Fdef~\eq{obj}  & Max~\eq{n-max}     &     1.6  &     66.2  &     90.3  \\
                        & \Fref~\eq{ref}  & Max~\eq{n-max}     &     6.8  &     17.8  &     54.5  \\ \midrule
\mr{2}{Opti-CAM (ours)} & \Fdef~\eq{obj}  & Sigmoid~\eq{n-sig} &     5.0  &     18.3  &     57.5  \\
                        & \Fref~\eq{ref}  & Sigmoid~\eq{n-sig} &     6.5  &     10.0  &     45.3  \\ \bottomrule
\end{tabular}
\caption{\emph{Ablation study} using VGG16 on 1000 images of ImageNet validation set. $\AD$/$\AI$: average drop/increase~\citep{chattopadhay2018grad}; $\AG$: average gain (ours); $\downarrow$ / $\uparrow$: lower / higher is better; bold: best, excluding Fake-CAM.}
\label{tab:ablate}
\end{table}

\pgfplotstableread{fig/eval/plain_ai.dat}{\plotAI}
\pgfplotstableread{fig/eval/plain_ad.dat}{\plotAD}
\pgfplotstableread{fig/eval/plain_ag.dat}{\plotAG}

\section{Discussion and conclusions}
\label{sec:conclusion}

Opti-CAM combines ideas of different saliency map generation methods, which are masking-based and CAM-based. Our method optimizes the saliency map at inference given a single input image. It does not require any additional data or training any other network, which would need interpretation too.

While Opti-CAM crafts a saliency map in the image space, it does not need any regularization. This is because the saliency map is expressed as a convex combination of feature maps and we only optimize one vector over the feature dimensions. The underlying assumption is that of all CAM-based methods: feature maps contain activations at all regions that are of interest for the classes that are present. Opti-CAM is more expensive than non-iterative gradient-based methods but as fast or faster than gradient-free methods that require as many forward passes as channels.

We find that Opti-CAM brings impressive performance improvement over the state of the art according to the most important classification metrics on several datasets. The saliency maps are more spread out compared with those of the competition, attending to larger parts of the object, multiple instances and background context, which may be helpful in classification.


\iavr{Our new classification metric $\AG$ aims to be paired $\AD$ as a replacement of $\AI$ and resolves a long-standing problem in evaluating attribution methods, without further increasing the number of metrics. We provide strong evidence supporting that the use of ground-truth object bounding boxes for localization is not necessarily optimal in evaluating the quality of a saliency map, because the primary objective is to explain how a classifier works.
}

\section*{Acknowledgements}
This publication has received funding from the Excellence Initiative of Aix-Marseille Universite - A*Midex, a French “Investissements d’Avenir programme” (AMX-21-IET-017), and the UnLIR ANR project (ANR-19-CE23-0009). Part of this work was performed using HPC resources from GENCI-IDRIS (Grant 2020-AD011013110).


%
%
\bibliographystyle{ieee}
\bibliography{refbib}

\clearpage

\clearpage




\appendix

\renewcommand{\theequation}{A\arabic{equation}}
\renewcommand{\thetable}{A\arabic{table}}
\renewcommand{\thefigure}{A\arabic{figure}}


\section*{Introduction}

Implementation details are provided in \autoref{sec:details}. We provide results on more classification metrics in \autoref{sec:cla-metrics}. In \autoref{sec:loc-metrics}, we define localization metrics and provide corresponding results. We provide results on medical data in \autoref{sec:medical}. We then provide more ablation results in \autoref{sec:more-ablation}, sanity check in \autoref{sec:sanity-check}, and results without input image normalization in \autoref{sec:without-norm}. Finally, we provide additional visualizations in \autoref{sec:more-vis}.


\section{Implementation details}
\label{sec:details}

All input images are resized to $224 \times 224 \times 3$. To optimize the saliency map with Opti-CAM~\eq{opt}, we use the Adam~\citep{kingma2014adam} optimizer with learning rate $0.1$ by default, setting the maximum number of iterations to $100$ and stopping early when the change in loss is less than $10^{-10}$. For VGG16, we generate the saliency map~\eq{v-sal} from the feature maps of the last convolutional layer before max pooling by default, \ie convolutional layer 3 of block 5. For ResNet50, we choose the last convolutional layer by default, \ie convolutional layer 3 of bottleneck 2 of block 4. For ViT and DeiT, we choose the last self-attention block by default, \ie layer normalization of self-attention block 12. Ablations concerning the layer $\ell$ and the convergence of Opti-CAM is included in \autoref{sec:more-ablation}.


\section{Classification metrics}
\label{sec:cla-metrics}

Classification metrics measure the effect on classification performance of masking (element-wise multiplying) the input image by the saliency map. We have used $\AD$, $\AG$ and $\AI$ in the main paper. Here we discuss Insertion/Deletion~\citep{petsiuk2018rise}, providing results and discussing failure cases for Opti-CAM.

\subsection{Insertion/Deletion}

\paragraph{Definition}

Insertion/Deletion~\citep{petsiuk2018rise} are based on the probability $p^{c_p}_i$ for the predicted class $c_p$ as pixels are ``inserted'' or ``deleted'' from image $\vx_i$, averaged over the number of pixels and over all images in the test set.

\emph{Deletion} measures the decrease in the probability of class $c_p$ when removing pixels one by one in decreasing order of saliency, where removal is taken as setting the value to zero; lower is better.

\emph{Insertion}, by contrast, measures the increase in the probability of class $c_p$ when adding pixels, again by decreasing order of saliency. In this case, we begin with a version of the image that is distorted by Gaussian blur and then addition is taken as setting the value of the pixel according to the original image. Higher is better.

\paragraph{Results}

The experimental results are shown in \autoref{tab:imagenet_cnn_hihd} for CNNs and \autoref{tab:imagenet-trans-hihd} for transformers. ExPerturbation~\citep{fong2019understanding} is expected to perform best in insertion because its optimization objective is very similar to this evaluation metric, using blurring for masked regions. However, ExPerturbation~\citep{fong2019understanding}  only performs best on ResNet50. TIBAV~\cite{chefer2021transformer}, which is designed for transformers, outperforms the other methods on DeiT and ViT. According to the results of Insertion/Deletion, Opti-CAM has low performance but there is no clear winner on either CNNs or transformers.

To further understand the behavior of Opti-CAM, we investigate in \autoref{fig:hihd} examples where Score-CAM succeeds (insertion score greater than $90$ and deletion score less than $10$) and Opti-CAM fails (insertion score less than $70$ and deletion score greater than $15$). Compared with Score-CAM, the saliency maps obtained by Opti-CAM are more spread out and highlight several parts of the object and background context. In most of the cases, Opti-CAM fails I/D because it not only finds the object but also attaches importance to the background.

We argue that this is not a failure. As our localization experiment in \autoref{tab:localization} indicates, background is useful in discriminating a class. Often, the network recognizes the background better than the object itself. \redred{For example, a gas pump is likely to be seen with a truck and a hare is often seen on grass. Several parts of the object are highlighted by Opti-CAM for the worm fence, terrier dog, hare, manhole cover. Finally, several instances of spaniel dog are found by Opti-CAM.}

Insertion/Deletion include 224 steps of binarization, with a set of 224 pixels being inserted/deleted at each step. If these  pixels are all inserted over a single small area, the effect on the classifier is more immediate than when sparsely inserting pixels over multiple areas. The same observation holds for deletion. By contrast, Opti-CAM attempts to find regions that contribute to the classification as a whole. There is no guarantee that those regions are effective when used in isolation.

\begin{table}
\centering
\footnotesize
\setlength{\tabcolsep}{8pt}
\begin{tabular}{lrr rr} \toprule
\mr{2}{\Th{Method}} & \mc{2}{\Th{ResNet50}} & \mc{2}{\Th{VGG16}} \\ \cmidrule{2-5}
                    & {{$\I\!\uparrow$}} & {{$\D\!\downarrow$}}& {{$\I\!\uparrow$}} & {{$\D\!\downarrow$}} \\ \midrule
Fake-CAM~\citep{poppi2021revisiting}&50.7&28.1&46.1&26.9\\\midrule
Grad-CAM~\citep{selvaraju2017grad}          &66.3&14.7&\tb{64.1}&11.6\\
Grad-CAM++~\cite{chattopadhay2018grad}     &66.0&14.7&62.9&12.2\\
Score-CAM~\citep{wang2020score}         &65.7&16.3&62.5&12.1\\
Ablation-CAM~\citep{ramaswamy2020ablation} &65.9&14.6&63.8&11.4\\
XGrad-CAM~\citep{fu2020axiom}             &66.3&14.7&\tb{64.1}&11.7\\
Layer-CAM~\citep{jiang2021layercam}&67.0&\tb{14.2}&58.3&\tb{6.4}\\
ExPerturbation~\citep{fong2019understanding}&\tb{70.7}&15.0&61.1&15.0\\
\rowcolor{cyan!10}
Opti-CAM (ours)                        &62.0&19.7&59.2&11.0\\
\bottomrule
\end{tabular}
\caption{
I/D: insertion/deletion~\citep{petsiuk2018rise} scores on ImageNet validation set; $\downarrow$ / $\uparrow$: lower / higher is better.}
\label{tab:imagenet_cnn_hihd}
\end{table}

\begin{table}
\centering
\footnotesize
\setlength{\tabcolsep}{8pt}
\begin{tabular}{lrr rr} \toprule
\mr{2}{\Th{Method}} & \mc{2}{\Th{DeiT-B}} & \mc{2}{\Th{ViT-B}} \\ \cmidrule{2-5}
                    & {{$\I\!\uparrow$}} & {{$\D\!\downarrow$}}&   {{$\I\!\uparrow$}} & {{$\D\!\downarrow$}} \\ \midrule
Fake-CAM~\citep{poppi2021revisiting}&57.5&34.2&57.4&33.3\\\midrule
Grad-CAM~\citep{selvaraju2017grad}          &61.8&17.5&62.9&19.8\\
Grad-CAM++~\cite{chattopadhay2018grad}     &60.5&21.9&56.7&29.3\\
Score-CAM~\citep{wang2020score}         &60.6&24.4&\tb{66.5}&15.1\\
XGrad-CAM~\citep{fu2020axiom}             &55.2&31.1&55.6&26.5\\
Layer-CAM~\citep{jiang2021layercam} &61.6&21.2&62.9&14.6\\
ExPerturbation~\citep{fong2019understanding}&62.1&27.0&64.4&18.4\\
RawAtt~\citep{dosovitskiy2020image}    &56.3&29.3&62.2&17.9\\
Rollout~\citep{abnar2020quantifying}    &56.7&32.8&64.8&15.2\\
TIBAV~\cite{chefer2021transformer}     &\tb{63.7}&\tb{16.3}&66.1&\tb{14.1}\\
\rowcolor{cyan!10}
Opti-CAM (ours)                        &59.2&22.8&60.5&22.0\\

\bottomrule
\end{tabular}
\caption{
\emph{I/D: insertion/deletion~\citep{petsiuk2018rise}} scores on ImageNet validation set; $\downarrow$ / $\uparrow$: lower / higher is better.}
\label{tab:imagenet-trans-hihd}
\end{table}

\begin{figure}[thpb]
\newcommand{\sizeP}{.12}
\newcommand{\sizeS}{.25}
\newcommand{\hh}{.175\textwidth}
\newcommand{\ww}{.200\textwidth}
\tiny
\centering
\setlength{\tabcolsep}{3pt}
\begin{tabular}{ccc}
\centering
Original & Opti-CAM & Score-CAM\\
\includegraphics[trim={28mm 8mm 28mm 8mm},clip, width=\sizeP\textwidth]{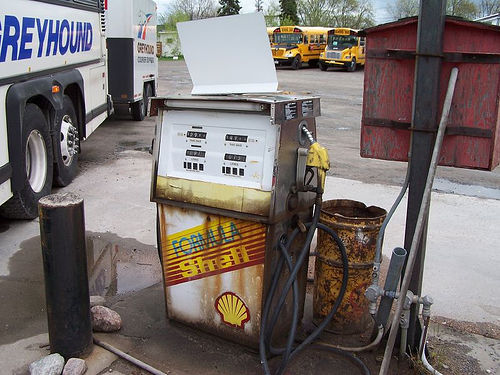}
&
\fig[\sizeS]{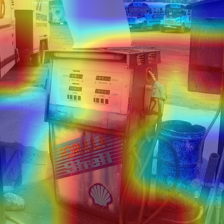} 
&  
\fig[\sizeS]{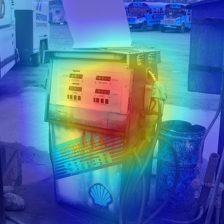} \\
gas pump&I$\uparrow$:66.3, D$\downarrow$:19.4&I$\uparrow$:94.2, D$\downarrow$:9.4\\
&AG$\uparrow$:100.0, AD$\downarrow$:0.0&AG$\uparrow$:0.0, AD$\downarrow$:0.0\\
\includegraphics[trim={32mm 14mm 36mm 1mm},clip, width=\sizeP\textwidth]{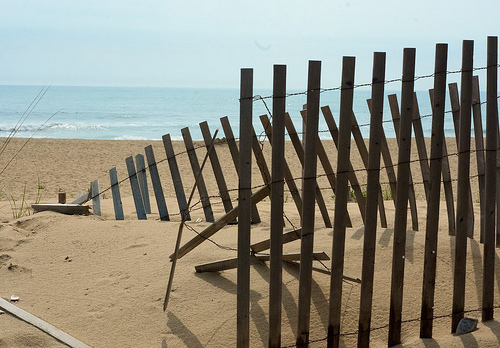}
&
\fig[\sizeS]{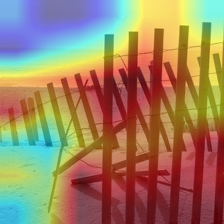} 
&          
\fig[\sizeS]{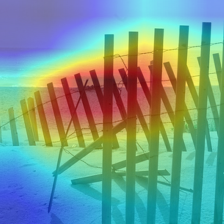} \\
worm fence&I$\uparrow$:69.7, D$\downarrow$:16.8&I$\uparrow$:91.9, D$\downarrow$:4.4\\
&AG$\uparrow$:73.2, AD$\downarrow$:0.0&AG$\uparrow$:0.0, AD$\downarrow$:28.8\\
\includegraphics[trim={10mm 14mm 10mm 4mm},clip, width=\sizeP\textwidth]{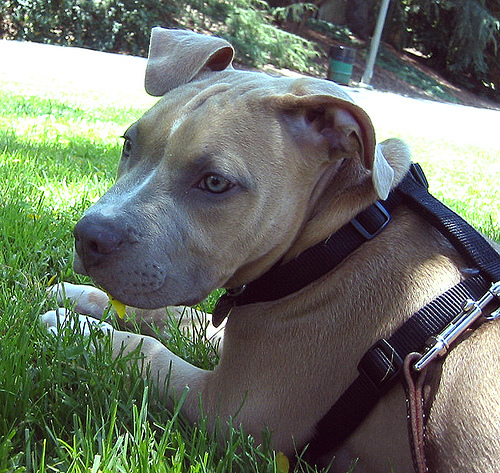}
&        
\fig[\sizeS]{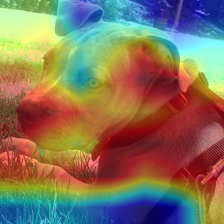} 
&
\fig[\sizeS]{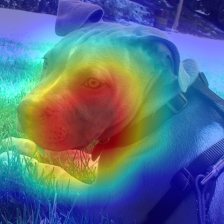} \\
staffordshire terrier&I$\uparrow$:62.1, D$\downarrow$:32.2&I$\uparrow$:93.4, D$\downarrow$:8.2\\
&AG$\uparrow$:41.3, AD$\downarrow$:0.0&AG$\uparrow$:0.0, AD$\downarrow$:0.3\\
\includegraphics[trim={18mm 6mm 10mm 12mm},clip, width=\sizeP\textwidth]{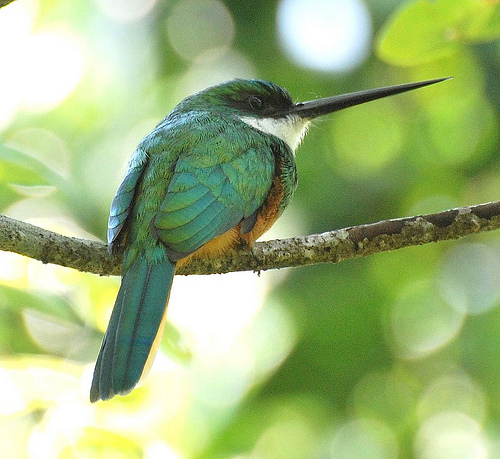}
&
\fig[\sizeS]{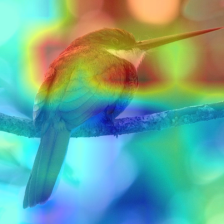} 
&                
\fig[\sizeS]{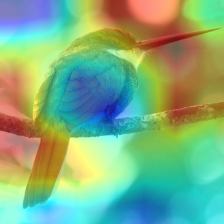} \\
jacamar&I$\uparrow$:66.3, D$\downarrow$:17.3&I$\uparrow$:94.6, D$\downarrow$:9.9\\
&AG$\uparrow$:91.4, AD$\downarrow$:0.0&AG$\uparrow$:56.5, AD$\downarrow$:0.0\\
\includegraphics[trim={6mm 1mm 6mm 1mm},clip, width=\sizeP\textwidth]{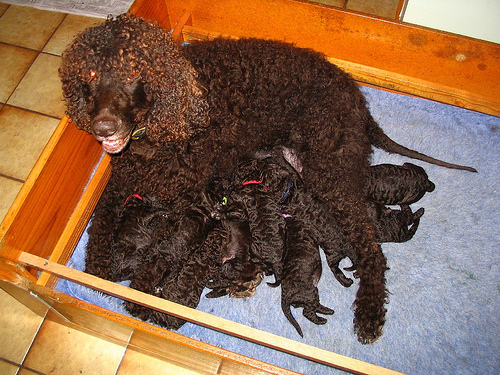}
&
\fig[\sizeS]{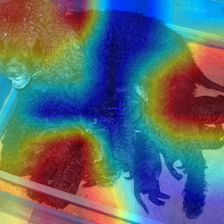} 
&     
\fig[\sizeS]{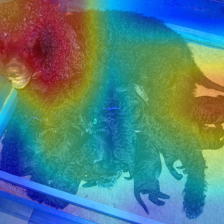} \\
Irish water spaniel&I$\uparrow$:52.6, D$\downarrow$:18.8&I$\uparrow$:90.5, D$\downarrow$:8.6\\
&AG$\uparrow$:86.4, AD$\downarrow$:0.0&AG$\uparrow$:65.1, AD$\downarrow$:0.0\\
\includegraphics[trim={28mm 5mm 22mm 5mm},clip, width=\sizeP\textwidth]{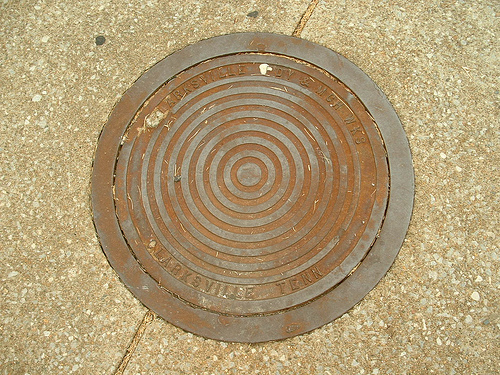}
&
\fig[\sizeS]{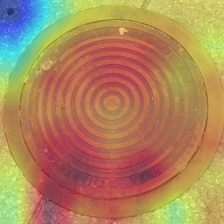} 
&          
\fig[\sizeS]{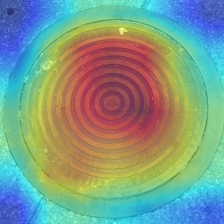} \\
manhole cover&I$\uparrow$:65.8, D$\downarrow$:29.6&I$\uparrow$92.7, D$\downarrow$:9.1\\
&AG$\uparrow$:24.0, AD$\downarrow$:0.0&AG$\uparrow$:0.0, AD$\downarrow$:59.9\\
\includegraphics[trim={12mm 5mm 12mm 5mm},clip, width=\sizeP\textwidth]{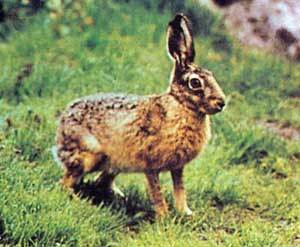}
&              
\fig[\sizeS]{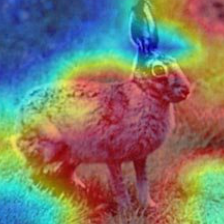} 
& 
\fig[\sizeS]{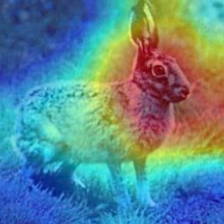} \\
hare&I$\uparrow$:61.3, D$\downarrow$:21.2&I$\uparrow$91.3, D$\downarrow$:8.9\\
&AG$\uparrow$:93.7, AD$\downarrow$:0.0&AG$\uparrow$:0.0, AD$\downarrow$:0.6\\
\end{tabular}
\caption{Failure examples of Opti-CAM regarding insertion/deletion.}
\label{fig:hihd}
\end{figure}


\section{Localization metrics}
\label{sec:loc-metrics}

Several works measure the localization ability of saliency maps, using metrics from the \emph{weakly-supervised object localization} (WSOL) task. While we show in the main paper that localization of the object and classifier interpretability are not well aligned as tasks, we still provide localization results here. We use the \emph{official metric} (OM), \emph{localization error} (LE), \emph{pixel-wise $F_1$ score}, \emph{box accuracy} (BoxAcc)~\citep{choe2020evaluating}, standard pointing game (SP)~\cite{zhang2018top}, \emph{energy pointing game} (EP)~\citep{wang2020score} and \emph{saliency metric} (SM)~\citep{dabkowski2017real} on the ILSVRC2014\footnote{\url{https://www.image-net.org/challenges/LSVRC/2014/index\#}} dataset. The goal of these metrics is to compare the saliency maps with bounding boxes around the object of interest. For simplicity, we define these metrics for a single image; the reported results are averaged over all images of the test set.

\subsection{Definitions}

We are given the saliency map $S^c$ obtained from test image $\vx$ for ground truth class $c$. We denote by $S^c_{\vp}$ its value at pixel $\vp$. We binarize the saliency map by thresholding at its average value and we take the bounding box of the largest connected component of the resulting mask as the predicted bounding box $B_p$, represented as a set of pixels. We compare this box against the set of ground truth bounding boxes $\cB$, which typically contains 1 or 2 boxes of the same class $c$, or with their union $U = \cup \cB$, again represented as a set of pixels. We also compare the predicted class label $c_p$ with the ground truth label $c$. All metrics take values in $[0,1]$ and are expressed as percentages, except SM~\eq{sm}, which is unbounded.

\paragraph{Official Metric (OM)}

measures the maximum overlap of the predicted bounding box with any ground truth bounding box, requiring that the predicted class label is correct:
\begin{equation}
	\OM \defn 1 - \paren{\max_{B \in \cB} \iou(B, B_p)} \ind_{c_p = c},
\label{eq:om}
\end{equation}
where $\iou$ is intersection over union.

\paragraph{Localization Error (LE)}

is similar but ignores the predicted class label:
\begin{equation}
	\LE \defn 1 - \max_{B \in \cB} \iou(B, B_p).
\label{eq:le}
\end{equation}

\paragraph{Pixel-wise $F_1$ score (F1)}

is defined as $F_1 = 2 \frac{P R}{P + R}$, where \emph{precision} $P$ is the fraction of mass of the saliency map that is within the ground truth union
\begin{equation}
	P \defn \frac{\sum_{\vp \in U} S^c_{\vp}}{\sum_{\vp} S^c_{\vp}}
\label{eq:prec}
\end{equation}
and \emph{recall} $R$ is the fraction of the ground truth union that is covered by the saliency map
\begin{equation}
	R \defn \frac{\sum_{\vp \in U} S^c_{\vp}}{\card{U}}.
\label{eq:rec}
\end{equation}

\paragraph{Box Accuracy (BA)~\citep{choe2020evaluating}}

Given threshold values $\eta$ and $\delta$, we find the bounding box $B^\eta_p$ of the largest connected component of the binary mask $\set{\vp: S_{\vp} > \eta}$ and require that it overlaps by $\delta$ with at least one ground truth box:
\begin{equation}
	\BA(\eta, \delta) \defn \max_{B \in \cB} \ind_{\iou(B^\eta_p, B) \ge \delta}.
\label{eq:ba}
\end{equation}
After averaging over the test images, we take the maximum of this measure over a set of values $\eta$ and then the average over a set of values $\delta$.

\paragraph{Standard Pointing game (SP)~\cite{zhang2018top}}

We find the pixel $\vp^* \defn \arg\max_{\vp} S^c_{\vp}$ having the maximum saliency value and require that it lands in any of the ground truth bounding boxes:
\begin{equation}
	\spg \defn \ind_{\vp^* \in U}.
\label{eq:spg}
\end{equation}

\paragraph{Energy Pointing game (EP)~\citep{wang2020score}}

is equivalent to precision~\eq{prec}.

\paragraph{Saliency Metric (SM)~\citep{dabkowski2017real}}

penalizes the size of the predicted bounding box $B_p$ relative to the image and the cross-entropy loss:
\begin{equation}
	\SM \defn \log \max\paren{ 0.05, \frac{\card{B_p}}{hw} } - \log p^c,
\label{eq:sm}
\end{equation}
where $h \times w$ is the input image resolution and $p^c$ is the precicted probability for ground truth class label $c$.

\begin{table}[ht]
\centering
\footnotesize
\setlength{\tabcolsep}{3pt}
\begin{tabular}{lccc|cccc} \toprule
\Th{method} & {OM$\downarrow$} & {LE$\downarrow$} & {F1$\uparrow$}&{BA$\uparrow$}& {SP$\uparrow$} & {EP$\uparrow$} & {SM$\downarrow$} \\ \midrule
\mc{8}{\Th{ResNet50}}     \\ \midrule
Fake-CAM~\citep{poppi2021revisiting}               &63.6&54.0&57.7&47.9&99.8&28.5&0.98\\ \midrule
Grad-CAM~\citep{selvaraju2017grad}         &72.9&65.8&49.8&\tb{56.2}&69.8&33.3&1.30            \\
Grad-CAM++~\cite{chattopadhay2018grad}     &73.1&66.1&\tb{50.4}&\tb{56.2}&69.9&33.1&1.29              \\
Score-CAM~\citep{wang2020score}            &\tb{72.2}&64.9&49.6&54.5&68.7&32.4&\tb{1.25}       \\
Ablation-CAM~\citep{ramaswamy2020ablation} &72.8&65.7&50.2&56.1&69.9&33.1&1.26            \\
XGrad-CAM~\citep{fu2020axiom}              &72.9&65.8&49.8&\tb{56.2}&69.8&33.3&1.30             \\
Layer-CAM~\citep{jiang2021layercam} &73.1&66.0&50.1&55.5&\tb{70.0}&33.0&1.29\\
ExPerturbation~\citep{fong2019understanding}  &73.6&66.6&37.5&44.2&64.8&\tb{38.2}&1.59\\
\rowcolor{cyan!10}
Opti-CAM (ours)                            &\tb{72.2}&\tb{64.8}&47.3&49.2&59.4&30.5&1.34             \\ \midrule
\mc{8}{\Th{VGG16}}                        \\ \midrule
Fake-CAM~\citep{poppi2021revisiting}               &64.7&54.0&57.7&47.9&99.8&28.5&1.07 \\ \midrule
Grad-CAM~\citep{selvaraju2017grad}         &71.1&62.3&42.0&54.2&64.8&32.0&1.39            \\
Grad-CAM++~\cite{chattopadhay2018grad}     &70.8&61.9&44.3&55.2&66.2&32.3&1.38           \\
Score-CAM~\citep{wang2020score}            &71.2&62.5&\tb{45.3}&\tb{58.5}&\tb{68.2}&33.4&1.40             \\
Ablation-CAM~\citep{ramaswamy2020ablation} &71.3&62.6&43.2&56.2&65.7&32.7&1.39            \\
XGrad-CAM~\citep{fu2020axiom}              &70.8&62.0&41.9&53.5&64.4&31.6&1.41             \\
Layer-CAM~\citep{jiang2021layercam} &70.5&61.5&28.0&54.7&65.0&32.4&1.45\\
ExPerturbation~\citep{fong2019understanding}  &74.1&66.4&37.8&43.3&62.7&\tb{36.1}&1.74\\
\rowcolor{cyan!10}
Opti-CAM (ours)                            &\tb{69.1}&\tb{59.9}&44.1&51.2&61.4&30.7&\tb{1.34}        \\ \bottomrule
\end{tabular}
\caption{\emph{Localization metrics} on ImageNet validation set. OM: \emph{official metric}; LE: \emph{localization error}; F1: \emph{pixel-wise $F_1$ score}; BA: box accuracy; SP: standard pointing game; EP: energy pointing game; SM: \emph{saliency metric}. $\downarrow$ / $\uparrow$: lower / higher is better. Bold: best, excluding Fake-CAM.}
\label{tab:imagenet-loc}
\end{table}
\begin{table}[t]
\centering
\footnotesize
\setlength{\tabcolsep}{3pt}
\begin{tabular}{lrrr|rrrr}
\toprule
\Th{method} & {OM$\downarrow$} & {LE$\downarrow$} & {F1$\uparrow$}&{BA$\uparrow$}& {SP$\uparrow$} & {EP$\uparrow$} & {SM$\downarrow$} \\ \midrule
\mc{8}{ViT-B} \\ \midrule
Fake-CAM~\citep{poppi2021revisiting}   &62.8&54.0&57.7&47.9&99.8&28.6&0.87 \\ \midrule
Grad-CAM~\citep{selvaraju2017grad}                   &79.6&74.3&29.4&45.0&58.1&31.0&3.27\\
Grad-CAM++~\cite{chattopadhay2018grad}               &84.2&80.6&14.8&23.8&51.4&27.3&4.15\\
Score-CAM~\citep{wang2020score}                   &77.6&71.6&46.0&54.3&\tb{66.1}&33.1&3.14 \\
XGrad-CAM~\citep{fu2020axiom}                      &82.0&76.9&19.6&41.3&52.8&28.5&3.31\\
Layer-CAM~\citep{jiang2021layercam}&70.7&63.9&20.6&50.5&60.7&32.6&1.44\\
ExPerturbation~\citep{fong2019understanding}&71.5&64.9&35.9&44.6&62.3&\tb{35.3}&1.34\\
RawAtt~\citep{dosovitskiy2020image}  &72.4&64.8&18.5&50.4&55.4&31.6&1.68\\
Rollout~\citep{abnar2020quantifying} &67.6&58.8&36.9&50.7&57.8&30.0&1.16\\
TIBAV~\cite{chefer2021transformer}&70.1&63.1&26.6&\tb{58.8}&\tb{66.1}&35.0&1.23\\
\rowcolor{cyan!10}
Opti-CAM (ours)     
&\tb{64.4}&\tb{54.6}&\tb{54.5}&48.0&58.2&28.7&\tb{0.98}\\\midrule
 \mc{8}{DeiT-B} \\ \midrule
Fake-CAM~\citep{poppi2021revisiting}    &61.4&54.0&57.7&47.9&99.8&28.7&0.83 \\ \midrule
Grad-CAM~\citep{selvaraju2017grad}                 &65.5&60.3&44.3&47.2&{62.8}&{30.2}&1.20\\
Grad-CAM++~\cite{chattopadhay2018grad}             &70.6&67.2&34.3&43.6&57.7&30.3&2.14\\
Score-CAM~\citep{wang2020score}                  &79.9&76.2&31.9&43.8&\tb{63.4}&32.2&3.14 \\
XGrad-CAM~\citep{fu2020axiom}                      &82.0&78.4&19.5&44.1&53.4&28.8&3.03\\
Layer-CAM~\citep{jiang2021layercam}&80.2&77.3&17.6&50.8&62.7&35.1&3.15\\
ExPerturbation~\citep{fong2019understanding}&69.9&64.3&36.2&44.2&63.1&\tb{35.5}&1.16\\
RawAtt~\citep{dosovitskiy2020image} &73.5&68.2&5.9&\tb{48.1}&46.5&27.3&1.91\\
Rollout~\citep{abnar2020quantifying} &63.9&57.0&27.8&47.9&36.5&27.2&0.94\\
TIBAV~\cite{chefer2021transformer}&68.2&62.2&28.1&59.6&64.1&33.5&1.08\\
\rowcolor{cyan!10}
Opti-CAM     
&\tb{62.3}&\tb{55.1}&\tb{53.9}&48.0&55.1&28.8&\tb{0.84}\\
\bottomrule
\end{tabular}
\vspace{5pt}
\caption{\emph{Localization metrics} with ViT and DeiT on ImageNet validation set. OM: \emph{official metric}; LE: \emph{localization error}; F1: \emph{pixel-wise $F_1$ score}; BA: box accuracy;
SP: standard pointing game; EP: energy pointing game; SM: \emph{saliency metric}.
$\downarrow$ / $\uparrow$: lower / higher is better. Bold: best, excluding Fake-CAM.}
\label{tab:ablate-loc-sup-deit}
\end{table}

\subsection{Results}

We evaluate the localization ability of saliency maps obtained by our Opti-CAM and we compare with other attribution methods quantitatively. \autoref{tab:imagenet-loc} and \autoref{tab:ablate-loc-sup-deit} report localization metrics on ImageNet. We observe different behavior in different metrics. In particular, Opti-CAM on ResNet and VGG performs best on OM and LE but poorly on the remaining metrics. On transformers, Opti-CAM performs best on OM, LE, F1, and SM.

Metrics where Opti-CAM does not perform well are mostly the ones that penalize saliency maps that are more spread out. For example, SP and EP penalize saliency outside the ground truth bounding box of an object. This is not necessarily a weakness of Opti-CAM, because rather than weakly supervised object localization, the objective here is to explain how the classifier works.


\section{Medical data}
\label{sec:medical}

Medical image recognition is a high-stakes task that crucially needs interpretable models. We thus evaluate our method on two standard medical image classification datasets.

\subsection{Datasets}

\paragraph{Chest X-ray}

\citep{kermany2018labeled} aims at recognizing chest images of patients with pneumonia from healthy ones with $5,216$ training images, $16$ for validation and $624$ for testing. Images are resized to $224 \times 224 \times 3$ to adapt to the pretrained models.

\paragraph{Kvasir}

\citep{pogorelov2017kvasir} contains $8$ classes and aims at recognizing anatomical landmarks, pathological findings and endoscopic procedures inside the gastrointestinal tract. The $8,000$ images are split into $6,000$ images for training, $1,000$ for validation and $1,000$ for testing. Images are resized as for the other datasets

\subsection{Network fine-tuning}

To train our models on the medical data, we first train the last fully-connected layer according to the classes in each dataset, while keeping the backbone frozen. On Chest X-ray, we use learning rate  $10^{-3}$ for both networks. On Kvasir, we use learning rate $10^{-4}$ for ResNet50 and $5\times10^{-3}$ for VGG16. We then fine-tune the entire network with learning rate $10^{-5}$ for 50 epochs, using SGD with momentum 0.9 for both networks on both datasets. On Chest X-ray data, we obtain accuracies of $83.2\%$ for VGG16 and $82.0\%$ for ResNet50; on Kvasir, $89.5\%$ for VGG16 and $89.8\%$ for ResNet50.

\begin{table}
\centering
\footnotesize
\setlength{\tabcolsep}{3pt}
\begin{tabular}{lrrr|rrr} \toprule
 \mr{2}{\Th{Method}}                        & \mc{3}{\Th{ResNet50}}                                                  & \mc{3}{\Th{VGG16}} \\
\cmidrule{2-7}
                               & {$\AD\!\downarrow$} & {$\AG\!\uparrow$} & {$\AI\!\uparrow$} & {$\AD\!\downarrow$} & {$\AG\!\uparrow$} & {$\AI\!\uparrow$} \\ \midrule
\mc{7}{\Th{Chest X-ray}}     \\ \midrule
 Fake-CAM~\citep{poppi2021revisiting} &0.1&0.9&49.7&0.1&0.4&29.8\\\midrule
 Grad-CAM~\citep{selvaraju2017grad}         &20.4&29.7&48.7&36.8&39.8&42.3\\
 Grad-CAM++~\cite{chattopadhay2018grad}     &24.7&24.1&41.2&36.9&43.4&45.8\\
 Score-CAM~\citep{wang2020score}            &21.6&27.7&44.2&35.3&47.4&48.9\\
 Ablation-CAM~\citep{ramaswamy2020ablation} &26.2&27.9&42.9&36.9&46.9&47.8\\
 XGrad-CAM~\citep{fu2020axiom}              &20.4&29.7&48.7&34.7&47.3&50.2\\
 Layer-CAM~\citep{jiang2021layercam} &24.5&23.4&39.1&36.6&45.9&47.6\\
 ExPerturbation~\citep{fong2019understanding}&21.4&5.5&17.9&29.7&21.8&28.7\\
 \rowcolor{cyan!10}
 Opti-CAM (ours)                            &\tb{0.1}&\tb{91.2}&\tb{98.4}&\tb{0.0}&\tb{85.9}&\tb{86.2}\\
\midrule
\mc{7}{\Th{Kvasir}}     \\ \midrule
 Fake-CAM~\citep{poppi2021revisiting} &0.1&0.4&48.3&0.0&0.3&45.0\\\midrule
 Grad-CAM~\citep{selvaraju2017grad}         &10.0&23.2&39.8&33.8&6.3&14.6\\
 Grad-CAM++~\cite{chattopadhay2018grad}     &11.2&18.7&32.9&20.7&9.3&20.4\\
Score-CAM~\citep{wang2020score}            &9.1&26.7&40.8&8.4&24.0&39.4\\
 Ablation-CAM~\citep{ramaswamy2020ablation} &10.7&21.6&35.4&10.6&20.9&36.9\\
 XGrad-CAM~\citep{fu2020axiom}              &10.0&23.2&39.8&12.1&21.6&35.2\\
 Layer-CAM~\citep{jiang2021layercam} &11.7&18.2&32.5&12.9&17.1&30.8\\
 ExPerturbation~\citep{fong2019understanding}&48.4&13.8&21.0&34.8&19.0&27.7\\
 \rowcolor{cyan!10}
 Opti-CAM (ours)                            &\tb{0.2}&\tb{91.1}&\tb{99.0}&\tb{0.0}&\tb{93.5}&\tb{98.1}\\
 \bottomrule
\end{tabular}
\caption{\emph{Classification metrics} on Chest X-ray and KVASIR datasets. $\AD$/$\AI$: average drop/increase~\citep{chattopadhay2018grad}; $\AG$: average gain (ours); $\downarrow$ / $\uparrow$: lower / higher is better; Bold: best, excluding Fake-CAM.}
\label{tab:xray-n-kvasir}
\end{table}

\subsection{Results}

\autoref{tab:xray-n-kvasir} reports metrics on Chest X-ray and Kvasir using \Th{ResNet50} and \Th{VGG16} networks. The conclusions remain the same as for ImageNet. More than that, AD and AI are near perfect in most cases and AG is also extremely high. Additional visualizations are presented in Section \autoref{sec:more-vis}.


\section{More ablations}
\label{sec:more-ablation}

\subsection{Selectivity}

We investigate the effect of selectivity of saliency maps on classification performance. In particular, before evaluation, we raise saliency maps element-wise to an exponent $\alpha$ that takes values in $\{0.01,0.05,0.1,0.5,1,1.5,2,3,5,10\}$. When $\alpha$ is small, the saliency maps become more uniform, so that more information about the original image is revealed to the network. Respectively, when $\alpha$ is large, the saliency maps become more selective, so that the network sees less parts of the input. The order of pixels is maintained.

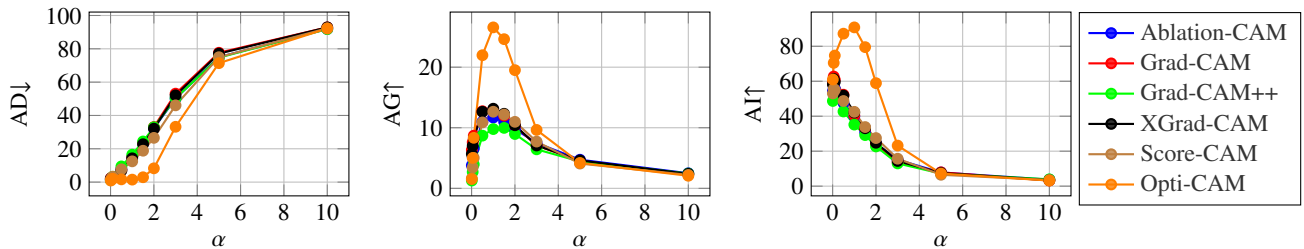
\begin{figure*}[htp!]
\tiny
\centering
\setlength{\tabcolsep}{3pt}
\footnotesize
\begin{tabular}{ccc}
\centering
\extfig{AD}{
\begin{tikzpicture}
\begin{axis}[
	height=4cm,
	width=5cm,
	ylabel={AD$\downarrow$},
	xlabel={$\alpha$},
	legend pos=outer north east,
]
	\addplot[mark=*,blue] table{fig/eval/adaiag/maskin_AblationCAM_ad.txt}; \leg{Ablation-CAM}
	\addplot[mark=*,red] table{fig/eval/adaiag/maskin_GradCAM_ad.txt}; \leg{Grad-CAM}
	\addplot[mark=*,green] table{fig/eval/adaiag/maskin_GradCAMPlusPlus_ad.txt}; \leg{Grad-CAM++}
	\addplot[mark=*,black] table{fig/eval/adaiag/maskin_XGradCAM_ad.txt}; \leg{XGrad-CAM}
	\addplot[mark=*,brown] table{fig/eval/adaiag/maskin_ScoreCAM_ad.txt}; \leg{Score-CAM}
	\addplot[mark=*,orange] table{fig/eval/adaiag/maskin_versionP0_ad.txt}; \leg{Opti-CAM}
    \legend{};
\end{axis}
\end{tikzpicture}
}
&
\extfig{AG}{
\begin{tikzpicture}
\begin{axis}[
	height=4cm,
	width=5cm,
	ylabel={AG$\uparrow$},
	xlabel={$\alpha$},
	legend pos=outer north east,
]
	\addplot[mark=*,blue] table{fig/eval/adaiag/maskin_AblationCAM_ag.txt}; \leg{Ablation-CAM}
	\addplot[mark=*,red] table{fig/eval/adaiag/maskin_GradCAM_ag.txt}; \leg{Grad-CAM}
	\addplot[mark=*,green] table{fig/eval/adaiag/maskin_GradCAMPlusPlus_ag.txt}; \leg{Grad-CAM++}
	\addplot[mark=*,black] table{fig/eval/adaiag/maskin_XGradCAM_ag.txt}; \leg{XGrad-CAM}
	\addplot[mark=*,brown] table{fig/eval/adaiag/maskin_ScoreCAM_ag.txt}; \leg{Score-CAM}
	\addplot[mark=*,orange] table{fig/eval/adaiag/maskin_versionP0_ag.txt}; \leg{Opti-CAM}
	\legend{};
\end{axis}
\end{tikzpicture}
}
&
\extfig{AI}{
\begin{tikzpicture}
\begin{axis}[
	height=4cm,
	width=5cm,
	ylabel={AI$\uparrow$},
	xlabel={$\alpha$},
	legend pos=outer north east,
]
	\addplot[mark=*,blue] table{fig/eval/adaiag/maskin_AblationCAM_ai.txt}; \leg{Ablation-CAM}
	\addplot[mark=*,red] table{fig/eval/adaiag/maskin_GradCAM_ai.txt}; \leg{Grad-CAM}
	\addplot[mark=*,green] table{fig/eval/adaiag/maskin_GradCAMPlusPlus_ai.txt}; \leg{Grad-CAM++}
	\addplot[mark=*,black] table{fig/eval/adaiag/maskin_XGradCAM_ai.txt}; \leg{XGrad-CAM}
	\addplot[mark=*,brown] table{fig/eval/adaiag/maskin_ScoreCAM_ai.txt}; \leg{Score-CAM}
	\addplot[mark=*,orange] table{fig/eval/adaiag/maskin_versionP0_ai.txt}; \leg{Opti-CAM}
\end{axis}
\end{tikzpicture}
}
\end{tabular}
\caption{Effect of \emph{selectivity} (raising element-wise to exponent $\alpha$) of saliency maps on classification performance. $\AD$/$\AI$: average drop/increase~\citep{chattopadhay2018grad}; $\AG$: average gain (ours); $\downarrow$ / $\uparrow$: lower / higher is better.}
\label{fig:aiadag-alpha}
\end{figure*}

Results in terms of $\AD, \AG, \AI$ are shown in \autoref{fig:aiadag-alpha}, averaged over $1,000$ ImageNet images. We observe that $\AD$ stays near zero for Opti-CAM for $\alpha < 2$, while it increases linearly with $\alpha$ for the other methods. The $\AG$ and $\AI$ of Opti-CAM has a strong peak at $\alpha = 1$, \ie for the original saliency maps. The other methods are less sensitive and their $\AI$ performance is not optimal at $\alpha = 1$.


\subsection{Opti-CAM components}

\paragraph{Objective function}

We consider more alternative definitions of the objective function $F^c_\ell$, taking into account not only the regions inside the saliency maps (In) but also their complement, outside (Out). In particular, relative to \Fdef, we define \MIODref as
\begin{equation}
	F^c_\ell(\vx; \vu) \defn g_c(f(\vx \odot \Vs)) - g_c(f(\vx \odot (1-\Vs))),
\label{eq:mi-dref}
\end{equation}
where $\Vs \defn n(\up(S_\ell(\vx; \vu)))$ for brevity. Similarly, relative to \Fref, we define \MIOFref as
\begin{equation}
\begin{split}
	F^c_\ell(\vx; \vu) \defn
		- \abs{g_c(f(\vx)) - g_c(f(\vx \odot \Vs))} \\
		+ \abs{g_c(f(\vx)) - g_c(f(\vx \odot (1-\Vs)))}.
\end{split}
\label{eq:mi-ref}
\end{equation}

According to \autoref{tab:ablate-loss}, \MIODref performs great on AD and AI but worse on AG, while \MIOFref is worse on all metrics. Therefore, including the complementary of the saliency map is not beneficial.

\begin{table}[t]
\centering
\footnotesize
\setlength{\tabcolsep}{1pt}
\begin{tabular}{lcrrrrr} \toprule
{\Th{Method}} & {$F^c_\ell$}& {$\AD\!\downarrow$}& {$\AG\!\uparrow$} & {$\AI\!\uparrow$}  \\ \midrule
Fake-CAM~\citep{poppi2021revisiting}              &          &     0.5  &      0.7  &     42.1  \\ \midrule
Grad-CAM~\citep{selvaraju2017grad}                &          &    15.0  &     15.3  &     40.4  \\
Grad-CAM++~\cite{chattopadhay2018grad}            &          &    16.5  &     10.6  &     35.2  \\
Score-CAM~\citep{wang2020score}                   &          &    12.5  &     16.1  &     42.6  \\
Ablation-CAM~\citep{ramaswamy2020ablation}        &          &    15.1  &     13.5  &     39.9  \\
XGrad-CAM~\citep{fu2020axiom}                     &          &    14.3  &     15.1  &     42.1  \\
Layer-CAM~\citep{jiang2021layercam}               &          &    49.2  &      2.7  &     12.7  \\
ExPerturbation~\citep{fong2019understanding}      &          &    43.8  &      7.1  &     18.9  \\
\midrule
\mr{4}{Opti-CAM}
& \Fdef~\eq{obj}        &1.4          & \tb{66.3} &92.5\\
& \Fref~\eq{ref}        &7.1&18.5&54.9     \\
& \MIODref~\eq{mi-dref} &\tb{0.2}&5.5&\tb{99.7}\\
& \MIOFref~\eq{mi-ref}  &25.9&7.6&42.6\\
\bottomrule
\end{tabular}
\caption{\emph{Ablation study on objective function} using VGG16 on 1000 images of ImageNet validation set.
Choices for objective function $F^c_\ell$: \Fdef:~\eq{obj}; \Fref:~\eq{ref}; \MIODref:~\eq{mi-dref}; \MIOFref:~\eq{mi-ref}.
Choice for normalization function $n$: Range~\eq{n-rng}. Iterations: 50.
$\AD$/$\AI$: average drop/increase~\citep{chattopadhay2018grad}; $\AG$: average gain (ours); $\downarrow$ / $\uparrow$: lower / higher is better.}
\label{tab:ablate-loss}
\end{table}

\begin{table}[ht!]
\footnotesize
\centering
\setlength{\tabcolsep}{6pt}
\begin{tabular}{crrr} \toprule
{\Th{Layer}} & {$\AD\!\downarrow$}& {$\AG\!\uparrow$} & {$\AI\!\uparrow$}\\ \midrule
42 &1.4&66.0&92.5\\
36 &1.7&66.1&90.3\\
32 &2.8&61.3&81.6\\
29 &1.6&78.0&93.9\\
26 &1.7&80.1&93.7\\
22 &3.3&68.8&84.8\\
19 &2.9&67.3&84.9\\
16 &2.3&72.4&89.1\\
12 &4.1&61.9&82.4\\
9  &4.3&44.2&71.9\\
6  &13.5&23.5&50.2\\ \bottomrule
\end{tabular}
\caption{\emph{Layer ablation} on $1,000$ images from ImageNet validation set, using various layers of VGG16. The last convolutional layer before max pooling is chosen as our default layer (layer 42). $\AD$/$\AI$: average drop/increase~\citep{chattopadhay2018grad}; $\AG$: average gain (ours); $\downarrow$ / $\uparrow$: lower / higher is better.}
\label{tab:layer}
\end{table}

\paragraph{Layers}

\autoref{tab:layer} shows how the performance of Opti-CAM, in terms of AD/AI/AG, depends on the layer $\ell$ of the VGG16 network used to compute the saliency map $S^c_\ell$~\eq{v-sal}. We can see that the layers 26, 29, and 42 are all competitive. We choose the last convolutional layer (42) to be compatible with the other CAM methods \citep{zhou2016learning,selvaraju2017grad,chattopadhay2018grad,wang2020score}.

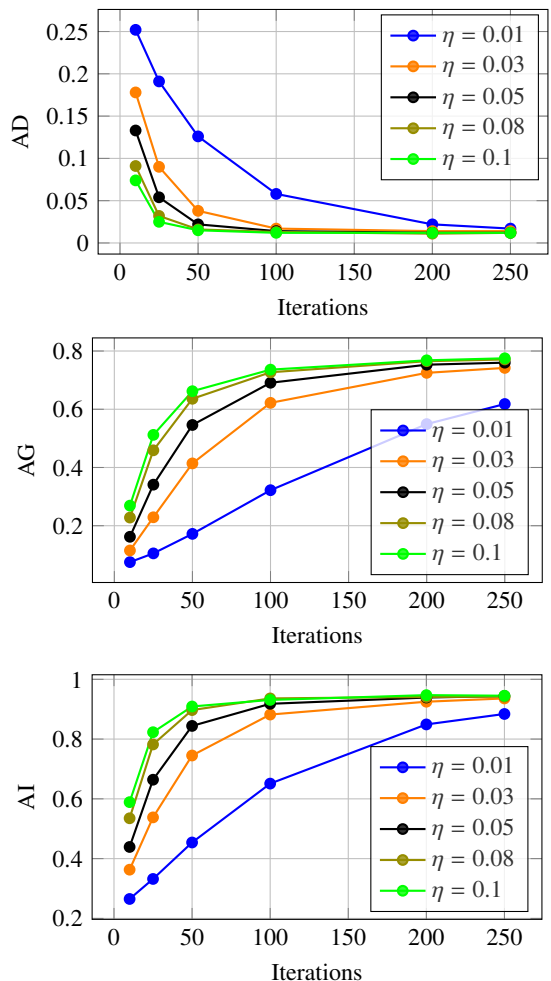
\begin{figure}[hpt]
\centering
\begin{tabular}{c}
\extfig{ab-ad}{
\begin{tikzpicture}
\begin{axis}[
    yticklabel style={
            /pgf/number format/fixed,
            /pgf/number format/precision=5
    },
    scaled y ticks=false,
	height=4.8cm,
	width=7.5cm,
	xlabel={Iterations},
	ylabel={AD},
]
	\addplot[mark=*,blue] table[x index=0, y index=1]{\plotAD};\leg{$\eta=0.01$}
	\addplot[mark=*,orange] table[x index=0, y index=2]{\plotAD};\leg{$\eta=0.03$}
	\addplot[mark=*,black] table[x index=0, y index=3]{\plotAD};\leg{$\eta=0.05$}
	\addplot[mark=*,olive] table[x index=0, y index=4]{\plotAD};\leg{$\eta=0.08$}
	\addplot[mark=*,green] table[x index=0, y index=5]{\plotAD};\leg{$\eta=0.1$}
\end{axis}
\end{tikzpicture}
}\\
\extfig{ab-ag}{
\begin{tikzpicture}
\begin{axis}[
    yticklabel style={
            /pgf/number format/fixed,
            /pgf/number format/precision=5
    },
    scaled y ticks=false,
	height=4.8cm,
	width=7.5cm,
	xlabel={Iterations},
	ylabel={AG},
	legend pos=south east,
]
	\addplot[mark=*,blue] table[x index=0, y index=1]{\plotAG};\leg{$\eta=0.01$}
	\addplot[mark=*,orange] table[x index=0, y index=2]{\plotAG};\leg{$\eta=0.03$}
	\addplot[mark=*,black] table[x index=0, y index=3]{\plotAG};\leg{$\eta=0.05$}
	\addplot[mark=*,olive] table[x index=0, y index=4]{\plotAG};\leg{$\eta=0.08$}
	\addplot[mark=*,green] table[x index=0, y index=5]{\plotAG};\leg{$\eta=0.1$}
\end{axis}
\end{tikzpicture}
}\\
\extfig{ab-ai}{
\begin{tikzpicture}
\begin{axis}[
	height=4.8cm,
	width=7.5cm,
	xlabel={Iterations},
	ylabel={AI},
	legend pos=south east,
]
	\addplot[mark=*,blue] table[x index=0, y index=1]{\plotAI};\leg{$\eta=0.01$}
	\addplot[mark=*,orange] table[x index=0, y index=2]{\plotAI};\leg{$\eta=0.03$}
	\addplot[mark=*,black] table[x index=0, y index=3]{\plotAI};\leg{$\eta=0.05$}
	\addplot[mark=*,olive] table[x index=0, y index=4]{\plotAI};\leg{$\eta=0.08$}
	\addplot[mark=*,green] table[x index=0, y index=5]{\plotAI};\leg{$\eta=0.1$}
\end{axis}
\end{tikzpicture}
}\\
\end{tabular}
\caption{
Classification metrics \vs number of iterations for different learning rates, using VGG-16 on 1000 images of ImageNet. $\AD$/$\AI$: average drop/increase~\citep{chattopadhay2018grad}; $\AG$: average gain (ours); $\downarrow$ / $\uparrow$: lower / higher is better.}
\label{fig:lr-epochs}
\end{figure}

\paragraph{Convergence}

Finally, \autoref{fig:lr-epochs} shows the classification performance of Opti-CAM \vs number of iterations for different learning rates. Optimal performance can be obtained at 100 iterations with learning rate $\eta = 0.1$. We use these settings by default. We note that by using 50 iterations allows us to double the speed at the cost of a 6\% drop of $\AG$ and very small drop of $\AI$ and $\AD$.

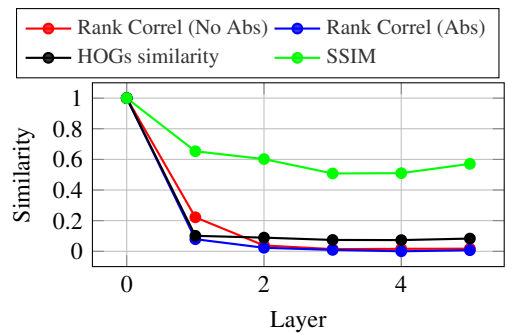
\begin{figure}[htpb]
\centering
\ref*{sanity_leg}

\extfig{scanity-check}{
\begin{tikzpicture}
\begin{axis}[
	height=4cm,
	width=7cm,
	xlabel={Layer},
	ylabel={Similarity},
	legend columns=2,
    legend to name=sanity_leg,
    legend style={font=\footnotesize}
]
	\addplot[mark=*,red] coordinates{(0,1.0)(1,0.222)(2,0.038)(3,0.014)(4,0.016)(5,0.017)};   \leg{Rank Correl (No Abs)}
	\addplot[mark=*,blue] coordinates{(0,1.0)(1,0.079)(2,0.023)(3,0.009)(4,0.000)(5,0.007)}; \leg{Rank Correl (Abs)}
	\addplot[mark=*,black] coordinates{(0,1.0)(1,0.101)(2,0.089)(3,0.074)(4,0.073)(5,0.083)}; \leg{HOGs similarity}
	\addplot[mark=*,green] coordinates{(0,1.0)(1,0.653)(2,0.602)(3,0.508)(4,0.510)(5,0.571)}; \leg{SSIM}
\end{axis}
\end{tikzpicture}
}
\caption{\emph{Sanity check} of Opti-CAM on $1,000$ images of ImageNet validation set using ResNet50. Similarity between saliency maps by original and randomized network, where layers are progressively replaced by random ones.}
\label{fig:sanity}
\end{figure}

\begin{figure}[htpb]
\newcommand{\sizeP}{.12}
\newcommand{\sizeS}{.15}
\newcommand{\hh}{.175\textwidth}
\newcommand{\ww}{.200\textwidth}
\centering
\small
\setlength{\tabcolsep}{3pt}
\begin{tabular}{cccccc}
Original & layer 1 & layer 2 & layer 3 & layer 4 & layer 5 \\
\fig[\sizeS]{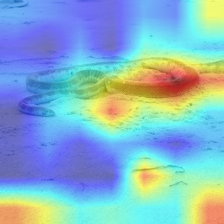} &
\fig[\sizeS]{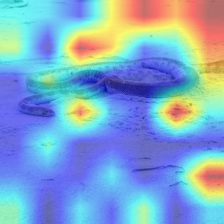} &
\fig[\sizeS]{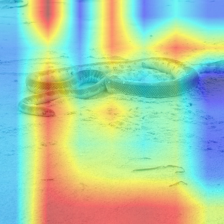} &
\fig[\sizeS]{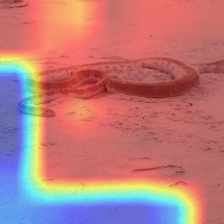} &
\fig[\sizeS]{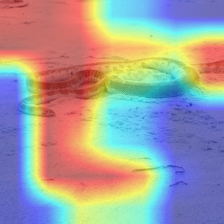} &
\fig[\sizeS]{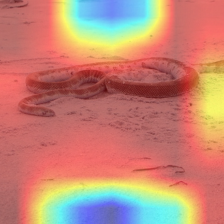} \\
\fig[\sizeS]{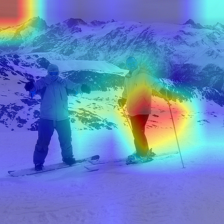} &
\fig[\sizeS]{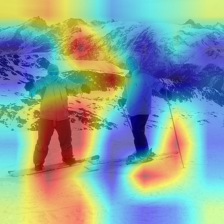} &
\fig[\sizeS]{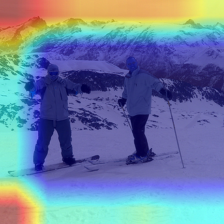} &
\fig[\sizeS]{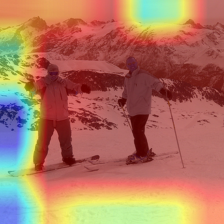} &
\fig[\sizeS]{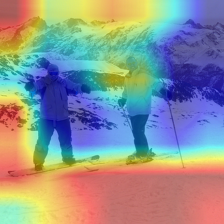} &
\fig[\sizeS]{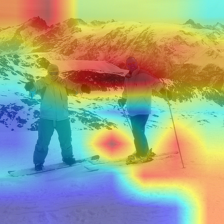} \\
\end{tabular}
\caption{\emph{Sanity check visualization} of Opti-CAM on two images of ImageNet validation set using ResNet50. First column: Opti-CAM saliency maps for the original network; remaining columns: Opti-CAM saliency maps where layers are progressively replaced by random ones.}
\label{fig:sanity-vis}
\end{figure}

\section{Sanity check}
\label{sec:sanity-check}

We use the model parameter randomization test proposed by~\citep{adebayosanity}. This test compares the saliency maps generated by a trained model with the ones generated by a partially randomly initialized network of the same architecture. In particular, we choose 5 layers of ResNet50 and we progressively replace them by random ones so that we have 6 different models with different amount of random parameters. The saliency maps are generated for the small subset of ImageNet validation set, as in the ablation study.

Following~\citep{adebayosanity}, we compute a number of similarity metrics between these saliency maps generated by the original and the randomized network, including Rank Correlation with/without absolute values, HOGs similarity, and SSIM. The results are shown in \autoref{fig:sanity} (saliency map similarity measurements) and \autoref{fig:sanity-vis} (saliency map visualizations). Our method passes the sanity check, as it is very sensitive to changes in the model parameters.

\begin{table}[htbp]
\centering
\footnotesize
\setlength{\tabcolsep}{2pt}
\begin{tabular}{lrrrr|rrrr} \toprule
\mr{2}{\Th{Method}}                                & \mc{4}{\Th{ResNet50}} & \mc{4}{\Th{VGG16}} \\ \cmidrule{2-9}
                                                   & {$\AD\!\downarrow$} & {$\AG\!\uparrow$} & {$\AI\!\uparrow$} & \mc{1}{T} & {$\AD\!\downarrow$} & {$\AG\!\uparrow$} & {$\AI\!\uparrow$} & \mc{1}{T} \\ \midrule
Fake-CAM~\citep{poppi2021revisiting}               &0.9&0.7&47.4&0.00&0.5&0.3&47.7&0.00  \\ \midrule
Grad-CAM~\citep{selvaraju2017grad}       & 36.4      &5.5& 27.0      &0.03     & 41.6     &3.3 & 25.2       &0.02     \\
Grad-CAM++~\cite{chattopadhay2018grad}    & 37.6    & 4.9 & 24.0       &0.04    & 46.3     &2.0 & 19.0        &0.02    \\
Score-CAM~\citep{wang2020score}           & 28.8     &8.8 & 33.6       &20.47& 39.3    & 3.5 & 24.6       &3.08     \\
Ablation-CAM~\citep{ramaswamy2020ablation}   & 36.6      &5.1& 25.6      &18.49 & 41.8     & 2.9& 24.0       &2.95    \\
XGrad-CAM~\citep{fu2020axiom}            & 36.4     &5.5 & 27.0       &0.03   & 40.6     &3.4 & 25.8       &0.02   \\
Layer-CAM~\citep{jiang2021layercam} &42.6&4.2&19.2&0.02&82.1&0.3&6.9&0.01   \\
ExPerturbation~\citep{fong2019understanding} &51.2&6.9&26.1&15.67&50.1&4.4&24.5&9.10  \\
\rowcolor{cyan!10}
Opti-CAM (ours)                          &\tb{2.0}&\tb{49.4}&\tb{91.2}&3.94&\tb{1.5}&\tb{52.7}&\tb{92.1}&3.95\\
\bottomrule
\end{tabular}
\caption{
\emph{Classification metrics} on ImageNet validation set, without input normalization. AD/AI: average drop/increase~\citep{chattopadhay2018grad}; $\AG$: average gain (ours); $\downarrow$ / $\uparrow$: lower / higher is better. T: Average time (sec) per batch of 8 images. Bold: best, excluding Fake-CAM.}
\label{tab:norm-imagenet}
\end{table}

\section{Results without input normalization}
\label{sec:without-norm}

It is standard that images are normalized to zero mean and unit standard deviation before feeding them to a network, because this is how networks are trained. For example, for ImageNet images, we subtract the mean vector $[0.485,0.456,0.406]$ and divide channel-wise by standard deviation $[0.229,0.224,0.225]$. By doing so however, we cannot reproduce the results published for several baseline methods; rather, all results are improved dramatically. We can obtain results similar to published ones by \emph{not} normalizing, thus we speculate that authors of related work do not normalize images. This is also suggested by our attempts to communicate with the authors.

We believe normalization is important and we include it in all our experiments. For reference and to allow for comparison with published results, we provide results without normalization in \autoref{tab:norm-imagenet} that correspond to \autoref{tab:imagenet-cnn}. Finally, code is provided to allow for reproduction and verification of our results.

\begin{figure*}
\newcommand{\sizeS}{.09}
\centering
\small
\setlength{\tabcolsep}{2pt}
\begin{tabular}{c cccc cccc}
     & \mc{4}{VGG16} & \mc{4}{ResNet50}\\
	Input image  &  Grad-CAM  & G-CAM++ & Score-CAM & Opti-CAM  &Grad-CAM  & G-CAM++ & Score-CAM & Opti-CAM \\
	\fig[\sizeS]{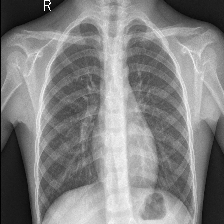} &
	\fig[\sizeS]{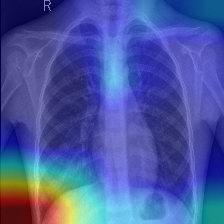} &\fig[\sizeS]{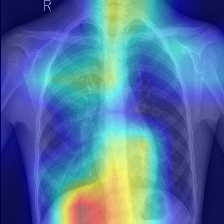} &
	\fig[\sizeS]{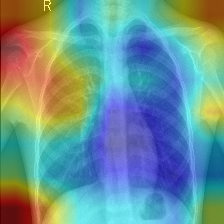} &\fig[\sizeS]{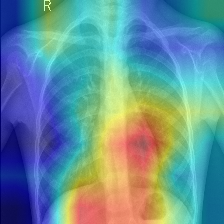} &
	\fig[\sizeS]{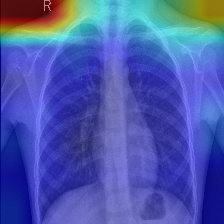} &\fig[\sizeS]{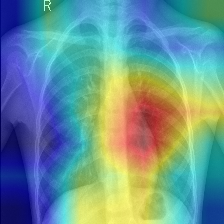} &
	\fig[\sizeS]{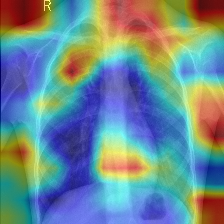} & \fig[\sizeS]{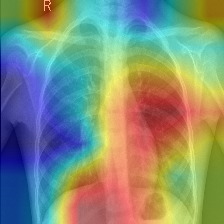} \\
 
\fig[\sizeS]{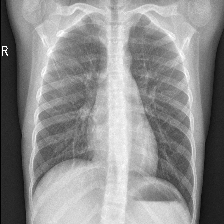} &
	\fig[\sizeS]{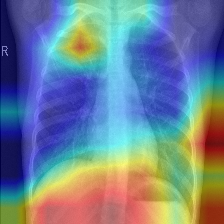} &\fig[\sizeS]{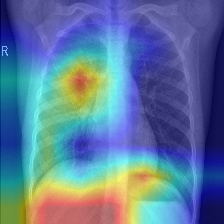} &
	\fig[\sizeS]{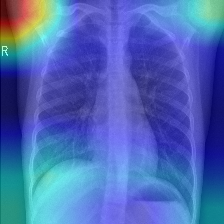} &\fig[\sizeS]{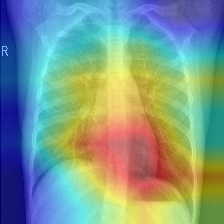} &
	\fig[\sizeS]{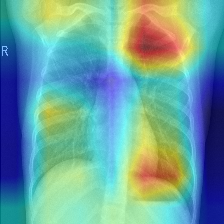} &\fig[\sizeS]{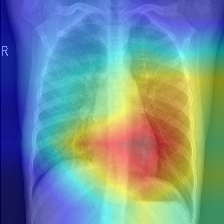} &
	\fig[\sizeS]{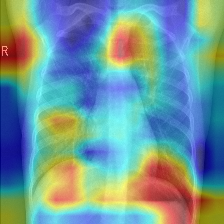} & \fig[\sizeS]{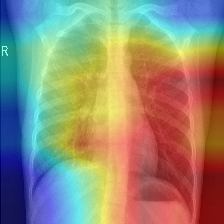} \\
 
\fig[\sizeS]{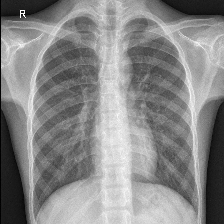} &
	\fig[\sizeS]{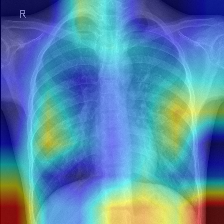} &\fig[\sizeS]{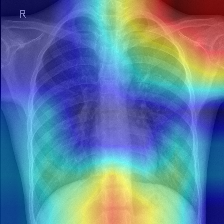} &
	\fig[\sizeS]{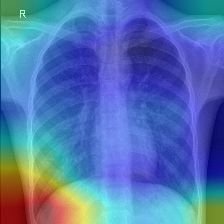} &\fig[\sizeS]{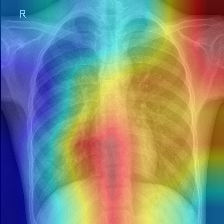} &
	\fig[\sizeS]{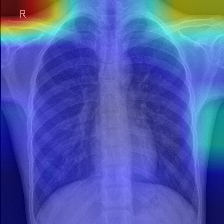} &\fig[\sizeS]{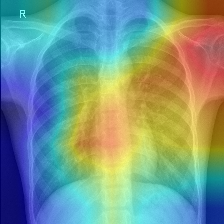} &
	\fig[\sizeS]{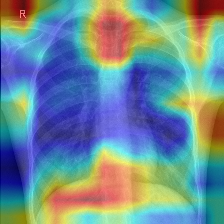} & \fig[\sizeS]{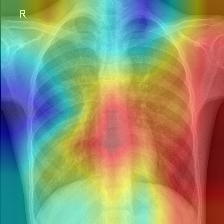} \\
 
\fig[\sizeS]{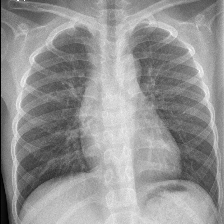} &
	\fig[\sizeS]{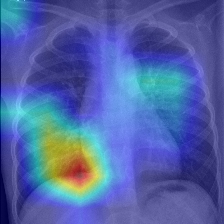} &\fig[\sizeS]{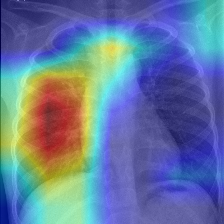} &
	\fig[\sizeS]{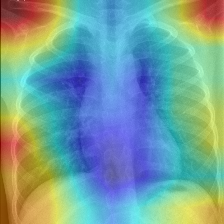} &\fig[\sizeS]{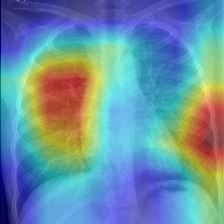} &
	\fig[\sizeS]{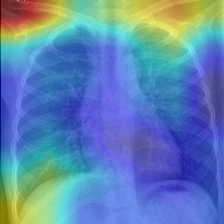} &\fig[\sizeS]{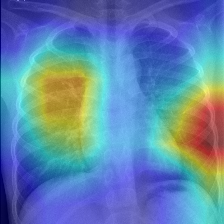} &
	\fig[\sizeS]{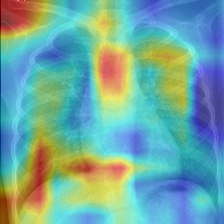} & \fig[\sizeS]{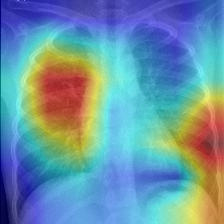} \\
 
	\fig[\sizeS]{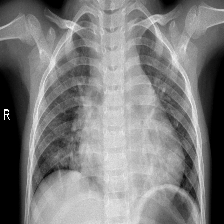} &
	\fig[\sizeS]{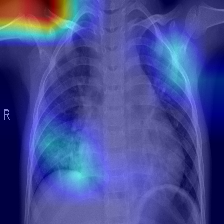} &\fig[\sizeS]{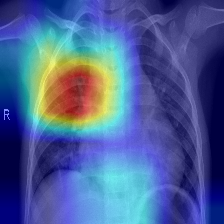} &
	\fig[\sizeS]{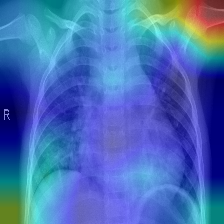} &\fig[\sizeS]{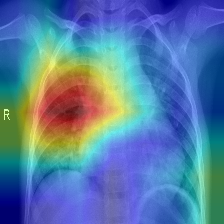} &
	\fig[\sizeS]{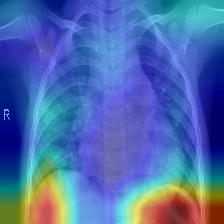} &\fig[\sizeS]{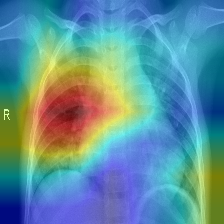} &
	\fig[\sizeS]{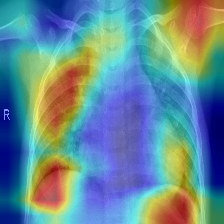} & \fig[\sizeS]{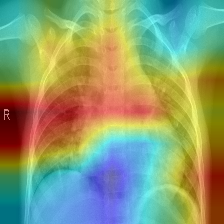} \\

 \fig[\sizeS]{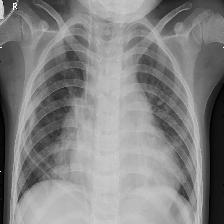} &
	\fig[\sizeS]{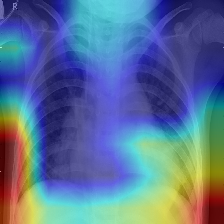} &\fig[\sizeS]{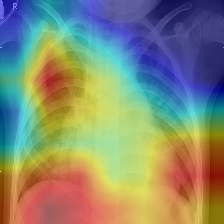} &
	\fig[\sizeS]{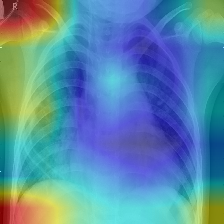} &\fig[\sizeS]{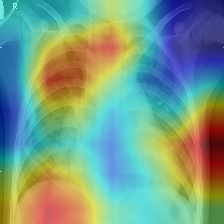} &
	\fig[\sizeS]{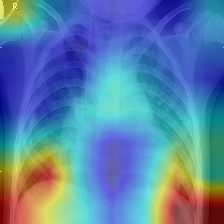} &\fig[\sizeS]{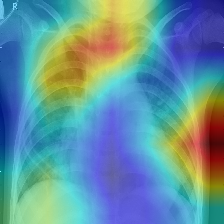} &
	\fig[\sizeS]{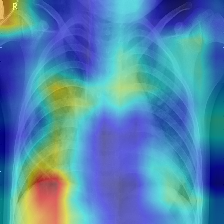} & \fig[\sizeS]{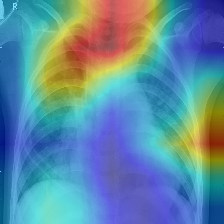} \\

 \fig[\sizeS]{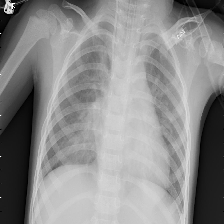} &
	\fig[\sizeS]{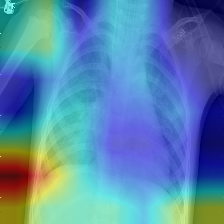} &\fig[\sizeS]{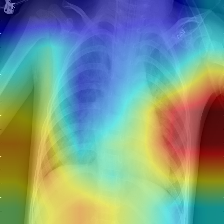} &
	\fig[\sizeS]{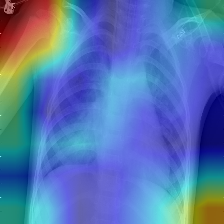} &\fig[\sizeS]{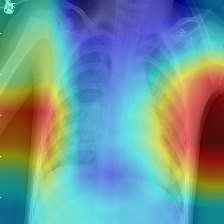} &
	\fig[\sizeS]{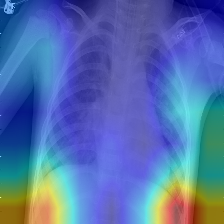} &\fig[\sizeS]{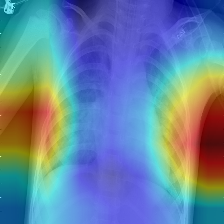} &
	\fig[\sizeS]{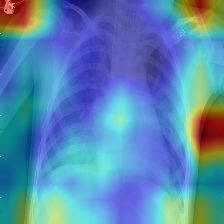} & \fig[\sizeS]{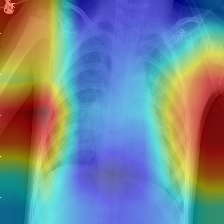} \\

 \fig[\sizeS]{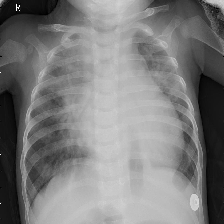} &
	\fig[\sizeS]{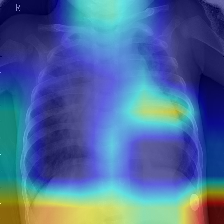} &\fig[\sizeS]{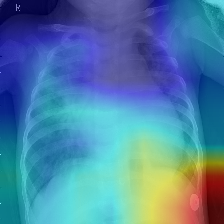} &
	\fig[\sizeS]{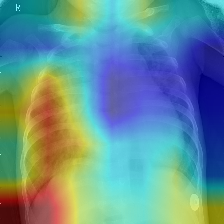} &\fig[\sizeS]{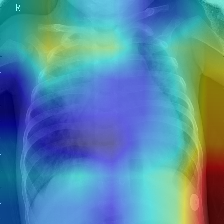} &
	\fig[\sizeS]{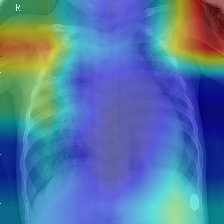} &\fig[\sizeS]{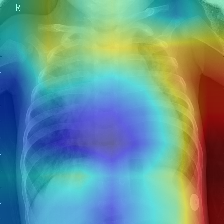} &
	\fig[\sizeS]{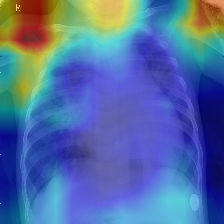} & \fig[\sizeS]{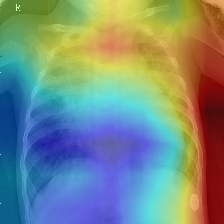} \\
\end{tabular}
\caption{Saliency maps obtained from Chest X-ray images.}
\label{fig:vis-chest-resnet}
\end{figure*}

\begin{figure*}
\newcommand{\sizeS}{.10}
\centering
\small
\setlength{\tabcolsep}{2pt}
\begin{tabular}{c cccc cccc}
     & \mc{4}{VGG16} & \mc{4}{ResNet50}\\
	Input image  &  Grad-CAM  & G-CAM++ & Score-CAM & Opti-CAM  &Grad-CAM  & G-CAM++ & Score-CAM & Opti-CAM \\
	\fig[\sizeS]{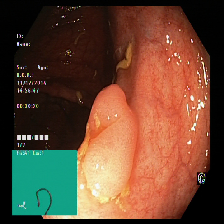} &
	\fig[\sizeS]{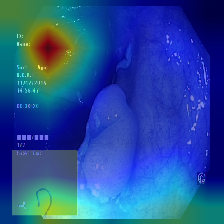} &
	\fig[\sizeS]{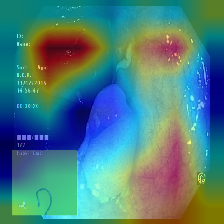} &
	\fig[\sizeS]{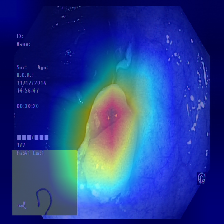} &
	\fig[\sizeS]{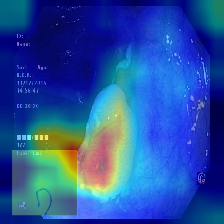} &
		\fig[\sizeS]{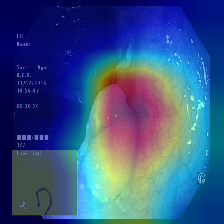} &
	\fig[\sizeS]{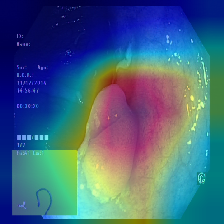} &
	\fig[\sizeS]{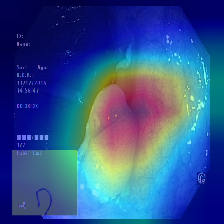} &
	\fig[\sizeS]{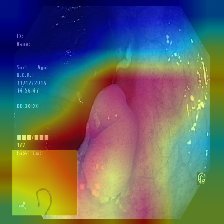}
	\\
	\fig[\sizeS]{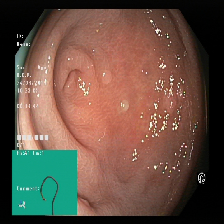} &
	\fig[\sizeS]{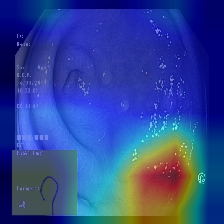} &
	\fig[\sizeS]{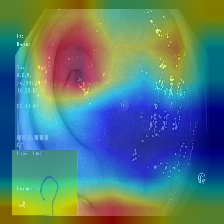} &
	\fig[\sizeS]{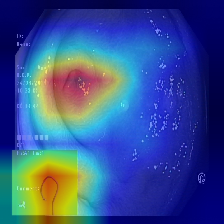} &
	\fig[\sizeS]{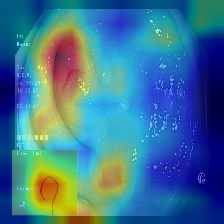} &
	\fig[\sizeS]{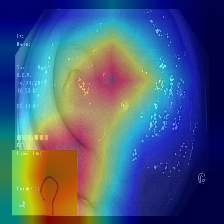} &
	\fig[\sizeS]{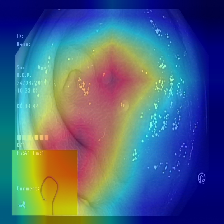} &
	\fig[\sizeS]{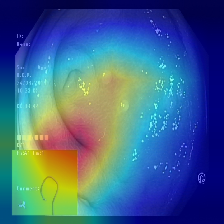} &
	\fig[\sizeS]{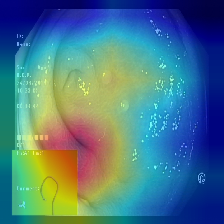} 
	
	\\
	\fig[\sizeS]{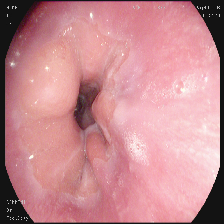} &
	\fig[\sizeS]{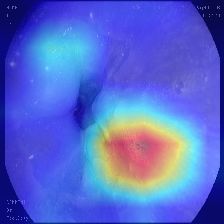} &
	\fig[\sizeS]{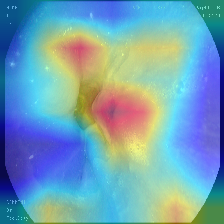} &
	\fig[\sizeS]{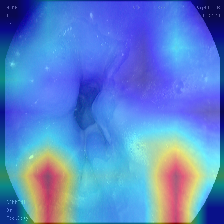} &
	\fig[\sizeS]{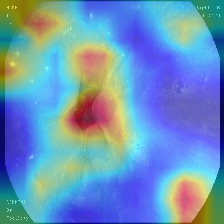} &
	\fig[\sizeS]{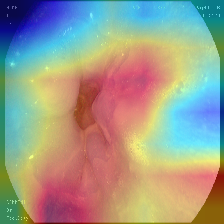} &
	\fig[\sizeS]{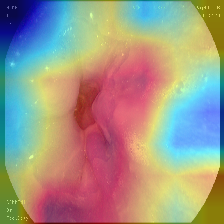} &
	\fig[\sizeS]{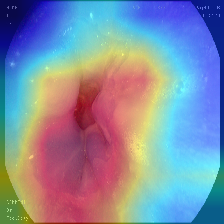} &
	\fig[\sizeS]{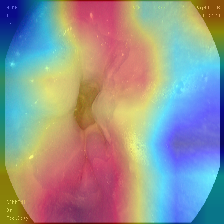}\\

	\fig[\sizeS]{medical/kvasir_Resnet50_GradCAM_3_img.png} &
	\fig[\sizeS]{medical/kvasir_VGG16_GradCAM_3_vis.png} &
	\fig[\sizeS]{medical/kvasir_VGG16_GradCAMPlusPlus_3_vis.png} &
	\fig[\sizeS]{medical/kvasir_VGG16_ScoreCAM_3_vis.png} &
	\fig[\sizeS]{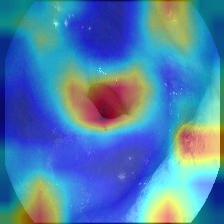} &
	\fig[\sizeS]{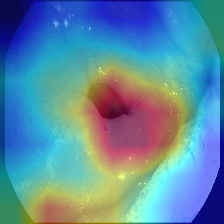} &
	\fig[\sizeS]{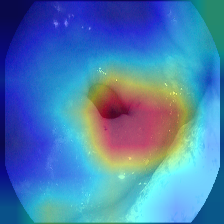} &
	\fig[\sizeS]{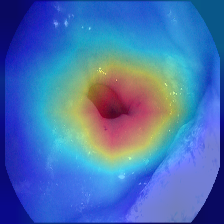} &
	\fig[\sizeS]{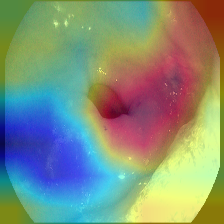}
	\\
	\fig[\sizeS]{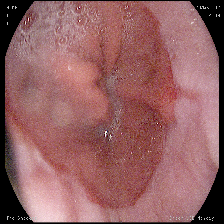} &
	\fig[\sizeS]{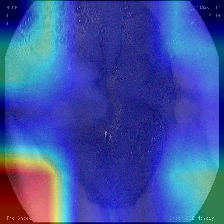} &
	\fig[\sizeS]{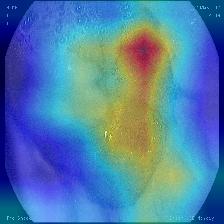} &
	\fig[\sizeS]{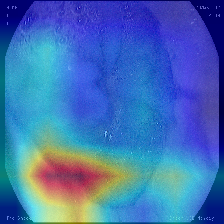} &
	\fig[\sizeS]{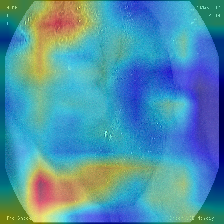} &
	\fig[\sizeS]{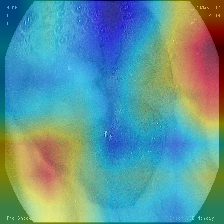} &
	\fig[\sizeS]{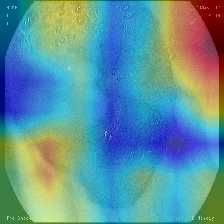} &
	\fig[\sizeS]{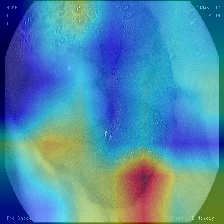} &
	\fig[\sizeS]{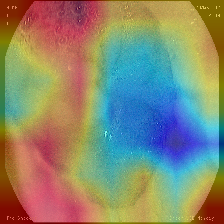}
	\\
	\fig[\sizeS]{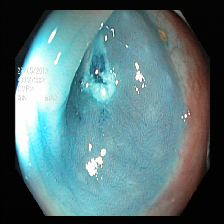} &
	\fig[\sizeS]{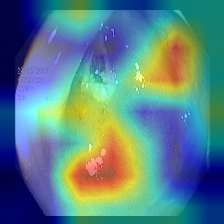} &
	\fig[\sizeS]{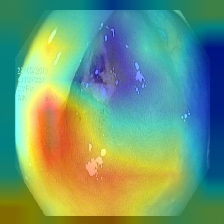} &
	\fig[\sizeS]{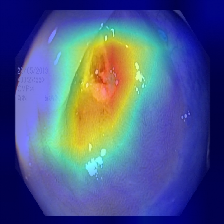} &
	\fig[\sizeS]{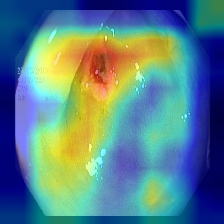}  &
	\fig[\sizeS]{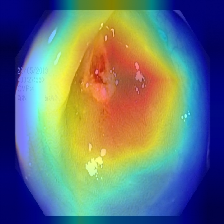} &
	\fig[\sizeS]{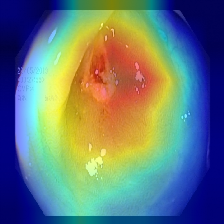} &
	\fig[\sizeS]{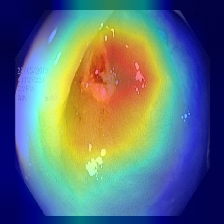} &
	\fig[\sizeS]{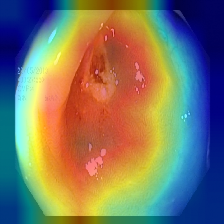} 
	\\
	\fig[\sizeS]{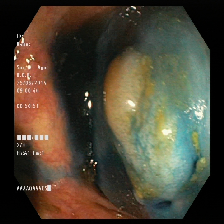} &
	\fig[\sizeS]{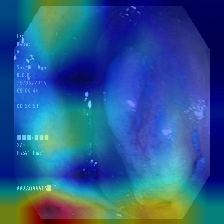} &
	\fig[\sizeS]{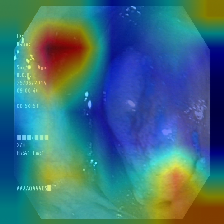} &
	\fig[\sizeS]{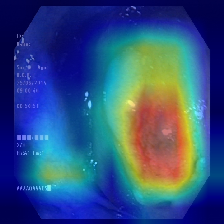} &
	\fig[\sizeS]{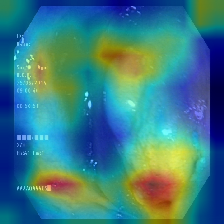} &
	\fig[\sizeS]{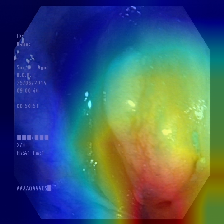} &
	\fig[\sizeS]{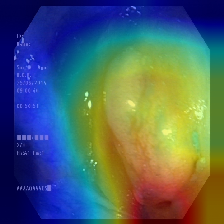} &
	\fig[\sizeS]{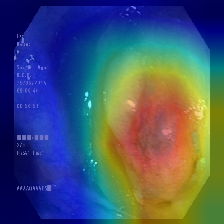} &
	\fig[\sizeS]{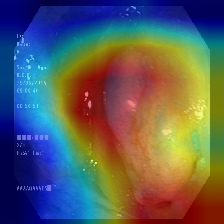} 
	\\
	\fig[\sizeS]{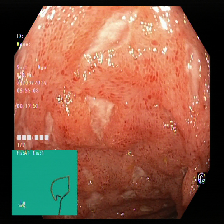} &
	\fig[\sizeS]{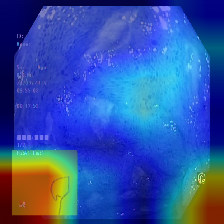} &
	\fig[\sizeS]{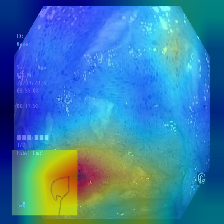} &
	\fig[\sizeS]{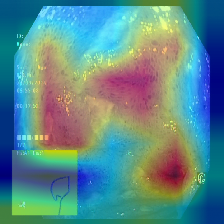} &
	\fig[\sizeS]{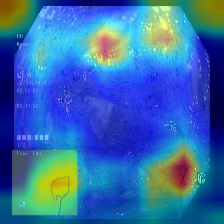} &
	\fig[\sizeS]{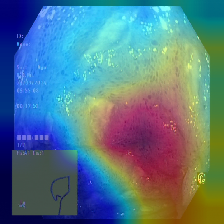} &
	\fig[\sizeS]{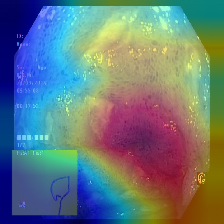} &
	\fig[\sizeS]{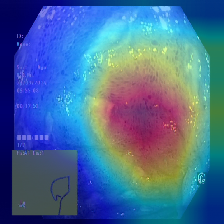} &
	\fig[\sizeS]{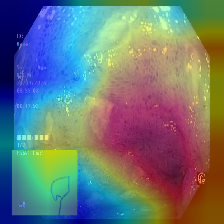} \\
\end{tabular}
\caption{Saliency maps obtained for KVASIR images.}
\label{fig:vis-kvasir-resnet}
\end{figure*}

\begin{figure*}[t]
\newcommand{\sizeP}{.12}
\newcommand{\sizeS}{.12}
\newcommand{\hh}{.175\textwidth}
\newcommand{\ww}{.200\textwidth}
\tiny
\centering
\setlength{\tabcolsep}{3pt}
\begin{tabular}{ccccccc}
	Input image  &  Grad-CAM  & G-CAM++ & Score-CAM & Ablation-CAM & XG-CAM & Opti-CAM (ours) \\
        \includegraphics[trim={12mm 14mm 12mm 14mm},clip, width=\sizeP\textwidth]{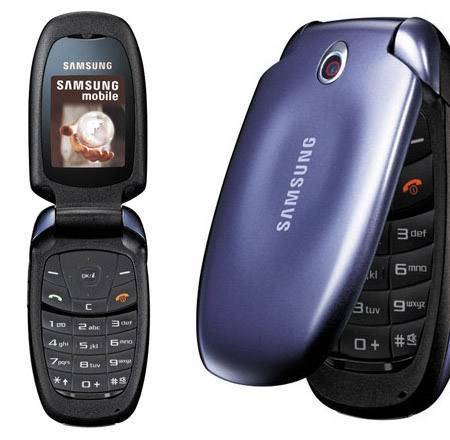}&
	\fig[\sizeS]{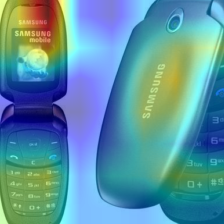} &
	\fig[\sizeS]{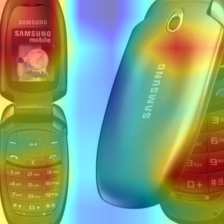} &
	\fig[\sizeS]{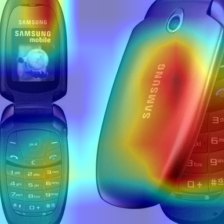} &
	\fig[\sizeS]{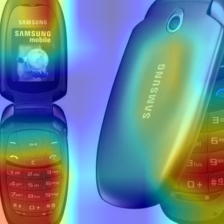} &
	\fig[\sizeS]{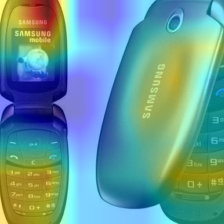} & 
	\fig[\sizeS]{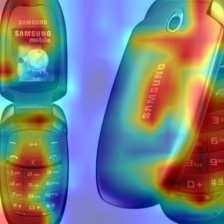}  \\
	Cellphone &&&&&& \\
	\fig[\sizeS]{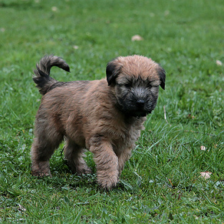}&
	\fig[\sizeS]{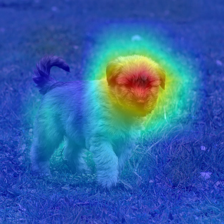} &
	\fig[\sizeS]{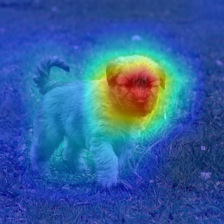} &
	\fig[\sizeS]{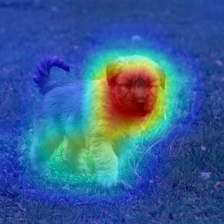} &
	\fig[\sizeS]{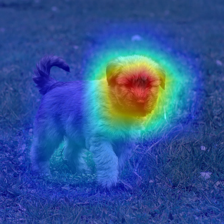} &
	\fig[\sizeS]{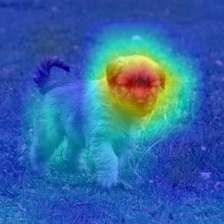} & 
	\fig[\sizeS]{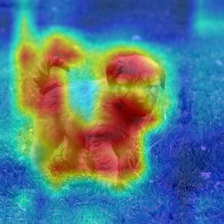}  \\
	Miniature Schnauzer &&&&&& \\
	\includegraphics[trim={2mm 3mm 6mm 1mm},clip, width=\sizeP\textwidth]{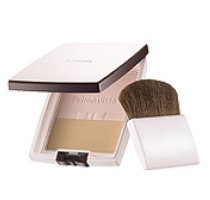}&
	\fig[\sizeS]{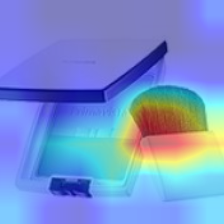} &
	\fig[\sizeS]{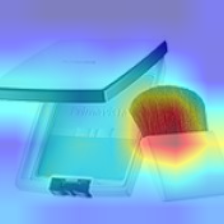} &
	\fig[\sizeS]{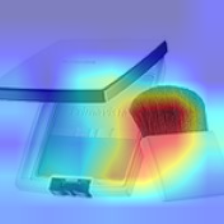} &
	\fig[\sizeS]{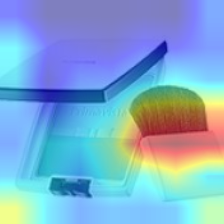} &
	\fig[\sizeS]{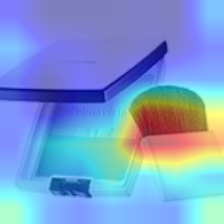} & 
	\fig[\sizeS]{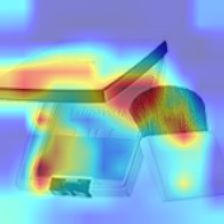}  \\
	Face Powder &&&&&& \\
	\includegraphics[trim={2mm 6mm 2mm 6mm},clip, width=\sizeP\textwidth]{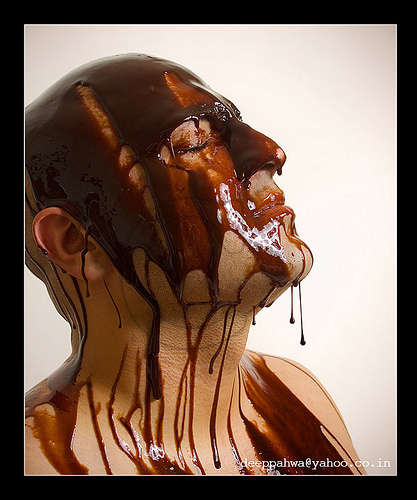}&
	\fig[\sizeS]{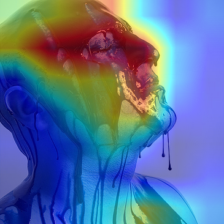} &
	\fig[\sizeS]{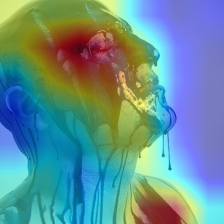} &
	\fig[\sizeS]{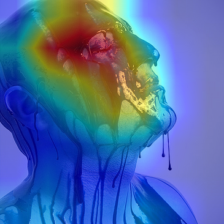} &
	\fig[\sizeS]{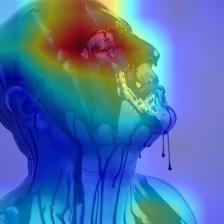} &
	\fig[\sizeS]{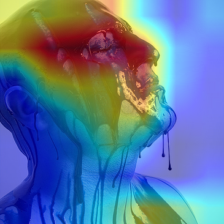} & 
	\fig[\sizeS]{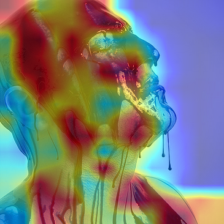}  \\
	Chocolate Sauce &&&&&& \\
	\fig[\sizeS]{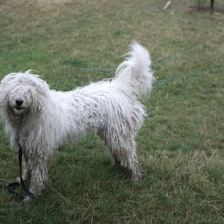}&
	\fig[\sizeS]{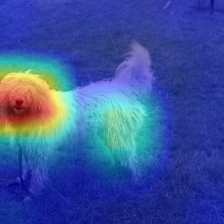} &
	\fig[\sizeS]{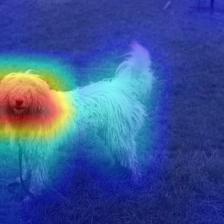} &
	\fig[\sizeS]{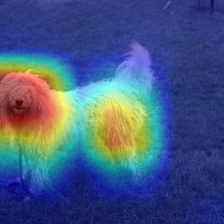} &
	\fig[\sizeS]{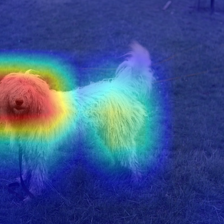} &
	\fig[\sizeS]{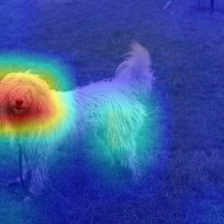} & 
	\fig[\sizeS]{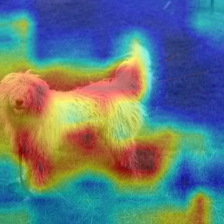}  \\
	Komondor &&&&&& \\
 \includegraphics[trim={14mm 16mm 14mm 12mm},clip, width=\sizeP\textwidth]{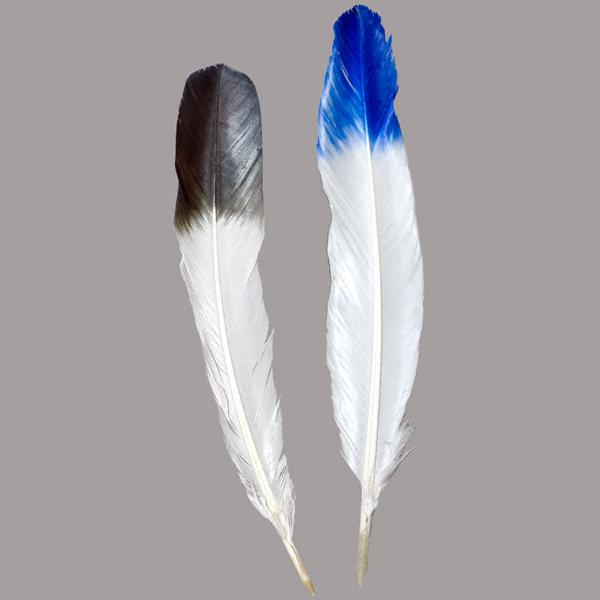}&
	\fig[\sizeS]{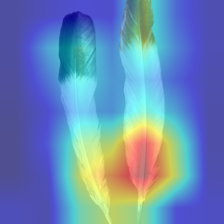} &
	\fig[\sizeS]{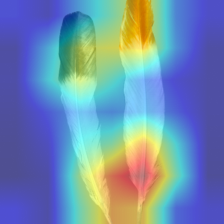} &
	\fig[\sizeS]{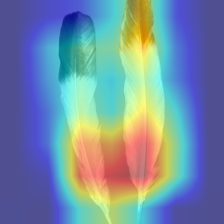} &
	\fig[\sizeS]{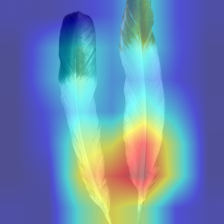} &
	\fig[\sizeS]{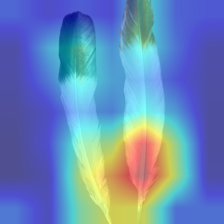} & 
	\fig[\sizeS]{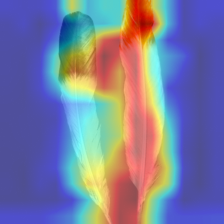}  \\
	Quill &&&&&& \\
 \includegraphics[trim={5mm 14mm 5mm 14mm},clip, width=\sizeP\textwidth]{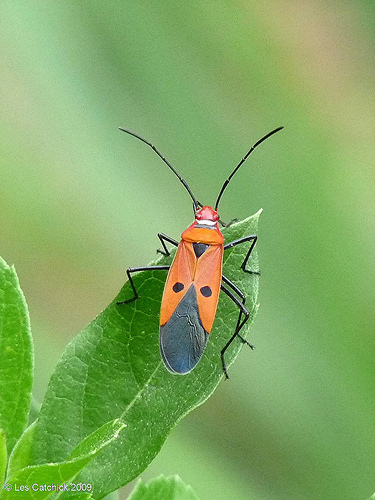}&
	\fig[\sizeS]{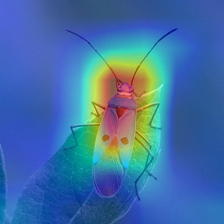} &
	\fig[\sizeS]{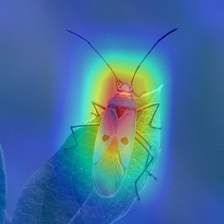} &
	\fig[\sizeS]{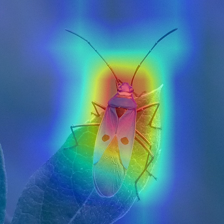} &
	\fig[\sizeS]{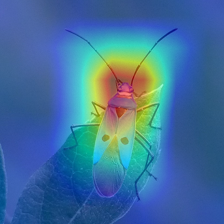} &
	\fig[\sizeS]{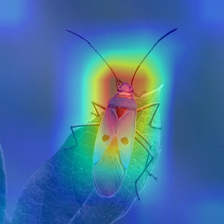} & 
	\fig[\sizeS]{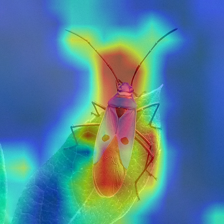}  \\
	Longicorn &&&&&& \\
 \includegraphics[trim={8mm 1mm 8mm 1mm},clip, width=\sizeP\textwidth]{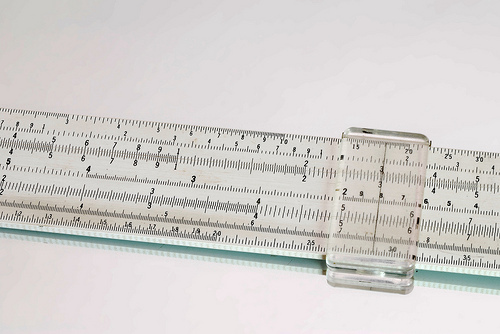}&
	\fig[\sizeS]{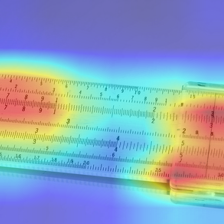} &
	\fig[\sizeS]{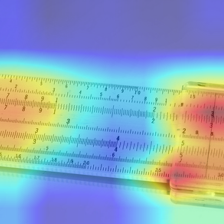} &
	\fig[\sizeS]{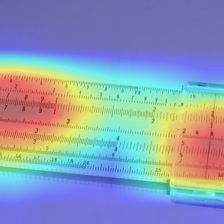} &
	\fig[\sizeS]{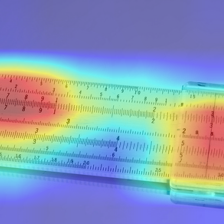} &
	\fig[\sizeS]{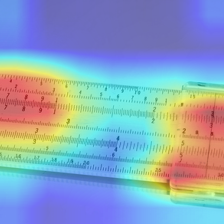} & 
	\fig[\sizeS]{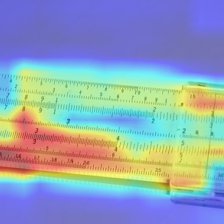}  \\
	Slide Rule &&&&&& \\
\end{tabular}
\caption{Saliency maps obtained from ImageNet example images using different methods on VGG16.}
\label{fig:imagenet-vis-more-vgg}
\end{figure*}

\begin{figure*}[t]
\newcommand{\sizeP}{.12}
\newcommand{\sizeS}{.12}
\newcommand{\hh}{.175\textwidth}
\newcommand{\ww}{.200\textwidth}
\tiny
\centering
\setlength{\tabcolsep}{3pt}
\begin{tabular}{ccccccc}
	Input image  &  Grad-CAM  & G-CAM++ & Score-CAM & Ablation-CAM & XG-CAM & Opti-CAM (ours) \\
        \includegraphics[trim={12mm 14mm 12mm 14mm},clip, width=\sizeP\textwidth]{fig/visual/ILSVRC2012_val_00000089.JPEG}&
	\fig[\sizeS]{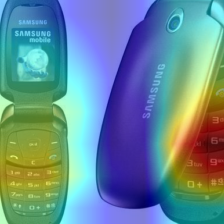} &
	\fig[\sizeS]{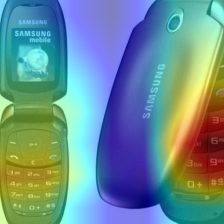} &
	\fig[\sizeS]{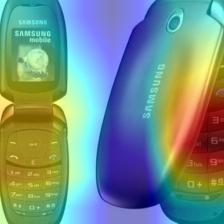} &
	\fig[\sizeS]{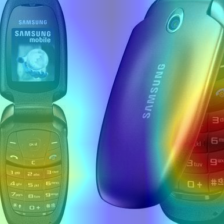} &
	\fig[\sizeS]{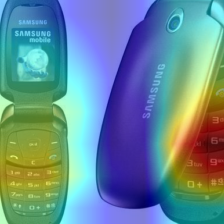} & 
	\fig[\sizeS]{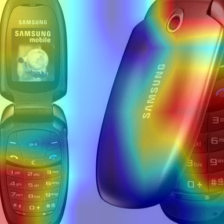}  \\
	Cellphone &&&&&& \\
	\fig[\sizeS]{visual/ILSVRC2012_val_00000748.png}&
	\fig[\sizeS]{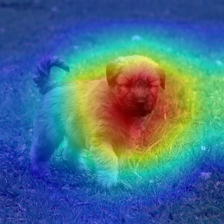} &
	\fig[\sizeS]{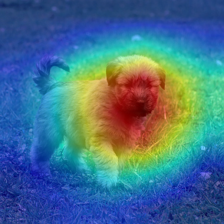} &
	\fig[\sizeS]{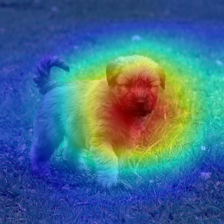} &
	\fig[\sizeS]{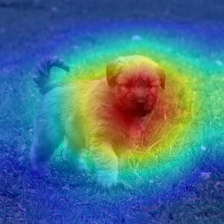} &
	\fig[\sizeS]{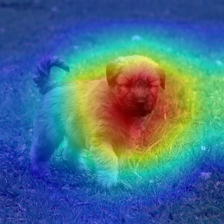} & 
	\fig[\sizeS]{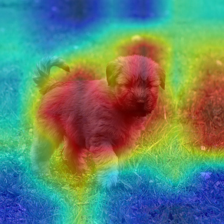}  \\
	Miniature Schnauzer &&&&&& \\
	\includegraphics[trim={2mm 3mm 6mm 1mm},clip, width=\sizeP\textwidth]{fig/visual/ILSVRC2012_val_00000769.JPEG}&
	\fig[\sizeS]{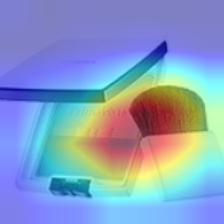} &
	\fig[\sizeS]{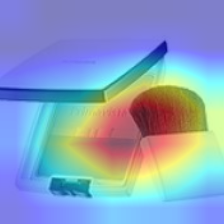} &
	\fig[\sizeS]{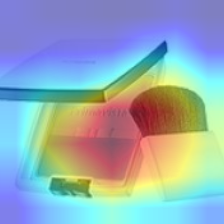} &
	\fig[\sizeS]{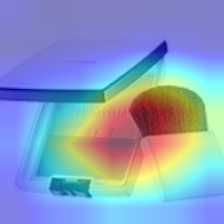} &
	\fig[\sizeS]{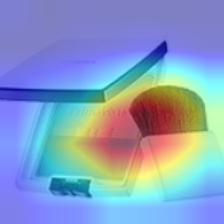} & 
	\fig[\sizeS]{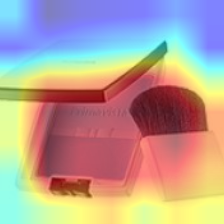}  \\
	Face Powder &&&&&& \\
	\includegraphics[trim={2mm 6mm 2mm 6mm},clip, width=\sizeP\textwidth]{fig/visual/ILSVRC2012_val_00000782.JPEG}&
	\fig[\sizeS]{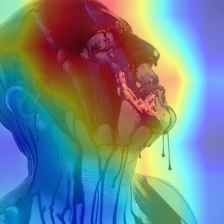} &
	\fig[\sizeS]{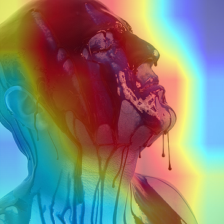} &
	\fig[\sizeS]{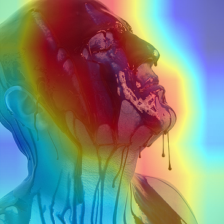} &
	\fig[\sizeS]{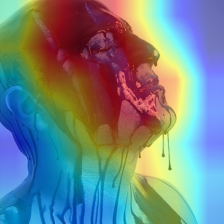} &
	\fig[\sizeS]{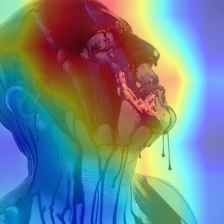} & 
	\fig[\sizeS]{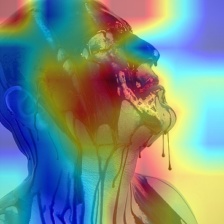}  \\
	Chocolate Sauce &&&&&& \\
	\fig[\sizeS]{visual/ILSVRC2012_val_00001113.png}&
	\fig[\sizeS]{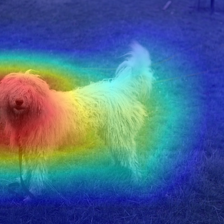} &
	\fig[\sizeS]{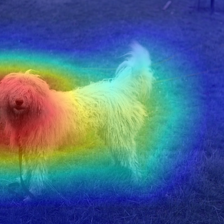} &
	\fig[\sizeS]{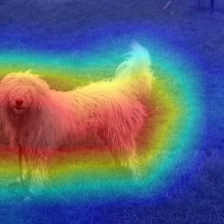} &
	\fig[\sizeS]{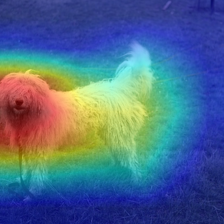} &
	\fig[\sizeS]{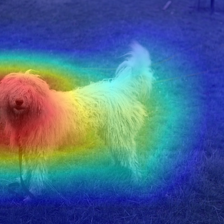} & 
	\fig[\sizeS]{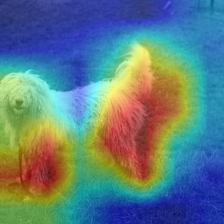}  \\
	Komondor &&&&&& \\
 \includegraphics[trim={14mm 16mm 14mm 12mm},clip, width=\sizeP\textwidth]{fig/visual/ILSVRC2012_val_00001345.JPEG}&
	\fig[\sizeS]{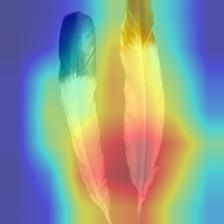} &
	\fig[\sizeS]{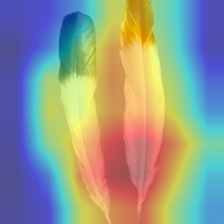} &
	\fig[\sizeS]{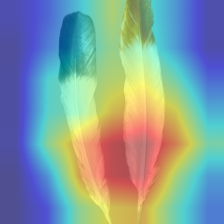} &
	\fig[\sizeS]{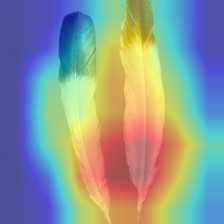} &
	\fig[\sizeS]{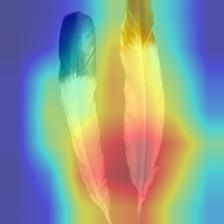} & 
	\fig[\sizeS]{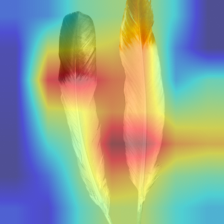}  \\
	Quill &&&&&& \\
 \includegraphics[trim={5mm 14mm 5mm 14mm},clip, width=\sizeP\textwidth]{fig/visual/ILSVRC2012_val_00001529.JPEG}&
	\fig[\sizeS]{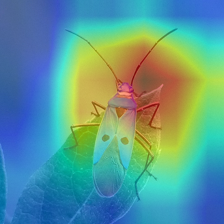} &
	\fig[\sizeS]{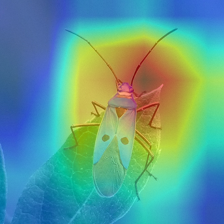} &
	\fig[\sizeS]{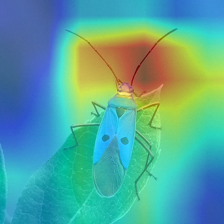} &
	\fig[\sizeS]{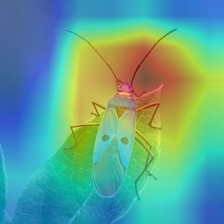} &
	\fig[\sizeS]{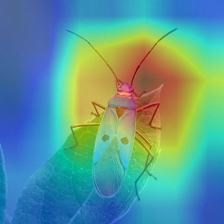} & 
	\fig[\sizeS]{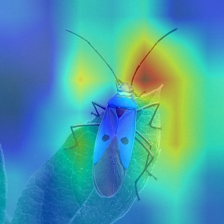}  \\
	Longicorn &&&&&& \\
 \includegraphics[trim={8mm 1mm 8mm 1mm},clip, width=\sizeP\textwidth]{fig/visual/ILSVRC2012_val_00001635.JPEG}&
	\fig[\sizeS]{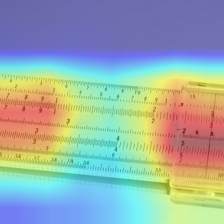} &
	\fig[\sizeS]{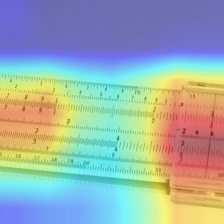} &
	\fig[\sizeS]{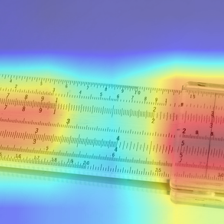} &
	\fig[\sizeS]{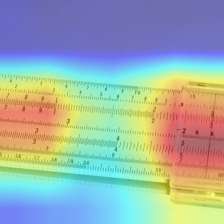} &
	\fig[\sizeS]{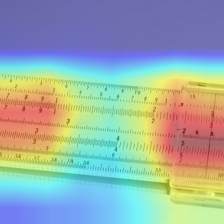} & 
	\fig[\sizeS]{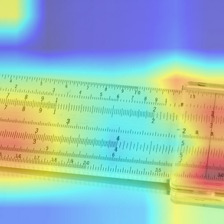}  \\
	Slide Rule &&&&&& \\
\end{tabular}
\caption{Saliency maps obtained from ImageNet example images using different methods on ResNet50.}
\label{fig:imagenet-vis-more-res}
\end{figure*}

\begin{figure*}[t]
\newcommand{\sizeP}{.12}
\newcommand{\sizeS}{.12}
\newcommand{\hh}{.175\textwidth}
\newcommand{\ww}{.200\textwidth}
\tiny
\centering
\setlength{\tabcolsep}{1pt}
\begin{tabular}{cccccccc}
	Input image  &  Grad-CAM  & G-CAM++ & Score-CAM & XG-CAM & Raw Att. & Rollout & Opti-CAM (ours) \\
        \includegraphics[trim={12mm 14mm 12mm 14mm},clip, width=\sizeP\textwidth]{fig/visual/ILSVRC2012_val_00000089.JPEG}&
	\fig[\sizeS]{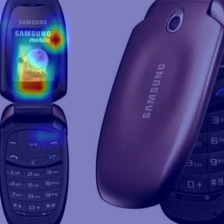} &
	\fig[\sizeS]{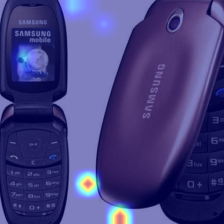} &
	\fig[\sizeS]{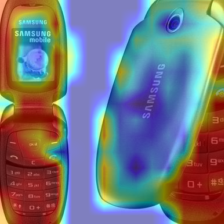} &
	\fig[\sizeS]{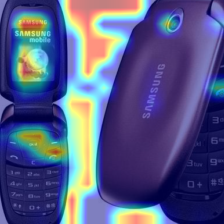} & 
        \fig[\sizeS]{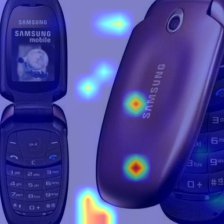} &
        \fig[\sizeS]{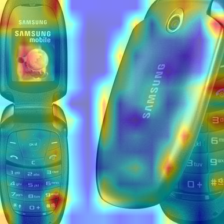} &
	\fig[\sizeS]{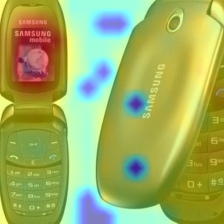}  \\
	Cellphone &&&&&& \\
	\fig[\sizeS]{visual/ILSVRC2012_val_00000748.png}&
	\fig[\sizeS]{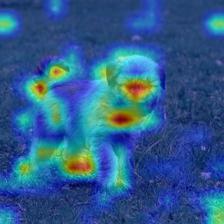} &
	\fig[\sizeS]{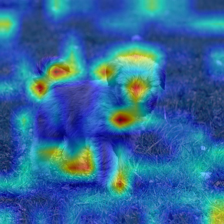} &
	\fig[\sizeS]{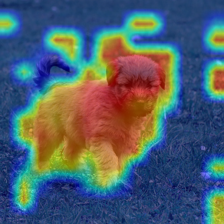} &
	\fig[\sizeS]{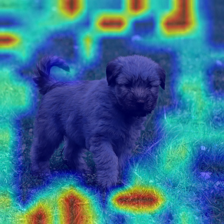} & 
        \fig[\sizeS]{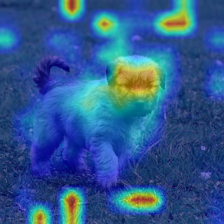} &
        \fig[\sizeS]{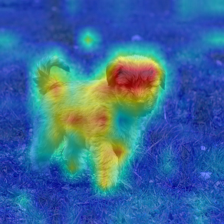} &
	\fig[\sizeS]{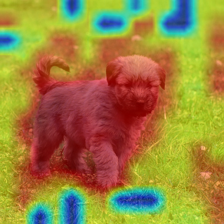}  \\
	Miniature Schnauzer &&&&&& \\
	\includegraphics[trim={2mm 3mm 6mm 1mm},clip, width=\sizeP\textwidth]{fig/visual/ILSVRC2012_val_00000769.JPEG}&
	\fig[\sizeS]{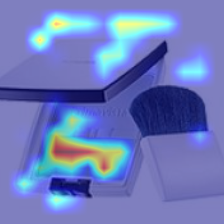} &
	\fig[\sizeS]{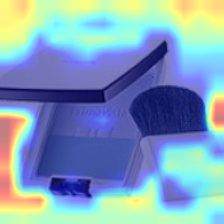} &
	\fig[\sizeS]{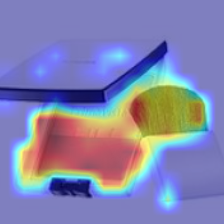} &
	\fig[\sizeS]{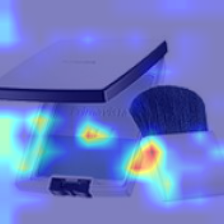} & 
        \fig[\sizeS]{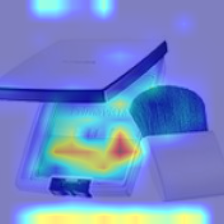} &
        \fig[\sizeS]{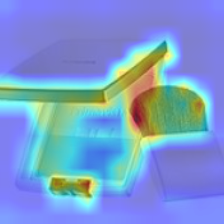} &
	\fig[\sizeS]{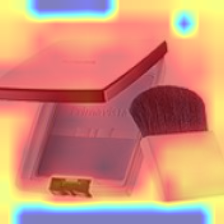}  \\
	Face Powder &&&&&& \\
	\includegraphics[trim={2mm 6mm 2mm 6mm},clip, width=\sizeP\textwidth]{fig/visual/ILSVRC2012_val_00000782.JPEG}&
	\fig[\sizeS]{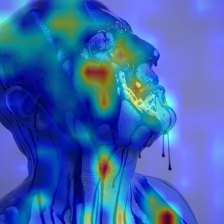} &
	\fig[\sizeS]{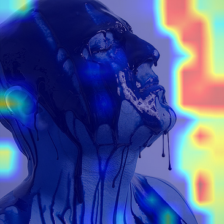} &
	\fig[\sizeS]{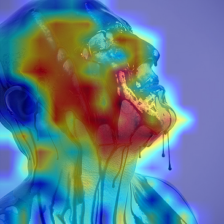} &
	\fig[\sizeS]{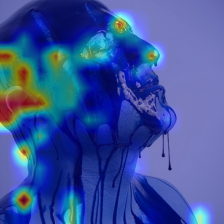} & 
        \fig[\sizeS]{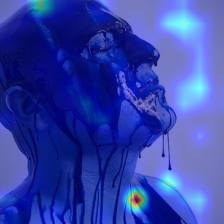} &
        \fig[\sizeS]{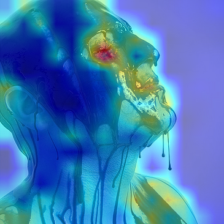} &
	\fig[\sizeS]{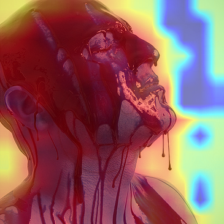}  \\
	Chocolate Sauce &&&&&& \\
	\fig[\sizeS]{visual/ILSVRC2012_val_00001113.png}&
	\fig[\sizeS]{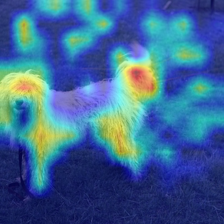} &
	\fig[\sizeS]{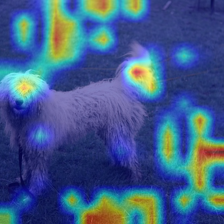} &
	\fig[\sizeS]{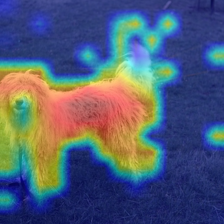} &
	\fig[\sizeS]{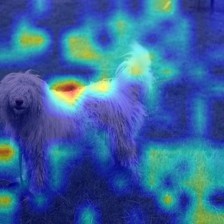} & 
        \fig[\sizeS]{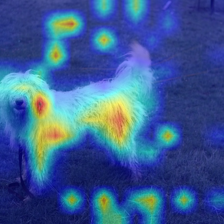} &
        \fig[\sizeS]{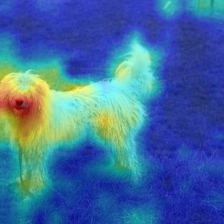} &
	\fig[\sizeS]{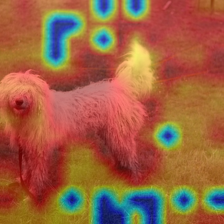}  \\
	Komondor &&&&&& \\
 \includegraphics[trim={14mm 16mm 14mm 12mm},clip, width=\sizeP\textwidth]{fig/visual/ILSVRC2012_val_00001345.JPEG}&
	\fig[\sizeS]{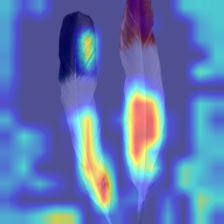} &
	\fig[\sizeS]{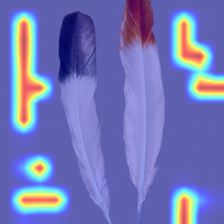} &
	\fig[\sizeS]{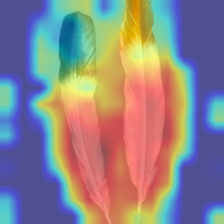} &
	\fig[\sizeS]{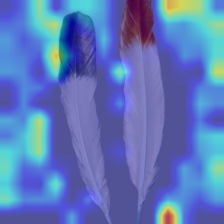} & 
        \fig[\sizeS]{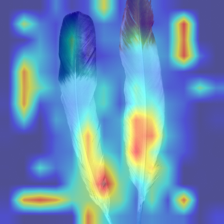} &
        \fig[\sizeS]{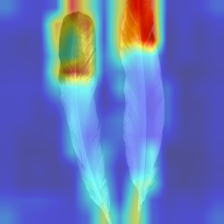} &
	\fig[\sizeS]{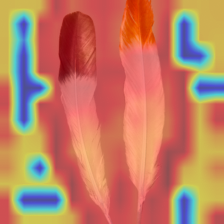}  \\
	Quill &&&&&& \\
 \includegraphics[trim={5mm 14mm 5mm 14mm},clip, width=\sizeP\textwidth]{fig/visual/ILSVRC2012_val_00001529.JPEG}&
	\fig[\sizeS]{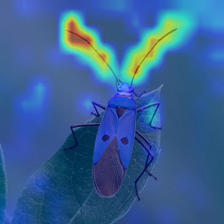} &
	\fig[\sizeS]{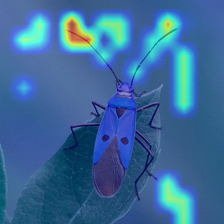} &
	\fig[\sizeS]{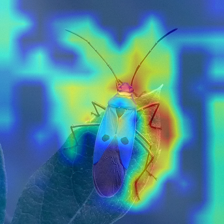} &
	\fig[\sizeS]{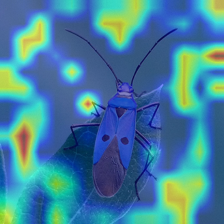} & 
        \fig[\sizeS]{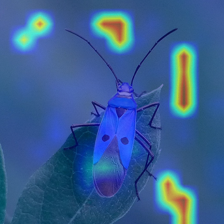} &
        \fig[\sizeS]{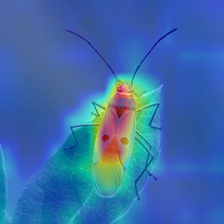} &
	\fig[\sizeS]{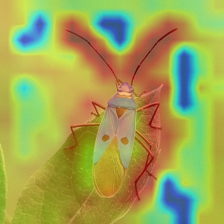}  \\
	Longicorn &&&&&& \\
 \includegraphics[trim={8mm 1mm 8mm 1mm},clip, width=\sizeP\textwidth]{fig/visual/ILSVRC2012_val_00001635.JPEG}&
	\fig[\sizeS]{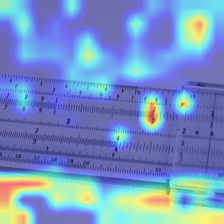} &
	\fig[\sizeS]{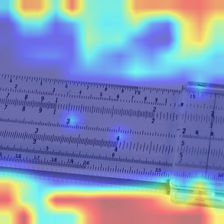} &
	\fig[\sizeS]{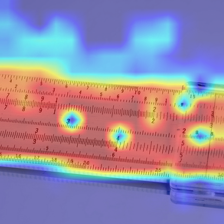} &
	\fig[\sizeS]{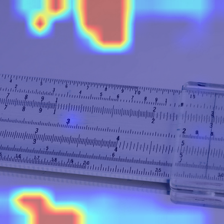} & 
        \fig[\sizeS]{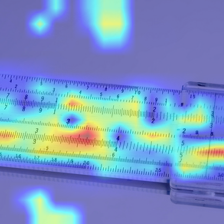} &
        \fig[\sizeS]{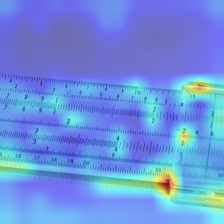} &
	\fig[\sizeS]{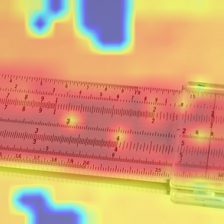}  \\
	Slide Rule &&&&&& \\
\end{tabular}
\caption{Saliency maps obtained from ImageNet example images using different methods on ViT.}
\label{fig:imagenet-vis-more-vit}
\end{figure*}


\section{More visualizations}
\label{sec:more-vis}

\autoref{fig:vis-chest-resnet} and \autoref{fig:vis-kvasir-resnet} present additional visualizations on Chest X-ray and Kvasir datasets using VGG16 and ResNet50. 
Then \autoref{fig:imagenet-vis-more-vgg}, \autoref{fig:imagenet-vis-more-res}, \autoref{fig:imagenet-vis-more-vit} show more results on ImageNet using VGG16, ResNet50, and ViT, respectively.

Overall, we still observe that Opti-CAM captures more of the object area compared with other saliency methods and sometimes background context as well \autoref{fig:imagenet-vis-more-res}.

\end{document}